%% file: main.tex
\documentclass[10pt,journal,compsoc]{IEEEtran}
\usepackage[utf8]{inputenc}

\ifCLASSOPTIONcompsoc
  \usepackage[nocompress]{cite}
\else
  \usepackage{cite}
\fi

\input{definitions}
\usepackage{lineno,hyperref}
\hypersetup{
    colorlinks=true,
    urlcolor=blue,
}
\usepackage{float}
\usepackage{booktabs}
\modulolinenumbers[5]
\usepackage{graphicx}
\usepackage{multirow}

\usepackage{cancel}

\usepackage{numprint}

\usepackage{changes}
\definechangesauthor[name={Rafael Molina}, color=red]{rms}
\setremarkmarkup{(#2)}

\definechangesauthor[name={Pablo Ruiz}, color=violet]{prm}

\definechangesauthor[name={Pablo Morales}, color=orange]{pma}

\begin{document}

\title{Scalable Variational Gaussian Processes\\for Crowdsourcing: Glitch Detection in LIGO}

\author{Pablo Morales-\'Alvarez,
Pablo Ruiz,
Scott Coughlin,\\
Rafael Molina, \IEEEmembership{Senior Member,~IEEE},
Aggelos K. Katsaggelos, \IEEEmembership{Fellow,~IEEE}
\IEEEcompsocitemizethanks{
\IEEEcompsocthanksitem PMA and RM are with the Department of Computer Science and Artificial Intelligence at the University of Granada (Spain).
\IEEEcompsocthanksitem PR and AKK are with the Department of Electrical Engineering and Computer Science at Northwestern University (USA).
\IEEEcompsocthanksitem SC is with the Center for Interdisciplinary Exploration and Research in Astrophysics (CIERA) at Northwestern University, and with the Department of Physics and Astronomy at Cardiff University (UK).
}
\thanks{This work was supported by
the Spanish Ministry of Economy and Competitiveness under project DPI2016-77869-C2-2-R, 
the US National Science Foundation through the NSF INSPIRE 15-47880 grant (Gravity Spy project), 
the US Department of Energy under grant DE-NA0002520,
and the University of Granada through the Visiting Scholar Program.
PMA is supported by La Caixa Banking Foundation (ID 100010434, Barcelona, Spain) through La Caixa Fellowship for Doctoral Studies LCF/BQ/ES17/11600011.
We would also like to thank Mike Zevin and Neda Rohani for useful discussions.
}
}







\IEEEtitleabstractindextext{%
\begin{abstract}

In the last years, the crowdsourcing paradigm is transforming the way classification training sets are obtained. Instead of relying on a single expert annotator, crowdsourcing shares the labelling effort among a large number of collaborators. 
For instance, this is being applied to the data acquired by the laureate Laser Interferometer Gravitational Waves Observatory (LIGO), in order to detect glitches which might hinder the identification of true gravitational-waves. 
The crowdsourcing scenario poses new challenging difficulties, as it has to deal with different opinions from a heterogeneous group of annotators with unknown degrees of expertise.
Probabilistic methods, such as Gaussian Processes (GP), have proven successful in modeling this setting. 
However, GPs do not scale up well to large data sets, which hampers their broad adoption in real-world problems (in particular LIGO).
This has led to the very recent introduction of deep learning based crowdsourcing methods, which have become the state-of-the-art for this type of problems.
However, the accurate uncertainty quantification provided by GPs has been partially sacrificed. This is an important aspect for astrophysicists working with LIGO data, since a glitch detection system should provide very accurate probability distributions of its predictions.
In this work, we leverage the most popular sparse GP approximations to develop a novel GP-based crowdsourcing method that factorizes into mini-batches. This makes it able to cope with previously-prohibitive data sets.
The approach, which we refer to as Scalable Variational Gaussian Processes for Crowdsourcing (SVGPCR), brings back GP-based methods to a state-of-the-art level, and excels at uncertainty quantification.
SVGPCR is shown to outperform deep learning based methods and previous probabilistic approaches when applied to the LIGO data.
Moreover, its behavior and main properties are carefully analyzed in a controlled experiment based on the MNIST data set.

\end{abstract}

}

\maketitle


\section{Introduction}\label{sec:intro}
\input{intro.tex}

\section{Probabilistic model and inference}\label{sec:theory}
\begin{figure}
    \centering
    \includegraphics[width=0.8\columnwidth]{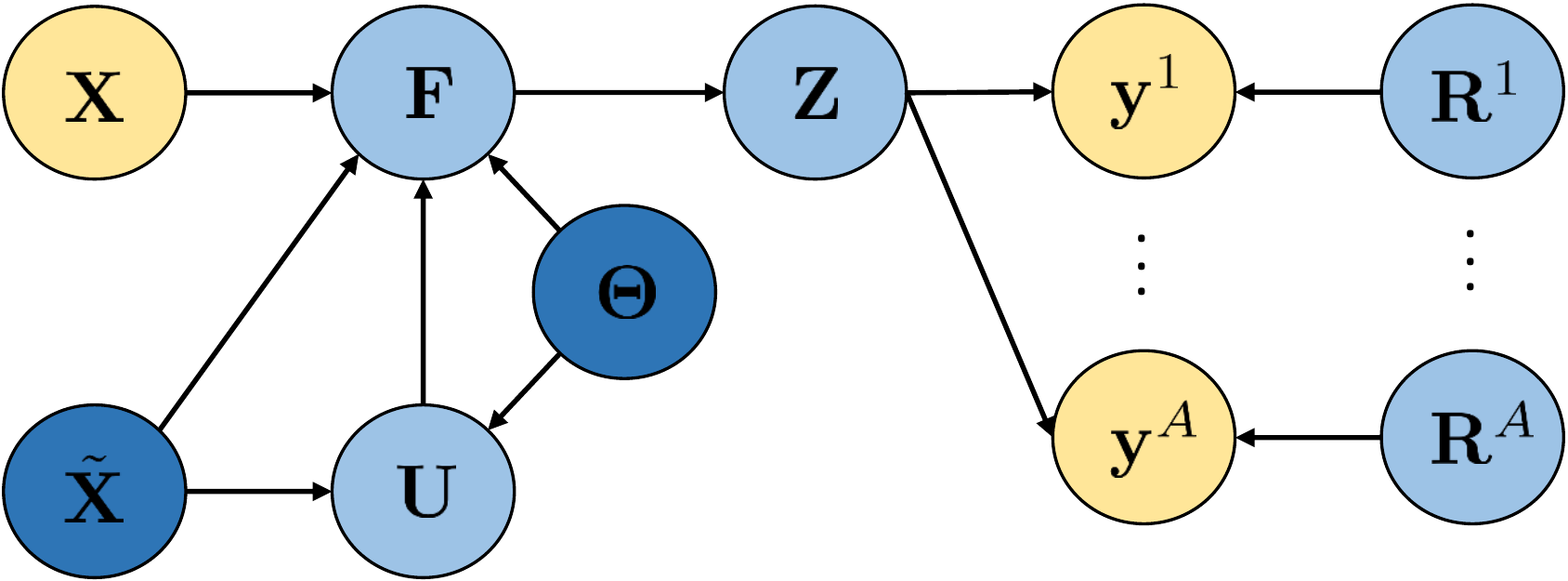}
    \caption{Probabilistic graphical model for the novel SVGPCR. Observed variables are depicted in yellow, and those to be estimated in blue. In the latter case, the intensity indicates whether the estimation is through a posterior distribution (light blue) or a point value (dark).}
    \label{fig:model}
\end{figure}

This section introduces the theoretical formulation of the proposed method. 
It follows the rationale of previous GP-based crowdsourcing approaches \cite{Rodrigues:14, CR_PR}, but achieves scalability through the sparse GP approximation behind SVGP. 
Figure \ref{fig:model} shows a graphical representation of the proposed model, which will be useful throughout this section.

\subsection{The model}\label{sec:model}
\input{model.tex}

\subsection{Variational inference}\label{sec:inference}
\input{inference.tex}



\section{LIGO data description}\label{sec:data}
\input{data.tex}

\section{Experimental results}\label{sec:experiments}
\input{experiments.tex}

\section{Conclusions and future work}\label{sec:conclusions}
\input{conclusion.tex}


\bibliographystyle{IEEEtran}
\bibliography{refs}

\begin{IEEEbiography}[{\includegraphics[width=1in,height=1.25in,keepaspectratio]{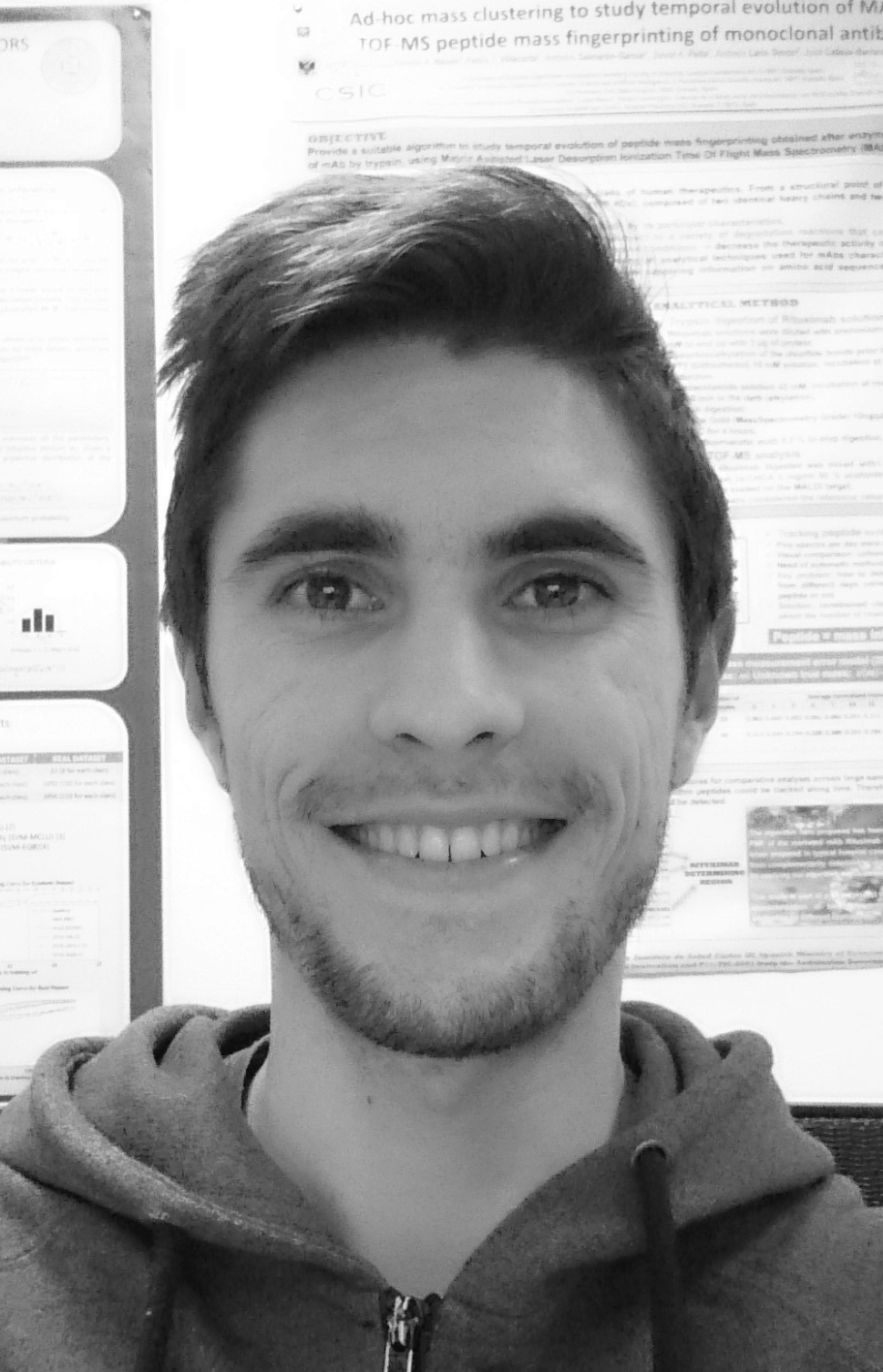}}]%
{Pablo Morales-\'{A}lvarez}
received the B.Sc. degree in Mathematics from the University of Granada (UGR), Spain, in 2014.
He obtained the First End of Studies Award by the Spanish Ministry of Science, Innovation and Universities, to the most outstanding undergraduate in mathematics.
Then, he received the M.Sc. degrees in Mathematical Physics (2015) and Data Science (2016), both from UGR.
Funded by the highly competitive Ph.D. fellowship from \emph{La Caixa} Foundation, he is currently pursuing the Ph.D. degree in the Department of Computer Science and Artificial Intelligence at UGR under the supervision of Prof. Rafael Molina, co-supervised by Prof. Aggelos K. Katsaggelos (Northwestern University, USA).
He was selected in a very competitive process (15\% international acceptance ratio) to attend the 2017 Machine Learning Summer School at the Max Planck Institute for Intelligent Systems at T\"{u}bingen (Germany), organized by Prof. Bernhard Sch\"{o}lkopf's group.
He will be visiting the Machine Learning Group at the University of Cambridge (UK) in 2019, to work with Prof. José Miguel Hernández-Lobato and his team.
His main research interests are probabilistic machine learning methods, specially (Deep) Gaussian Processes and Bayesian Neural Networks, and their applications to crowdsourcing and citizen science problems. 
\end{IEEEbiography}

\begin{IEEEbiography}[{\includegraphics[width=1in,height=1.25in,keepaspectratio]{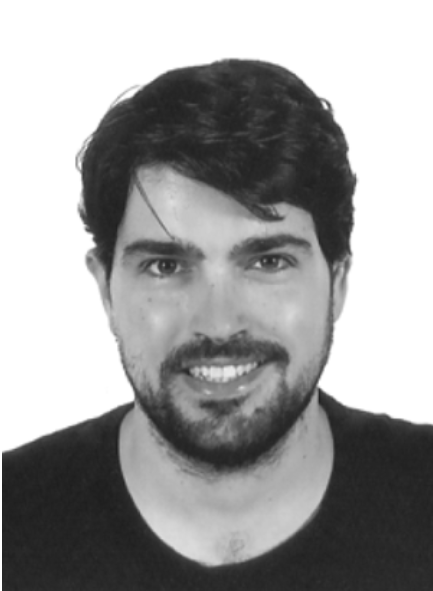}}]%
{Pablo Ruiz}
received his BS Degree in Mathematics (2008), Master Degree in Multimedia Technologies (2009) and PhD in Information and Communication Technologies (2015), all of them from the University of Granada. Currently he is in his third year as a postdoctoral fellow at Northwestern University under the supervision of Prof. Katsaggelos. In his research he studies probabilistic models and inference applied to the solution of inverse problems. Using these tools he is addressing a number of signal and image processing problems such as image restoration, blind deconvolution, image and 3D volume restoration from images acquired with computational tomography, phase retrieval, as well as Machine Learning problems, where he is using the Gaussian Processes theory to solve regression and classification problems, active learning, feature learning and crowdsourcing among others.
\end{IEEEbiography}

\begin{IEEEbiography}[{\includegraphics[width=1in,height=1.25in,keepaspectratio]{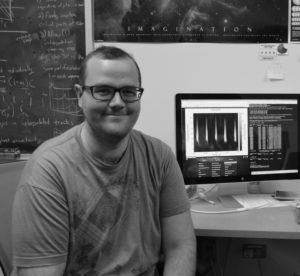}}]%
{Scott Coughlin}
received a BA degree from Northwestern University, in 2014. He is currently working as a research specialist at the Center for Interdisciplinary Exploration and Research in Astrophysics (CIERA, Northwestern University), and enrolled as a graduate student at the School of Physics and Astronomy at Cardiff University (UK). He leads and manages the citizen science project known as Gravity Spy. Gravity Spy is an interdisciplinary research project that combines his interests of gravitational-wave data analysis, machine learning and citizen science with the goal of classifying non-astrophysical noise artifacts in gravitational wave data.
\end{IEEEbiography}

\begin{IEEEbiography}[{\includegraphics[width=1in,height=1.25in,keepaspectratio]{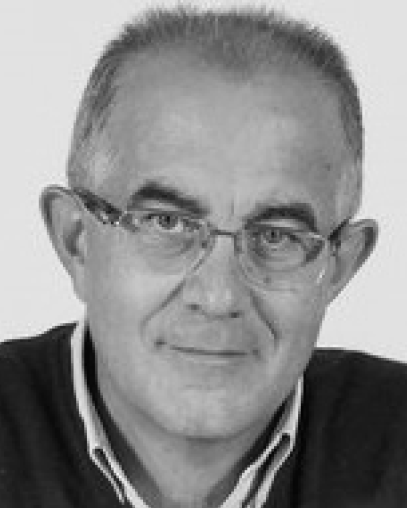}}]%
{Rafael Molina}
received the degree in mathematics
(statistics) and the Ph.D. degree in optimal design
in linear models from the University of Granada,
Granada, Spain, in 1979 and 1983, respectively.
He was the Dean of the Computer Engineering
School, University of Granada, from 1992 to 2002.
In 2000, he joined the University of Granada, as a
Professor of computer science and artificial intelligence.
He was the Head of the Computer Science
and Artificial Intelligence Department, University of
Granada, from 2005 to 2007. He has co-authored the
paper that received the runner-up prize at reception for early stage researchers
at the House of Commons. His research interest focuses mainly on using
Bayesian modeling and inference in problems like image restoration (applications
to astronomy and medicine), super-resolution of images and video,
blind deconvolution, computational photography, source recovery in medicine,
compressive sensing, low-rank matrix decomposition, active learning, fusion,
and machine learning. He was a recipient of the IEEE International Conference
on Image Processing Paper Award in 2007 and the ISPA Best Paper Award
in 2009. He served as an Associate Editor for Applied Signal Processing from
2005 to 2007 and the IEEE TRANSACTIONS ON IMAGE PROCESSING from
2010 to 2014. He has been serving as an Associate Editor for Progress in
Artificial Intelligence since 2011. He has been serving as an Area Editor for
Digital Signal Processing since 2011.
\end{IEEEbiography}

\begin{IEEEbiography}[{\includegraphics[width=1in,height=1.25in,keepaspectratio]{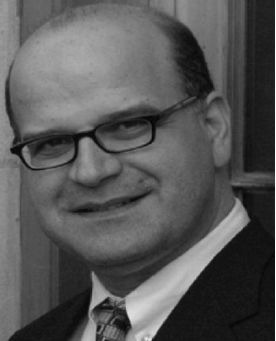}}]%
{Aggelos K. Katsaggelos}
received the Diploma
degree in electrical and mechanical engineering from
the Aristotelian University of Thessaloniki, Greece,
in 1979, and the M.S. and Ph.D. degrees in electrical
engineering from the Georgia Institute of Technology
in 1981 and 1985, respectively.
In 1985, he joined the Department of Electrical
Engineering and Computer Science, Northwestern
University. He was the Ameritech Chair of information
technology and the AT\&T Chair. He is currently
a Professor of the Joseph Cummings Chair, Northwestern
University. He is also an Academic Staff Member with NorthShore
University Health System and an Affiliated Faculty Member with the Department
of Linguistics. He has an appointment with the Argonne National
Laboratory. He has published extensively in the areas of multimedia signal
processing and communications, computational imaging, and machine learning
(over 250 journal papers, 600 conference papers, and 40 book chapters).
He holds 25 international patents. He has co-authored Rate-Distortion Based
Video Compression (Kluwer, 1997), Super-Resolution for Images and Video
(Claypool, 2007), Joint Source-Channel Video Transmission (Claypool, 2007),
and Machine Learning Refined (Cambridge University Press, 2016). He has
supervised 56 Ph.D. dissertations.
Among his many professional activities, he was the Editor-in-Chief of the
IEEE Signal Processing Magazine from 1997 to 2002, a BOG Member of
the IEEE Signal Processing Society from 1999 to 2001, a member of the
Publication Board of the IEEE PROCEEDINGS from 2003 to 2007, and a
member of the Award Board of the IEEE Signal Processing Society. He is
a fellow of the IEEE in 1998, SPIE in 2009, EURASIP in 2017, and OSA
in 2018. He was a recipient of the IEEE Third Millennium Medal in 2000, the
IEEE Signal Processing Society Meritorious Service Award in 2001, the IEEE
Signal Processing Society Technical Achievement Award in 2010, the IEEE
Signal Processing Society Best Paper Award in 2001, the IEEE ICME Paper
Award in 2006, the IEEE ICIP Paper Award in 2007, the ISPA Paper Award
in 2009, and the EUSIPCO Paper Award in 2013. He was a Distinguished
Lecturer of the IEEE Signal Processing Society from 2007 to 2008.
\end{IEEEbiography}

\enlargethispage{-9.5cm}


\end{document}


\title{Scalable Variational Gaussian Processes\\for Crowdsourcing: Glitch Detection in LIGO.\\
\emph{Supplemental material.}}

\author{Pablo Morales-\'Alvarez, Pablo Ruiz, Scott Coughlin,\\ Rafael Molina \IEEEmembership{Senior Member,~IEEE}, Aggelos K. Katsaggelos \IEEEmembership{Fellow,~IEEE}
\thanks{This work was supported by
the Spanish Ministry of Economy and Competitiveness under project DPI2016-77869-C2-2-R, 
the US National Science Foundation through the NSF INSPIRE 15-47880 grant (Gravity Spy project), 
the US Department of Energy under grant DE-NA0002520,
and the University of Granada through the Visiting Scholar Program.
PMA is supported by La Caixa Banking Foundation (ID 100010434, Barcelona, Spain) through La Caixa Fellowship for Doctoral Studies LCF/BQ/ES17/11600011.
We would also like to thank Mike Zevin and Neda Rohani for useful discussions.}
\thanks{PMA and RM are with the Department of Computer Science and Artificial Intelligence at the University of Granada (Spain). PR and AKK are with the Department of Electrical Engineering and Computer Science at Northwestern University (USA).
SC is with the Center for Interdisciplinary Exploration and Research in Astrophysics (CIERA) at Northwestern University, and with the Department of Physics and Astronomy at Cardiff University (UK). 
}
}







\maketitle

This supplemental material contains two sections. The first one provides a detailed derivation for the ELBO, recall eq. (15) in the article.
Then, section \ref{sec:terms} presents the explicit expression for each term of the ELBO.
These expressions are implemented in the GPflow code that will be released.

\section{Derivation of the ELBO}\label{sec:ELBO}
Let us derive step by step the eq. (15) in the paper:

\begin{align}
\mathrm{ELBO}(\bV,\bTheta,\tilde\bX)&=
\mathbb{E}_{\q(\bZ)\p(\bF|\bU)\q(\bU)\q(\bR)}\left(
    \log\frac{\p(\bY|\bZ,\bR)\p(\bZ|\bF)\cancel{\p(\bF|\bU)}\p(\bU)\p(\bR)}{\q(\bZ)\cancel{\p(\bF|\bU)}\q(\bU)\q(\bR)}
    \right)
\nonumber\\
    &\hspace*{2cm}\textit{[Grouping and integrating straightforward terms]}\nonumber\\
    &=
    \mathbb{E}_{\q(\bZ)\q(\bR)}\log\p(\bY|\bZ,\bR)+
    \mathbb{E}_{\q(\bZ)\p(\bF|\bU)\q(\bU)}\log\p(\bZ|\bF)-
    \mathbb{E}_{\q(\bZ)}\log\q(\bZ)\nonumber\\
    &+\mathbb{E}_{\q(\bU)}\left(\log\frac{\p(\bU)}{\q(\bU)}\right)+
    \mathbb{E}_{\q(\bR)}\left(\log\frac{\p(\bR)}{\q(\bR)}\right)\nonumber\\
    &\hspace*{2cm}\textit{[Using the factorizations of  the different terms and the definition of the KL divergence]}\nonumber\\
    &=\sum_{n=1}^N\sum_{a\in A_n}\sum_{\by\in\bY_n^a}\mathbb{E}_{\q(\bz_n)\q(\bR^a)}\log\p(\by|\bz_n,\bR^a)+
    \sum_{n=1}^N\mathbb{E}_{\q(\bz_n)\q(\bff_{n,:})}\log\p(\bz_n|\bff_{n,:})-
    \sum_{n=1}^N \mathbb{E}_{\q(\bz_n)}\log\q(\bz_n)
    \nonumber\\
    &
    -\sum_{k=1}^K \textrm{KL}(\q(\bu_k)||\p(\bu_k))-
    \sum_{a=1}^A\sum_{k=1}^K \textrm{KL}(\q(\br_k^a)||\p(\br_k^a)) \nonumber\\ 
    &\hspace*{2cm}\textit{[Explicitly taking the expectation over the discrete distribution $\q(\bz_n)$]}\nonumber\\
    &= \sum_{n=1}^N\sum_{a\in A_n}\sum_{\by\in\bY_n^a}\sum_{k=1}^K q_{nk}\mathbb{E}_{\q(\br_k^a)}\log\p(\by|\be_k,\br_k^a)+
    \sum_{n=1}^N\sum_{k=1}^K q_{nk} \mathbb{E}_{\q(\bff_{n,:})}\log\p(\be_k|\bff_{n,:})-
    \sum_{n=1}^N\sum_{k=1}^K q_{nk}\log q_{nk}
    \nonumber\\
    &
    -\sum_{k=1}^K \mathrm{KL}(\q(\bu_k)||\p(\bu_k))
    -\sum_{a=1}^A\sum_{k=1}^K\mathrm{KL}(\q(\br_k^a)||\p(\br_k^a)). \label{eq:ELBO_final}
\end{align}

Indeed, this final expression is the same formula as that in eq. (15) in the article. The posterior over $\bF$, $\q(\bF)$, whose marginal $\q(\bff_{n,:})$ is used in the second term, can be easily obtained in closed form, since both $\q(\bF|\bU)=\p(\bF|\bU)$ and $\q(\bU)$ are Gaussians:

\begin{align*}
    \q(\bF)=\prod_{k=1}^K \q(\bff_k),\quad\mathrm{with}\quad
    \q(\bff_k)=\mathcal{N}\left(\bff_k|\bB\cdot\bm_k,\bK_{\bX\bX}+\bB(\bS_k-\bK_{\tilde\bX\tilde\bX})\bB^\intercal\right),\quad\bB=\bK_{\bX\tilde\bX}\bK_{\tilde\bX\tilde\bX}^{-1}.
\end{align*}

\section{Explicit expressions for the ELBO terms}\label{sec:terms}

In order to maximize the ELBO w.r.t. the variational paremeters $\bV$, the kernel parameters $\bTheta$, and the inducing locations $\tilde\bX$, we need explicit expressions on these variables.
All the five terms in eq.~\eqref{eq:ELBO_final} but the second one can be analytically expressed in terms of $\bV$, $\bTheta$, $\tilde\bX$.
Moreover, the second one admits an approximated expression through Gaussian-Hermite quadrature \cite{gaussHermite}.

For the first term, we use the expectation of the logarithm of a component of a Dirichlet distribution (which is in the exponential family): if $(X_1,\dots,X_S)\rightarrow \textrm{Dir}(\gamma_1,\dots,\gamma_S)$, then $\mathbb{E}(\log X_i) = \psi(\gamma_i)-\psi(\sum_{i=1}^S \gamma_i)$ \cite{Bishop:06}. Here, $\psi$ is the digamma function, which is the logarithmic derivative of the gamma function and is implemented in TensorFlow as \texttt{tf.math.digamma}. With this, we have:
\begin{equation*}
    q_{nk}\cdot\mathbb{E}_{\q(\br_k^a)}\log\p(\by|\be_k,\br_k^a) = 
    q_{nk}\cdot\prod_{i=1}^K\left(\psi(\tilde\alpha^a_{ik})-\psi\left(\sum_{c=1}^K\tilde\alpha^a_{ck}\right) \right)^{y_i},
\end{equation*}
which depends on $\{\tilde\alpha_{ij}^a\}$ and $\{q_{nk}\}$, recall table 1 in the article.

The second term involves the expectation of the log-likelihood over a Gaussian.
If the likelihood is also defined through a Gaussian (like in most regression problems) then this can be obtained in closed-form. 
However, this is not possible for the classification likelihoods that are used in crowdsourcing. 
In this case, Gaussian-Hermite quadrature is applied in order to approximate the result \cite{gaussHermite}. 
For the particular case of our RobustMax likelihood, the approximation is already integrated within the GPflow library, see the \texttt{variational\_expectations} method of the \texttt{Multiclass} likelihood class \cite{matthews2017gpflow}.
As summarized in table 1 in the main paper, this second term depends on $\{q_{nk}\}$, $\{\bm_k,\bS_k\}$, $\bTheta$, and $\tilde\bX$.
The third term is already expressed in terms of $\{q_{nk}\}$ in eq.~\eqref{eq:ELBO_final}.

For the fourth term, only the well-known expression of the KL divergence between (multivariate) Gaussians is needed \cite{Rasmussen:06}:
\begin{equation*}
    \textrm{KL}(\q(\bu_k)||\p(\bu_k)) = 
    \textrm{KL}\left(\mathcal{N}(\bmu_k,\bS_k)||\mathcal{N}(\mathbf{0},\bK)\right) = 
    \frac{1}{2}\left(
    \textrm{tr}\left(\bK^{-1}\bS_k\right)+
    \bmu_k^\intercal\bK^{-1}\bmu_k - M
    +\log\left(\frac{\textrm{det }\bK}{\textrm{det }\bS_k} \right)
    \right),
\end{equation*}
where $\bK = \bK_{\tilde\bX\tilde\bX}$ is the prior covariance matrix for the inducing points, and $M$ is the dimension of the variable $\bu_k$ (i.e. the amount of inducing points).
This expression depends on $\{\bm_k,\bS_k\}$, $\bTheta$, and $\tilde\bX$ (the last two through $\bK$).

In the fifth term, it is straightforward to obtain a closed-form expression for the KL divergence between Dirichlet distributions:
\begin{equation*}
    \mathrm{KL}(\q(\br_k^a)||\p(\br_k^a))=
    \log B(\balpha_{k}^a)-
    \log B(\tilde\balpha_{k}^a)+
    \sum_{i=1}^K(\tilde\alpha_{ik}^a-\alpha_{ik}^a)\cdot\left(\psi(\tilde\alpha_{ik}^a)-\psi\left(\sum_{j=1}^K \tilde\alpha_{jk}^a \right)\right).
\end{equation*}
Here, $\psi$ is the aforementioned digamma function, and $B$ is the beta function, which is also implemented in TensorFlow and is defined as $B(\balpha)=B(\alpha_1,\dots,\alpha_K)=(\prod_i \Gamma(\alpha_i))/\Gamma(\sum_i \alpha_i)$, with $\Gamma$ the Gamma function.
This term only depends on $\{\tilde\alpha_{ij}^a\}$.

\bibliographystyle{IEEEtran}
\bibliography{refs}

%% file: definitions.tex
\usepackage{bm}
\usepackage{color}
\usepackage{amssymb}

\usepackage{algorithmic}
\usepackage{algorithm}
\usepackage[cmex10]{amsmath}
\usepackage{url}

\usepackage{multirow}

\def\p{{\mathrm{p}}}
\def\q{{\mathrm{q}}}

\def\balpha{{\boldsymbol{\alpha}}}

\def\btheta{{\boldsymbol{\theta}}}

\def\bnu{{\boldsymbol{\nu}}}

\def\bTheta{{\boldsymbol{\Theta}}}

\def\be{{\mathbf e}}
\def\bff{{\mathbf f}}

\def\bm{{\mathbf m}}

\def\bq{{\mathbf q}}
\def\br{{\mathbf r}}

\def\bu{{\mathbf u}}

\def\bx{{\mathbf x}}
\def\by{{\mathbf y}}
\def\bz{{\mathbf z}}

\def\bB{{\mathbf B}}

\def\bF{{\mathbf F}}

\def\bK{{\mathbf K}}

\def\bR{{\mathbf R}}
\def\bS{{\mathbf S}}

\def\bU{{\mathbf U}}
\def\bV{{\mathbf V}}

\def\bX{{\mathbf X}}
\def\bY{{\mathbf Y}}
\def\bZ{{\mathbf Z}}

\newcommand{\R}{\mathbb{R}}

%% file: intro.tex
Crowdsourcing, also known as citizen science, is revolutionizing the way real-world data sets are obtained nowadays \cite{irwin2018no, Guerrini134}.
Traditionally, the task of labelling has been accomplished by a single expert annotator in a process that is time-consuming, expensive and difficult to scale.
The proliferation of web services such as Amazon Mechanical Turk (\url{www.mturk.com})
and Figure-Eight (\url{www.figure-eight.com}, formerly Crowdflower) allows for outsourcing this process to a distributed workforce that can collaborate virtually, sharing the effort among a huge number of annotators \cite{Snow:08, buhrmester2011amazon}.
This approach is rapidly growing in popularity, and is being applied to many different fields such as medical imaging \cite{Albarqouni:16}, genetics \cite{saez2016crowdsourcing}, remote sensing \cite{fritz2017global}, topic modelling \cite{Rodrigues:2017}, and object segmentation \cite{heim2018clickstream}.

\begin{figure}
\centering
    \begin{tabular}{cc}
     \includegraphics[width=0.48\columnwidth]{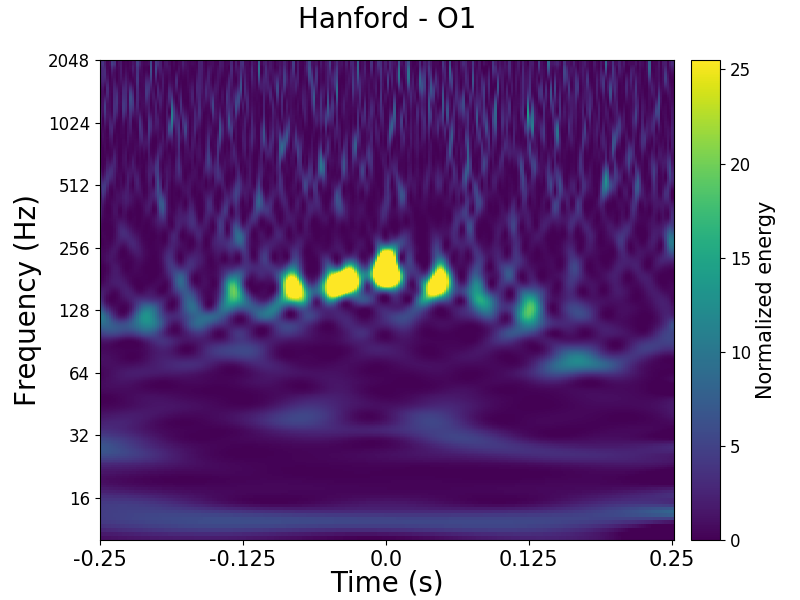} & 
     \includegraphics[width=0.48\columnwidth]{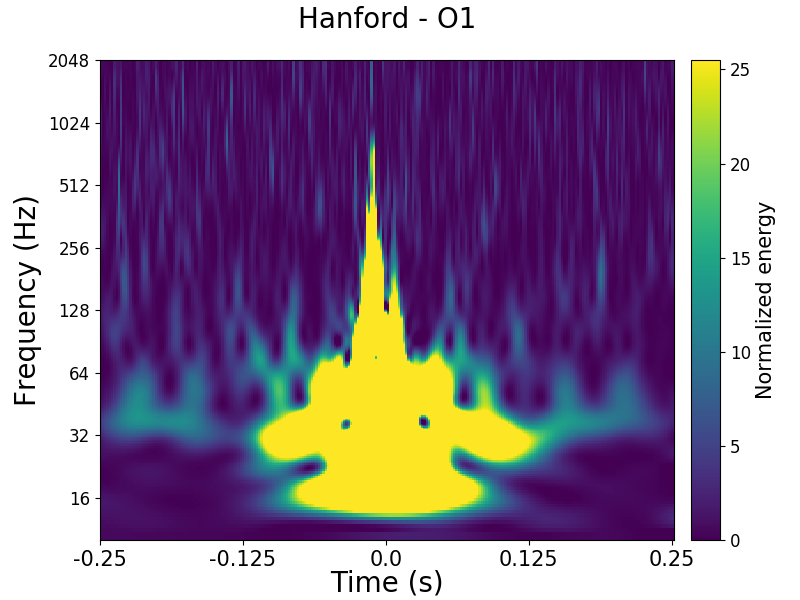} 
    \end{tabular}
    \caption{Two examples of glitches observed by the LIGO detector. The fifteen types considered in this work will be carefully described in section \ref{sec:data}, see also figure \ref{fig:glitches_full}.}
    \label{fig:glitches_intro}
\end{figure}

A very recent application of crowdsourcing in the field of astrophysics is the GravitySpy project \cite{LIGO:17}, which aims at detecting glitches in the Laser Interferometer Gravitational Waves Observatory (LIGO).
The LIGO collaboration is one of the most exciting and recognized scientific international initiatives \cite{Abramovici325}. It was awarded the 2017 Physics Nobel Prize for the first empirical detection of a gravitational-wave in September 2015 \cite{gwDiscovery}. These waves are ripples in the fabric of spacetime, their existence was theoretically predicted by Einstein's General Relativity theory in 1916, and open a whole new way to explore the universe (beyond the electromagnetic signals available so far) \cite{castelvecchi2016einstein}. 
However, the LIGO detector is equipped with extremely delicate technology, which is sensitive to many different sources of noise.
This produces a wide variety of glitches, see figure \ref{fig:glitches_intro}, which make the detection of true gravitational-waves difficult. 
The goal of GravitySpy is to leverage citizen science to label the large data set of glitches produced by LIGO, and then develop a machine learning system (based on crowdsourcing methods) to help astrophysicists classify them \cite{LIGO:17}.

The crowdsourcing scenario introduces new challenges in machine learning, such as combining the unknown expertise of annotators, dealing with disagreements on the labelled samples, or detecting the existence of spammer and adversarial annotators \cite{Sheng:08, Donmez:08}.
The first approaches to deal with multiple-annotated data used to rely on some kind of
label aggregation mechanism prior to training. 
The most straightforward one is majority voting, which assumes that every annotator is equally reliable.
More elaborated methods consider the biases of the different annotators, yielding a better calibrated set of training labels, see \cite{Dawid:79} (which is usually considered the first crowdsourcing work) and \cite{Ipeirotis:10, Whitehill:09}. In all these cases, the idea is to obtain a set of clean true labels, which are then fed to the preferred standard (no-crowdsourcing) classification algorithm.

However, recent works show that jointly modelling the classifier and the annotators behavior leads to superior performance, since the features provide information to puzzle out the true labels \cite{Raykar:10, Yan:14}.
In this joint model, Bayesian methods based on Gaussian Processes (GPs) have proved extremely successful to accurately quantify uncertainty \cite{Rodrigues:14, CR_PR, Rodrigues:2017}.
However, in real-world applications they have been gradually replaced by deep learning based approaches \cite{Albarqouni:16, Rodrigues:18, guan2018said}, since GPs do not scale well to large data sets \cite{Rodrigues:2017, Rasmussen:06}.
As a result, the sound probabilistic formulation of GPs has been sacrificed. 
However, large scale problems could greatly benefit from such a solid modelling.
In particular, in order to develop a reliable glitch detection system, astrophysicists with the GravitySpy project are particularly interested in the Bayesian formulation given by GPs \cite{LIGO:17}.
Therefore, their scalability issues must be addressed.


GP is a popular Bayesian non-parametric model for supervised learning that excels at uncertainty quantification \cite{Rasmussen:06}.
Due to the kernel matrix inversion, its computational complexity at training is $\mathcal{O}(N^3)$, where $N$ is the size of the training set.
This makes it prohibitive for practical applications beyond a few thousands data points \cite{damianou2015deep}. 
To overcome this problem, different sparse GP approximations have been proposed in the last years \cite{bauer:2016}. 
Most of them rely on the notion of inducing points, a reduced set of $M$ ($M\ll N$) instances which condense the information contained in the whole data set 
\cite{snelson:06, titsias:09}\footnote{There exist other sparse GP approximations which alternatively rely on Fourier features \cite{Morales:18}, and which have been already used for crowdsourcing problems \cite{CR_INFFUS}. In the experiments, the proposed method will be shown to clearly outperform these alternative approaches too.}.
One of the most widespread methods is the Scalable (sometimes also referred to as Sparse) Variational Gaussian Processes (SVGP) method \cite{Hensman:15}, which uses Variational Inference \cite{blei2017variational} and is inspired by the earlier regression method \cite{hensman:2013}.
SVGP stands out as one of the most efficient GP approximations, since it can be trained through mini-batches and yields a training computational cost of $\mathcal{O}(N_b M^2)$, with $N_b$ the mini-batch size. This allows SVGP to handle data sets of practically any size \cite{Hensman:15}.

In this work, we extend the sparse GP approximation behind SVGP to the multiple-annotated crowdsourcing setting. 
Interestingly, the form of the Evidence Lower Bound (ELBO) is still suitable for Stochastic Variational Inference \cite{hoffman2013stochastic}, which allows for training through mini-batches.
To the best of our knowledge, this allows GPs to be used for crowdsourcing problems of virtually any size for the first time.
This novel method is refered to as Scalable (or Sparse) Variational Gaussian Processes for Crowdsourcing (SVGPCR). 
The annotators noise model is also fully Bayesian, described by per-user confusion matrices which are assigned Dirichlet priors. The underlying true labels are modelled in a probabilistic manner as well. Variational inference \cite{blei2017variational} is used to approximate the posterior distribution of the model. 

In order to deal with the LIGO data, SVGPCR is modelled and implemented as a multi-class method.
The implementation is based on GPflow, a very popular GP library that benefits from GPU acceleration through TensorFlow \cite{matthews2017gpflow}.
Three sets of experiments are provided. 
First, a controlled crowdsourcing problem specified for MNIST illustrates the main properties and behavior of SVGPCR. Among these, we may highlight its accurate identification of annotators' expertise degree, reconstruction of the real underlying label, and how the number of inducing points influences its performance.
Secondly, SVGPCR is compared against previous probabilistic crowdsourcing methods in a relevant binary LIGO problem\footnote{Most of these previous probabilistic crowdsourcing approaches were originally proposed for binary problems, and the code is available accordingly.}. SVGPCR stands out as the indisputably best performing approach, thanks to its innovative scalability through mini-batches.
Third, SVGPCR is shown to outperform state-of-the-art DL-based methods in the full LIGO data set, specially in terms of test likelihood, due to the more robust uncertainty control.

The rest of the paper is organized as follows.
Section \ref{sec:theory} describes the proposed model and inference procedure. 
Section \ref{sec:data} presents the LIGO data available in the GravitySpy project.
The experimental results are discussed in Section \ref{sec:experiments}.
Finally, Section \ref{sec:conclusions} contains some remarks and future outlook.


%% file: model.tex
In a crowdsourcing problem with $K$ classes, we observe training data $\mathcal{D}=\{(\bx_n,\bY_n^a):\; n=1,\dots,N;\; a\in A_n\}$,
where $\bx_n\in\R^D$ are the training features, and $\bY_n^a$ is the set\footnote{Notice that annotators are allowed to label the same instance more than once (possibly with different labels). This happens in a few cases in the LIGO data.} of annotations provided by the $a$-th annotator for the $n$-th instance.
That is, each $\by\in\bY_n^a$ is an one-hot encoded vector in $\{\be_1,\ldots,\be_K\}$ that represents the $k$-th class (i.e., all elements of $\be_k$ are zero but the $k$-th one, which is one).
There are $N$ data points, $A$ annotators, and $A_n\subseteq \{1,\dots,A\}$ contains the annotators that labelled the $n$-th instance. All training instances will be grouped in $\bX = \{\bx_n: n=1,\dots,N\}$, and analogously all annotations in $\bY = \{\bY_n^a: n=1,\dots,N; a\in A_n\}$.

As with previous approaches \cite{Rodrigues:2017, Raykar:10, Yan:14, Rodrigues:14, CR_PR}, the proposed model assumes an (unknown) real label for each instance, $\bz_n\in\{\be_1,\ldots,\be_K\}$.
The actual annotations depend on this real label and the degree of expertise of each annotator, which is modelled by the confusion matrix $\bR^a=(r_{ij}^a)_{1\leq i,j\leq K}$.
Each $r_{ij}^a\in[0,1]$ represents the probability that the $a$-th annotator labels as class $i$ an instance whose real class is $j$. 
Notice that this matrix must add up to one by columns.
Mathematically, this is given by 
\begin{equation}\label{eq:annotations_ind}
    \p(\bY_n^a|\bz_n,\bR^a) = \prod_{\by\in\bY_n^a} \by^\intercal\bR^a \bz_n.
\end{equation}

Assuming that all annotators label the different instances independently, we have
\begin{equation}\label{eq:annotations}
    \p(\bY|\bZ,\bR)=
    \prod_{n=1}^N\prod_{a\in A_n}\p(\bY_n^a|\bz_n,\bR^a),
\end{equation}
where $\bZ=\{\bz_n: n=1,\dots,N\}$ and $\bR=\{\bR^a:a=1,\dots,A\}$ group the corresponding individual variables, and $\p(\bY_n^a|\bz_n,\bR^a)$ is given by eq.~\eqref{eq:annotations_ind}. 

The prior distribution for the behavior of annotators is modelled through (independent) Dirichlet distributions, which are conjugate to the categorical one in eq.~\eqref{eq:annotations_ind} \cite{Bishop:06}. This yields
\begin{equation}\label{eq:annot_prior}
    \p(\bR)=
    \prod_{a=1}^A\prod_{j=1}^K \p(\br^a_j) =
    \prod_{a=1}^A\prod_{j=1}^K \mathrm{Dir}(\br^a_j|\alpha_{1j}^a,\dots,\alpha_{Kj}^a),
\end{equation}
where $\br^a_j=(r^a_{1j},\dots,r^a_{Kj})^\intercal$ denotes the $j$-th column of the confusion matrix $\bR^a$.
The hyperparameters $\balpha=\{\alpha_{ij}^a:i,j=1,\dots,K,\; a=1,\dots,A\}$ codify any prior belief on the behavior of the annotators. If this is not available, the default choice $\alpha_{ij}^a=1$ corresponds to uniform distributions. 

For each instance, the true underlying label $\bz$ is modelled through $K$ latent variables $f_1,\dots,f_K$.
Both parts are related by means of the likelihood model
\begin{equation}\label{eq:likelihood_ind}
    \p(\bz|f_1,\dots,f_K)=\bz^\intercal \bnu(f_1,\dots,f_K),
\end{equation}
where $\bnu(f_1,\dots,f_K)$ is any vector with $K$ positive components that add up to $1$. In this work we will use the popular robust-max likelihood, which prevents overfitting in GP classification \cite{hernandez2011robust}. It is given by $\bnu(f_1,\dots,f_K)=(\nu_1,\dots,\nu_K)$, with $\nu_i=1-\varepsilon$ for $i=\arg\max(f_1,\dots,f_K)$ and $\nu_j=\varepsilon/(K-1)$ for $j\neq i$.
The value of $\varepsilon$ is set to the default value $10^{-3}$ and is kept fixed.
This likelihood is implemented in the GPflow library \cite{matthews2017gpflow}, and in practice it can be substituted by any other one available in GPflow. 
For instance the soft-max likelihood, which generalizes the sigmoid likelihood to multi-class, i.e. $\nu_i=e^{f_i}/\sum_j e^{f_j}$.

Assuming that the underlying real labels for the different instances are independent given the latent variables, we have
\begin{equation}\label{eq:likelihood}
    \p(\bZ|\bF)=\prod_{n=1}^N \p(\bz_n|\bff_{n,:}),
\end{equation}
where $\p(\bz_n|\bff_{n,:})$ is given by eq.~\eqref{eq:likelihood_ind}, and $\bF$ gathers the $K$ latent variables for the $N$ instances. Specifically, $\bF$ is a $N\times K$ matrix, whose $(n,k)$ term is the value of the $k$-th latent variable for the $n$-th instance. As usual, the $n$-th row of $\bF$ is denoted by $\bff_{n,:}$, and the $k$-th column by $\bff_k$.

Finally, independent GP priors are utilized for the latent variables $f_1,\dots,f_K$. This yields the joint prior
\begin{equation}\label{eq:GP_prior}
    \p(\bF|\bTheta,\bX)=
    \prod_{k=1}^K \p(\bff_k|\btheta_k,\bX)=
    \prod_{k=1}^K \mathcal{N}(\bff_k|\mathbf{0},\bK(\btheta_k,\bX)),
\end{equation}
where $\btheta_k$ are the kernel hyperparameters for the $k$-th GP.
In this work we will use the well-known squared exponential kernel, $k(\bx,\by)=\gamma\cdot\exp(-||\bx-\by||^2/(2\sigma^2))$, which has the hyperparameters of variance $\gamma$ and length-scale $\sigma$.
However, as before, the flexibility of GPflow allows us to use any other kernel \cite{matthews2017gpflow}.

In summary, the full probabilistic model is given by
\begin{equation}\label{eq:model_full}
    \p(\bY,\bZ,\bF,\bR|\bTheta)=
    \p(\bY|\bZ,\bR)\p(\bR)\p(\bZ|\bF)\p(\bF|\bTheta),
\end{equation}
with the four factors on the right hand side defined through eqs.~\eqref{eq:annotations}, \eqref{eq:annot_prior}, \eqref{eq:likelihood} and \eqref{eq:GP_prior}, respectively. 
Here, the dependency on the observed features $\bX$ has been omitted for simplicity.

In order to introduce the sparse GP approximation, let us expand this model by introducing $M$ inducing points for each GP. 
Namely, each GP prior $\p(\bff_k)$ can be naively rewritten as the marginal of $\p(\bff_k,\bu_k)$, where $\bu_k=(u_{1k},\dots,u_{Mk})$ are $M$ inducing points. 
These represent the value of the $k$-th GP on $M$ new locations called inducing inputs, $\tilde\bX = \{\tilde\bx_1, \dots, \tilde\bx_M\}$, just like $\bff_k$ does for $\bX$\footnote{Notice that the inducing locations $\tilde\bX$ do not depend on $k$. Although different inducing locations could be used for each GP from a theoretical viewpoint, in practice they are usually considered the same. However, the inducing points $\bu_k$ do depend on $k$, as each GP models a different function.}.
Analogously to $\bF$, we write $\bU$ for the $M\times K$ matrix gathering all the inducing points, whose rows and columns are denoted by $\bu_{m,:}$ and $\bu_k$ respectively. 
Then, if the joint GP $\p(\bff_k,\bu_k)$ is factorized as $\p(\bff_k|\bu_k)\p(\bu_k)$, the model in eq.~\eqref{eq:model_full} can be analogously rewritten as
\begin{align}\label{eq:model_full_U}
    \p(\bY,\bZ,\bF, \bU, \bR|\bTheta)&=
    \p(\bY|\bZ,\bR)\cdot\p(\bR)\cdot\p(\bZ|\bF)\nonumber\\
    &\times \p(\bF|\bU,\bTheta)\cdot\p(\bU|\bTheta).
\end{align}

It is worth stressing that, by marginalizing out $\bU$, this model is equivalent to the one in eq.~\eqref{eq:model_full}. 
This is important because the sparse GP approximations are grouped into two big categories: those which approximate the model and perform exact inference (like FITC \cite{snelson:06}), and those which keep the model unaltered and introduce the approximation at the inference step. Our approach, like SVGP, belongs to the second group, and the approximation is carried out next. 

%% file: inference.tex
Given the model in eq.~\eqref{eq:model_full_U}, an exact solution would involve calculating the marginal likelihood $\p(\bY|\bTheta)$, in order to estimate the optimal kernel hyperparameters $\tilde\bTheta$ and then obtain the posterior $\p(\bZ,\bF, \bU, \bR|\bY,\tilde\bTheta)$.
However, integrating out $\bZ$, $\bF$, $\bU$ and $\bR$ is analytically intractable, and we resort to variational inference to approximate the computations \cite{blei2017variational}.

The core of variational inference is the following decomposition of the log marginal likelihood (associated to the observations), which is straightforward and holds for any distribution $\q(\bz,\bF,\bU,\bR)$\footnote{Observe that, in order to ``lighten'' the notation, we use the integral symbol also for the discrete variable $\bZ$.}:
\begin{align}
&\log\p(\bY|\bTheta) =
\mathrm{KL}(\q(\bZ,\bF,\bU,\bR)||\p(\bZ,\bF,\bU,\bR|\bY,\bTheta))\nonumber\\
&+
\underbrace{
\int\q(\bZ,\bF,\bU,\bR)
\log\frac{\p(\bY,\bZ,\bF,\bU,\bR|\bTheta)}{\q(\bZ,\bF,\bU,\bR)}{\rm d}\bZ{\rm d}\bF{\rm d}\bU{\rm d}\bR.
}_\mathrm{ELBO}\label{eq:VI_decomposition}
\end{align}
This distribution $\q$ must be understood as an approximation to the true posterior $\p(\bZ,\bF,\bU,\bR|\bY,\bTheta)$. 
The second term in the right hand side of eq.~\eqref{eq:VI_decomposition} is called the Evidence Lower Bound (ELBO), since it is a lower bound for the model evidence or log marginal likelihood $\log\p(\bY|\bTheta)$ (recall that the first term, the KL divergence, is always non-negative, and is zero if and only if both distributions coincide). Moreover, notice that this KL divergence is precisely between the approximate posterior $\q$ and the true one. 

The idea of variational inference is to propose a parametric form for $\q$.
Then, the ELBO in eq.~\eqref{eq:VI_decomposition} is maximized with respect to these new variational parameters, the kernel hyperparameters $\bTheta$, and the inducing locations $\tilde\bX$ (which are not usually considered fixed).
Notice that, by maximizing the ELBO, we are at the same time considering the log marginal likelihood $\log\p(\bY|\bTheta)$ and the KL divergence between $\q$ and the real posterior (just solve for the ELBO in eq.~\eqref{eq:VI_decomposition}).
Thus, variational inference converts the problem of posterior distribution approximation into an optimization one \cite{blei2017variational}, \cite[Section 10.1]{Bishop:06}, which in practice is addressed through optimization algorithms such as Adam Optimizer \cite{Adam}.



Here, the following parametric form is proposed for $\q$:
\begin{align}
    \label{eq:parametric_form0}
    &\q(\bZ,\bF,\bU,\bR)
    =\q(\bZ)\q(\bF|\bU,\bTheta)\q(\bU)\q(\bR),\textrm{ with}\\
    \label{eq:parametric_form1}
    &\q(\bZ)
    =\prod_{n=1}^N \q(\bz_n)=\prod_{n=1}^N \bz_n^\intercal \bq_{n},\\
    \label{eq:parametric_form2}
    &\q(\bF|\bU,\bTheta)
    =\p(\bF|\bU,\bTheta),\\
    \label{eq:parametric_form3}
    &\q(\bU)
    =\prod_{k=1}^K \q(\bu_k) = \prod_{k=1}^K \mathcal{N}(\bu_k|\bm_k,\bS_k),\\
    \label{eq:parametric_form4}
    &\q(\bR)
    =\prod_{a=1}^A\prod_{k=1}^K \q(\br^a_k) = \prod_{a=1}^A\prod_{j=1}^K \mathrm{Dir}(\br^a_j|\tilde\alpha_{1j}^a,\dots,\tilde\alpha_{Kj}^a).
\end{align}

The proposed posterior on $\bZ$ factorizes across data points, and each $\bq_n=(q_{n1},\dots,q_{nK})\in[0,1]^K$ describes the probability that $K$ is the real class for $\bx_n$ (i.e., $\sum_k q_{nk} = 1$).
The prior conditional $\bF|\bU$ does not introduce any new variational parameter.
The posterior on $\bU$ factorizes across dimensions, and each one is given by a Gaussian with mean $\bm_k\in \R^M$ and (positive-definite) covariance matrix $\bS_k\in\R^{M\times M}$.
Finally, $\q(\bR)$ factorizes across annotators and dimensions, and they are assigned Dirichlet distributions with parameters $\tilde\alpha_{ij}^a>0$. 
In the sequel, all these variational parameters $\{\bq_n: n=1,\dots, N\}$, $\{\bm_k,\bS_k:k=1,\dots,K\}$, $\{\tilde\alpha_{ij}^a: i,j=1,\dots,K;\ a=1,\dots,A\}$ will be jointly denoted by $\bV$.

In the proposed form described by eqs.~\eqref{eq:parametric_form0}--\eqref{eq:parametric_form4}, the prior conditional $\p(\bF|\bU,\bTheta)$ arises in a natural way if the GP values are assumed conditionally independent on any other value given the inducing points $\bU$.
This is the original assumption of Titsias in \cite{titsias:09}, and intuitively implies that all the information is condensed by and propagated through the inducing points $\bU$. 
This form of $\bF|\bU$, plus that of $\q(\bU)$, are at the core of the sparse GP approximation that we are inspired by, SVGP \cite{Hensman:15}.
The distributions $\q(\bZ)$ and $\q(\bR)$ are given the functional form that would arise if a mean-field approach was applied \cite[Eq. (10.9)]{Bishop:06}.
For that, the conjugacy between the Dirichlet distribution in $\p(\bR)$ and the categorical in $\p(\bY|\bZ,\bR)$ is essential.

Now, we can compute the explicit expression for the ELBO in our case, which must be maximized w.r.t. $\bV$, $\bTheta$, and $\tilde\bX$: 
\begin{align}
    &\mathrm{ELBO}(\bV,\bTheta,\tilde\bX)=\nonumber
    \\ \nonumber
    &\mathbb{E}_{\q(\bZ)\p(\bF|\bU)\q(\bU)\q(\bR)}\!
    \log\frac{\p(\bY|\bZ,\bR)\p(\bZ|\bF)\cancel{\p(\bF|\bU)}\p(\bU)\p(\bR)}{\q(\bZ)\cancel{\p(\bF|\bU)}\q(\bU)\q(\bR)}
    \\ \nonumber
    &=\sum_{n=1}^N\sum_{a\in A_n}\sum_{\by\in\bY_n^a}\sum_{k=1}^K q_{nk}\mathbb{E}_{\q(\br_k^a)}\left[\log\p(\by|\be_k,\br_k^a)\right]
    \\ \nonumber
    &+\sum_{n=1}^N\sum_{k=1}^K q_{nk} \mathbb{E}_{\q(\bff_{n,:})}\left[\log\p(\be_k|\bff_{n,:})\right]-\sum_{n=1}^N\sum_{k=1}^K q_{nk}\log q_{nk}
    \\
    &
    -\sum_{k=1}^K \mathrm{KL}(\q(\bu_k)||\p(\bu_k))
    -\sum_{a=1}^A\sum_{k=1}^K\mathrm{KL}(\q(\br_k^a)||\p(\br_k^a)), \label{eq:ELBO}
\end{align}
A detailed derivation of this expression is provided in the supplemental material.
Notice that the inclusion of the prior conditional $\p(\bF|\bU)$ in the approximate posterior makes the highlighted cancellation possible, which is essential for the scalability of the method.
All these five terms in eq.~\eqref{eq:ELBO} but the second one can be expressed in closed-form as a function of $\bV$, $\bTheta$, and $\tilde\bX$.
Similarly, the second one can be approximated explicitly through Gaussian-Hermite quadrature \cite{olver2010nist}, which is already implemented in GPflow for many different likelihoods (like the robust-max used here) \cite{matthews2017gpflow}.
Further details and the specific expressions can be found in the supplemental material.
As a summary, Table \ref{tab:ELBO_dependence} shows which parameters each term in eq.~\eqref{eq:ELBO} depends on.

{
\renewcommand{\arraystretch}{1.7}
\begin{table}
    \centering
    \caption{Specifying the dependence of the ELBO on the variational parameters, the kernel hyperparameters, and the inducing locations through its five terms in eq.~\eqref{eq:ELBO}.}
    \begin{tabular}{c|c}
        ELBO term & Parameters it depends on \\
        \hline
        $\sum\q_{nk}\mathbb{E}_{\q(\br_k^a)}\left[\log\p(\by|\be_k,\br_k^a)\right]$
        & 
        $q_{nk},\tilde\alpha_{ij}^a$
        \\
        
        $\sum\q_{nk}\mathbb{E}_{\q(\bff_{n,:})}\left[\log\p(\be_k|\bff_{n,:})\right]$
        &
        $q_{nk},\bm_k,\bS_k,\bTheta,\tilde\bX$
        \\
        
        $\sum q_{nk}\log q_{nk}$
        &
        $q_{nk}$
        \\
        $\sum \mathrm{KL}(\q(\bu_k)||\p(\bu_k))$
        &
        $\bm_k,\bS_k,\bTheta,\tilde\bX$
        \\
        $\sum\mathrm{KL}(\q(\br_k^a)||\p(\br_k^a))$
        &
        $\tilde\alpha_{ij}^a$
    \end{tabular}
    \label{tab:ELBO_dependence}
\end{table}
}

Importantly, observe that the expression for the ELBO factorizes across data points, which allows for stochastic optimization through mini-batches \cite{hoffman2013stochastic}. 
To the best of our knowledge, this allows GP-based crowdsourcing methods to scale up to previously prohibitive data sets for the first time.   
More specifically, the computational complexity to evaluate the ELBO in eq.~\eqref{eq:ELBO} in terms of the training set size is $\mathcal{O}(N_b (M^2+A_bK))$, where $N_b$ is the mini-batch size, $M$ the number of inducing points, $K$ the number of classes, and $A_b$ the number of annotations per instance in the mini-batch.
This extends the scalability of the sparse GP approximation SVGP \cite{Hensman:15} to the crowdsourcing setting.
It is also interesting to compare eq.~\eqref{eq:ELBO} with the expression for the ELBO in SVGP \cite[Eq. (19)]{Hensman:15}.
The second and fourth terms, which come from the prior and the classification likelihood, are analogous to the two terms in \cite{Hensman:15}. The other three terms arise naturally from the crowdsourcing modelling.

Once the ELBO is maximized w.r.t. $\bV$, $\bTheta$ and $\tilde\bX$, we can make predictions for previously unseen data points. Given a new $\bx^*$, we have
\begin{multline}
    \p(f^*_k|\bx^*,\mathcal{D})\!=\!
    \int \p(f^*_k|\bu_k)\p(\bu_k|\mathcal{D})\mathrm{d}\bu\approx
    \mathbb{E}_{\q(\bu_k)}\p(f^*_k|\bu_k)
    \\ 
    =\mathcal{N}\left(f_k^*|
    \bB_{\bx^*\tilde\bX}\bm_k,
    k_{\bx^*\bx^*}+\bB_{\bx^*\tilde\bX}(\bS_k-\bK_{\tilde\bX\tilde\bX})\bB_{\tilde\bX\bx^*}\right),
\end{multline}
where $\bB_{\bx^*\tilde\bX}$ stands for $\bK_{\bx^*\tilde\bX}\bK_{\tilde\bX\tilde\bX}^{-1}$, and we are using the values of $\bm_k$, $\bS_k$, $\bTheta$, and $\tilde\bX$ estimated after training. 
The predictive distribution on the real label $\bz^*$ is obtained as $\p(\bz^*)=\int\p(\bz^*|\bff^*)\p(\bff^*)\mathrm{d}\bff^*$. For classification likelihoods like ours, this is computed by GPflow through Gaussian-Hermite quadrature.
Moreover, as we will illustrate in the experiments, the posterior distributions $\q(\bZ)$ and $\q(\bR)$ provide an estimation for the underlying real label of the training points and for the annotators degree of expertise, respectively.
Finally, in order to exploit GPU acceleration through TensorFlow, the novel SVGPCR is implemented within the popular GP framework GPflow \cite{matthews2017gpflow}.
The code will be made publicly available in GitHub upon acceptance of the paper, and will be listed in the ``projects using GPflow'' section of the GPflow site \url{https://github.com/GPflow/GPflow}.  



%% file: data.tex
The Laser Interferometer Gravitational-Waves Observatory (LIGO) is a large-scale physics experiment and observatory to detect gravitational waves (GWs) \cite{Abramovici325}.
These are ripples in the space-time produced by non-symmetric movements of masses, being their energy much higher for events such as binary black holes or neutron stars mergers. 
Their existence is a direct consequence of the General Relativity theory postulated in 1916.
However, Albert Einstein himself believed they would be extremely difficult to detect by any technology foreseen at that time \cite{kennefick2016traveling}.

The first direct observation of GWs was made one hundred years later by LIGO, on September 14th, 2015.
The discovery had a tremendous impact in the scientific community. Not only as an empirical validation of one of the most recognized Physics theories, but also as a whole new way to explore the universe.
So far, astrophysicists could perceive the outer space only through one ``sense'' (electromagnetic radiation), but were ``deaf'' to GWs.
This detection has inaugurated a new era of the so-called GWs astronomy, and has been awarded the 2017 Physics Nobel Prize \cite{gwDiscovery}.

\begin{figure*}
    \centering
    \begin{tabular}{ccc}
    1080LINE & 1400RIPPLE & BLIP \\
    \includegraphics[width=0.3\textwidth]{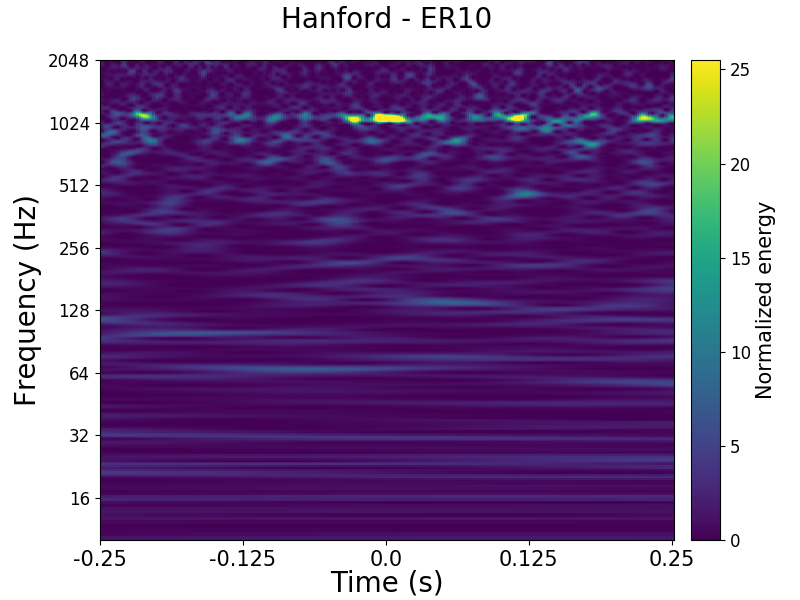} &
    \includegraphics[width=0.3\textwidth]{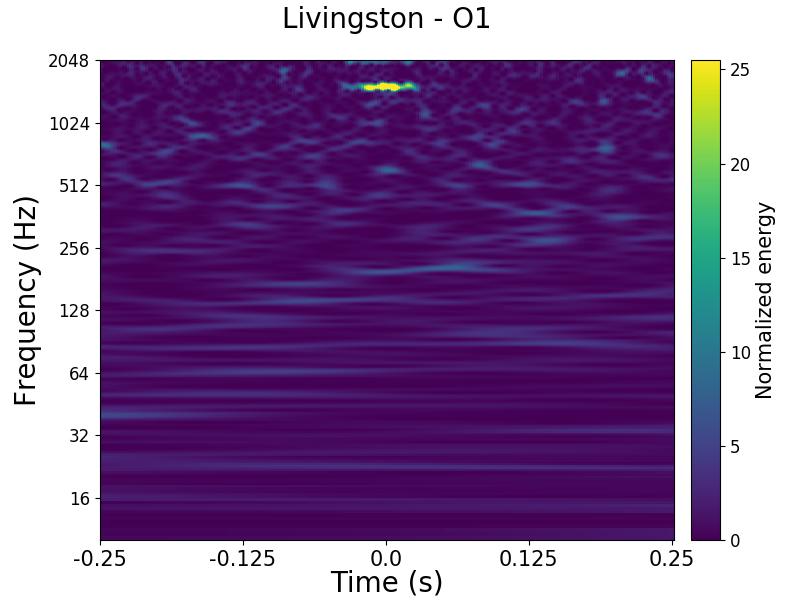} &
    \includegraphics[width=0.3\textwidth]{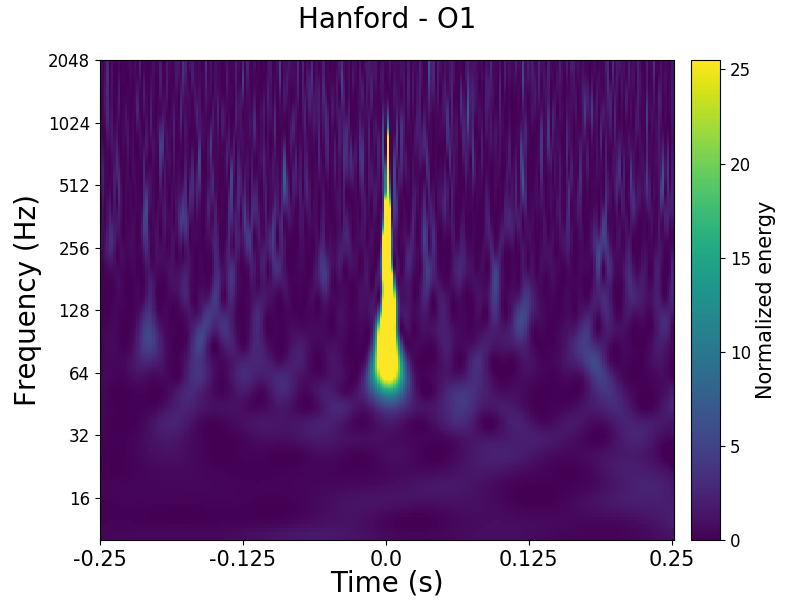} \\
    EXTREMELY LOUD & KOIFISH & LOW FREQUENCY BURST \\
    \includegraphics[width=0.3\textwidth]{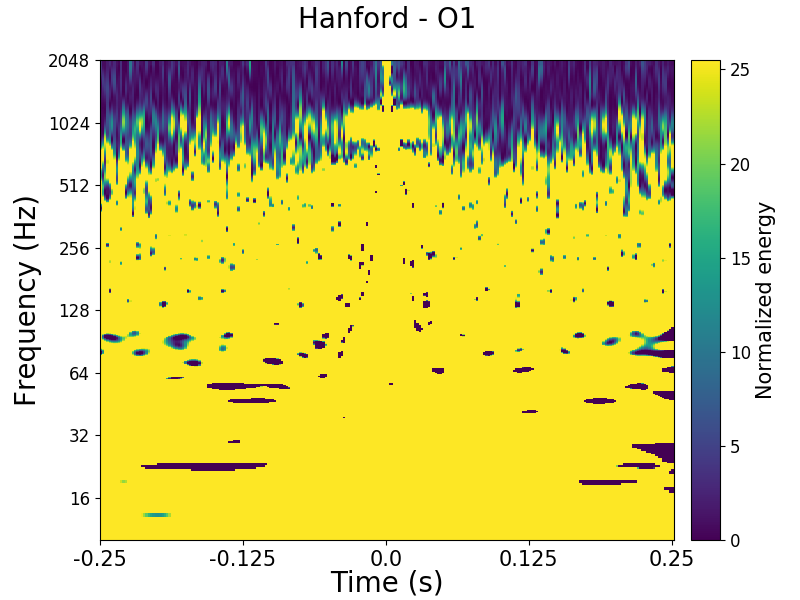} &
    \includegraphics[width=0.3\textwidth]{figures/data/KOIFISH.png} &
    \includegraphics[width=0.3\textwidth]{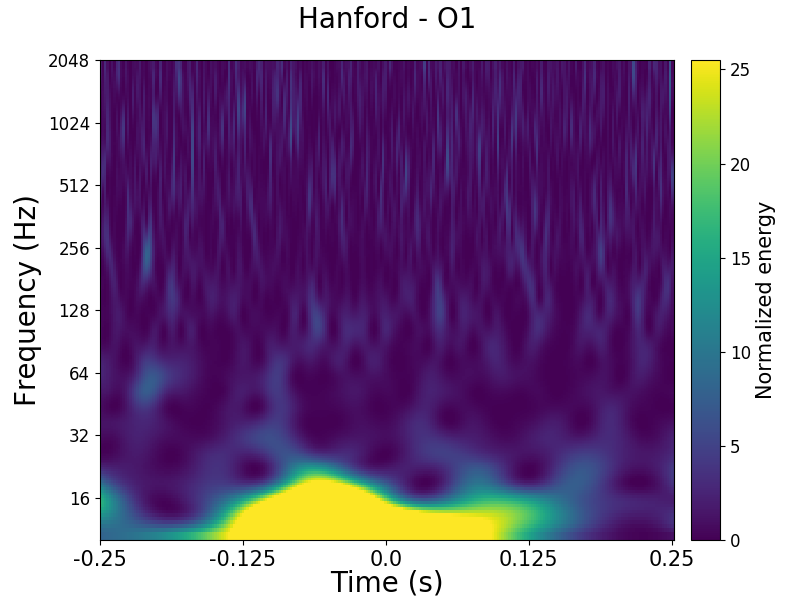} \\
    LOW FREQUENCY LINES & NO GLITCH & OTHER \\
    \includegraphics[width=0.3\textwidth]{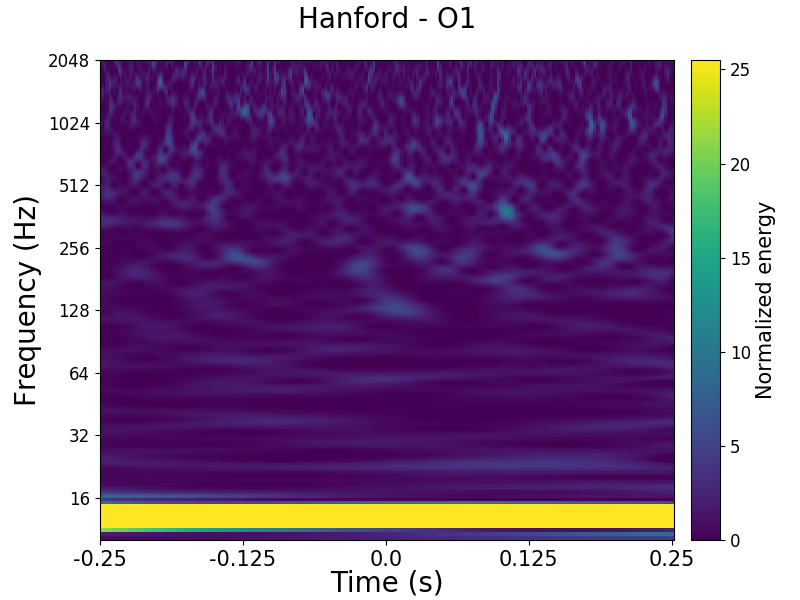} &
    \includegraphics[width=0.3\textwidth]{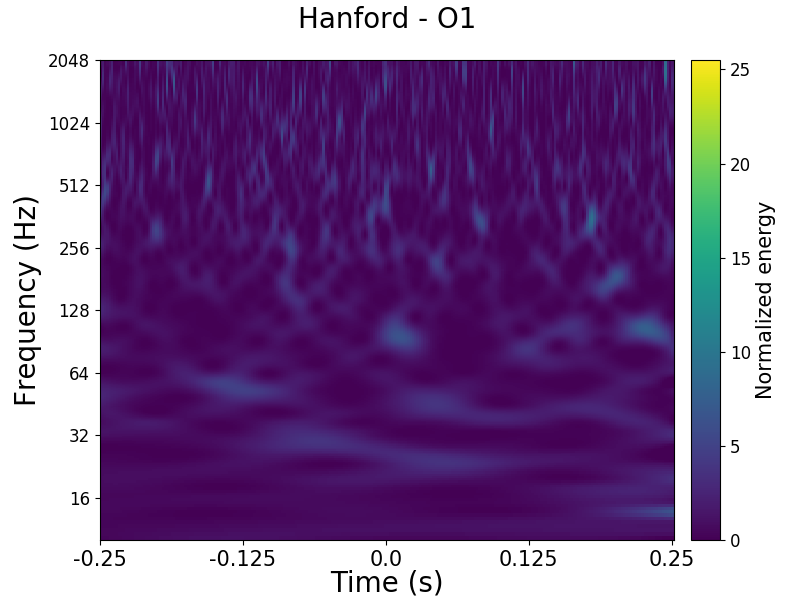} &
    \includegraphics[width=0.3\textwidth]{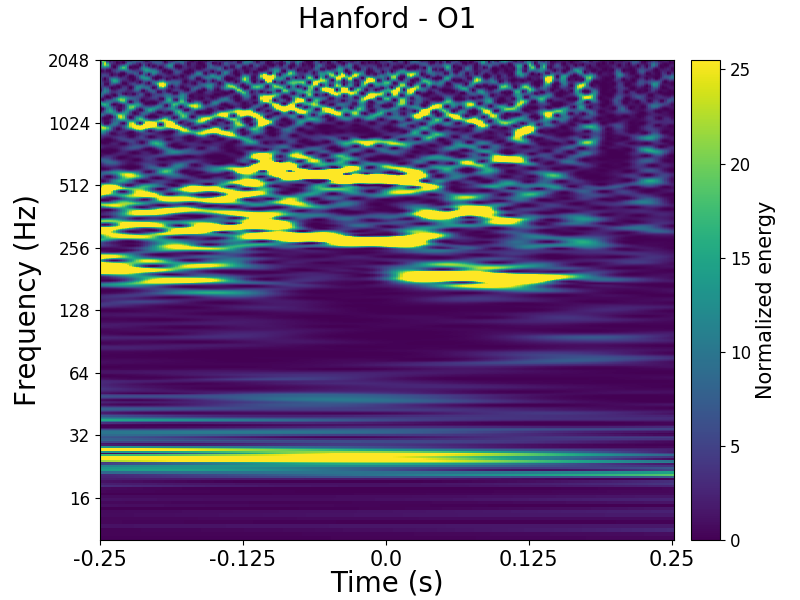} \\
    POWER-LINE 60HZ & REPEATING BLIPS & SCATTERED LIGHT \\
    \includegraphics[width=0.3\textwidth]{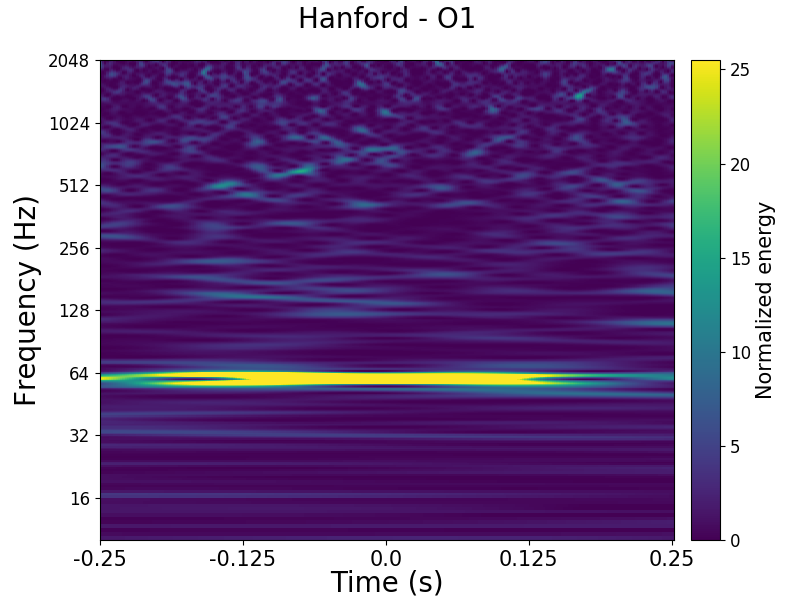} &
    \includegraphics[width=0.3\textwidth]{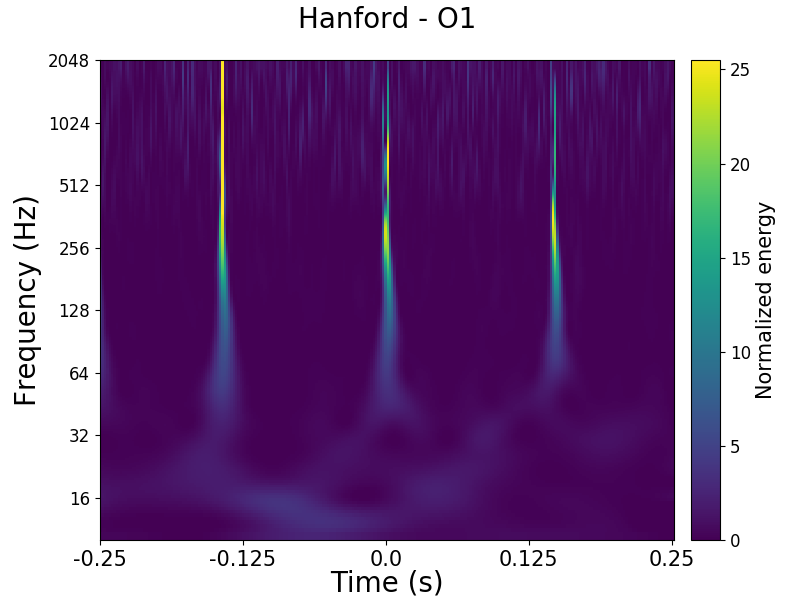} &
    \includegraphics[width=0.3\textwidth]{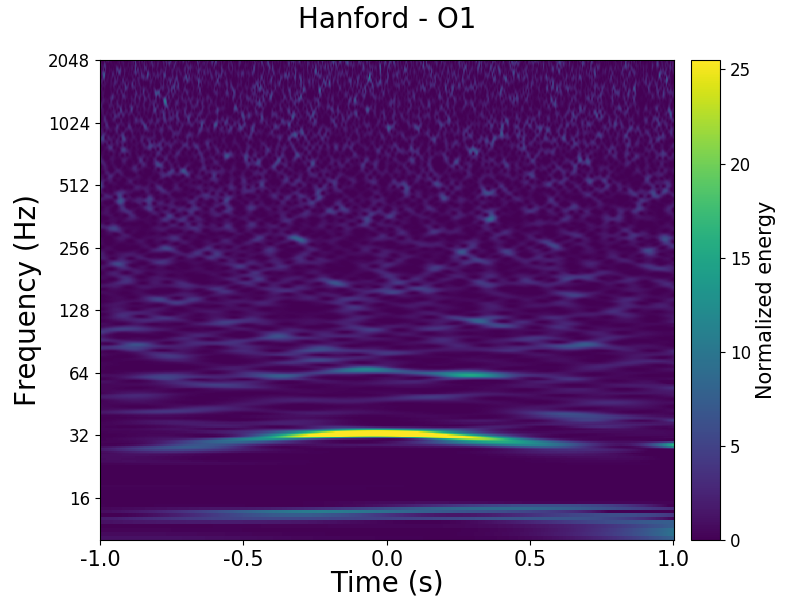} \\
    SCRATCHY & VIOLIN MODE HARMONIC & WHISTLE \\
    \includegraphics[width=0.3\textwidth]{figures/data/SCRATCHY.png} &
    \includegraphics[width=0.3\textwidth]{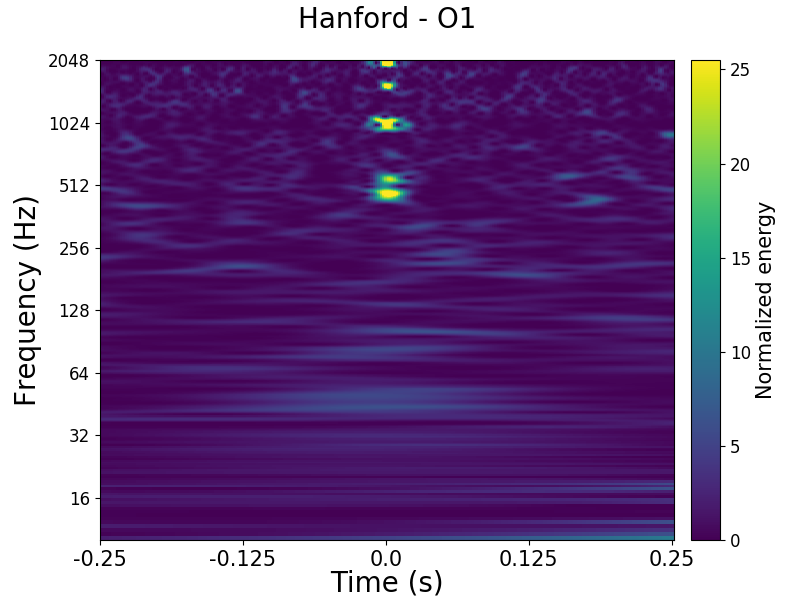} &
    \includegraphics[width=0.3\textwidth]{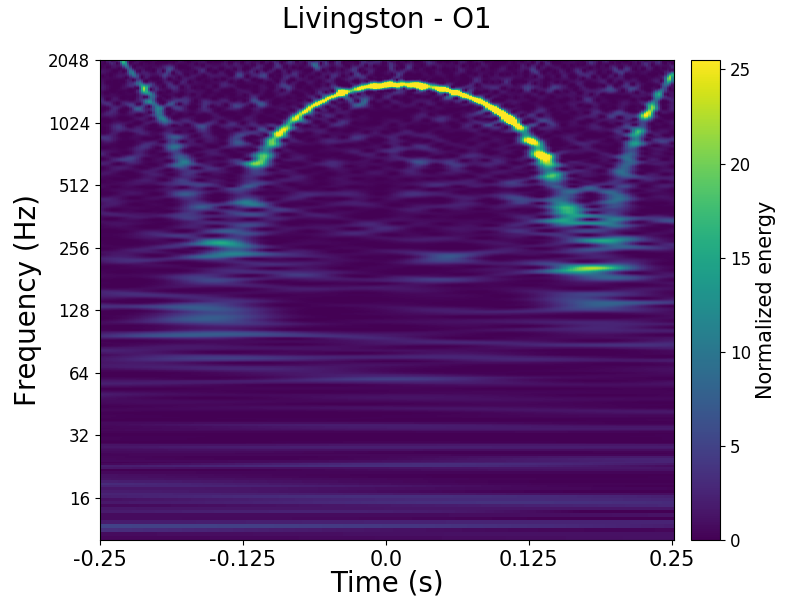}
    \end{tabular}
    \caption{Representative spectrograms for the 15 different types of glitches considered in this work. Hanford and Livinsgton refer to the two observatories that LIGO comprises, and ER10/O1 to two different observation runs. A brief description of each glitch is provided in the text. The goal of the GravitySpy project is to learn a machine learning system to automatically classify these glitches. The labels for the training set are obtained through crowdsourcing.}
    \label{fig:glitches_full}
\end{figure*}

To identify GWs, LIGO is able to detect changes of the length of a 4 kilometers arm by a thousandth of the width of a proton \cite{Abramovici325}.
This is proportionally equivalent to changing the distance to the nearest star outside the Solar System by one hair's width.
Such precision requires cutting-edge technology that is also extremely sensitive to different instrumental and environmental sources of noise. 
In the spectrograms that astrophysicists analyze to search for GWs, this contamination manifests itself in the form of glitches, which are noisy patterns that adopt many different morphologies \cite{LIGO:17}. 
The presence of these glitches hinders the detection of true GWs.
Figure \ref{fig:glitches_full} shows the 15 types of glitches considered in this work, which will be later described.

The goal of the GravitySpy project is to develop a system to accurately classify the different types of glitches \cite{LIGO:17}. 
This would help astrophysicists to gain insights on their taxonomy and potential causes, enhancing detection of true GWs.
Since LIGO produces a constant stream of data, GravitySpy leverages crowdsourcing techniques through the Zooniverse platform in order to label a training set \url{https://www.zooniverse.org/projects/zooniverse/gravity-spy}.
Then, machine learning crowdsourcing algorithms that can learn from this multiple-annotated data must be applied (like the SVGPCR presented here).

Our training set contains 173565 instances (glitches) and 1828981 annotations (i.e., a mean value of more than 10 labels per instance), which have been provided by 3443 collaborators through the Zooniverse platform. 
These instances are time-frequency plots (spectrograms) like those in figures \ref{fig:glitches_intro} and \ref{fig:glitches_full}, taken with four time windows.
For each one, we will use 256 relevant features extracted in \cite{GravitySpy_NN}. 
These glitches have been classified into 15 different classes proposed by astrophysicists (recall figure \ref{fig:glitches_full}).
Next, we provide a brief description of them (see \cite{GravitySpy_NN} for a more detailed explanation). 

\textbf{1080 Line:}
It appears as a string of short yellow dots, always around 1080Hz.
It was reduced after an update on 2017, although it is still present.

\textbf{1400 Ripple:}
Glitches of 0.05s or longer around 1400Hz. So far, their origin is unknown. They are commonly confused with 1080Line and Violin Mode Harmonic.

\textbf{Blip:}
Short glitches with a symmetric ``teardrop'' shape in time-frequency.
Blips are extremely important since they hamper the detection of binary black hole mergers \cite{abbott2016characterization}.

\textbf{Extremely Loud:}
These are caused by major disturbances, such as an actuator reaching the end of its range and ``railing'', or a photodiode ``saturating''.
They look very bright, due to their very high energy.

\textbf{Koifish:}
Similar to Blips, but resemble a fish with the head at the low frequency end, pectoral fins around 30 Hz, and a thin tail around 500Hz. LIGO scientists do not understand the physical origin of this glitch.

\textbf{Low Frequency Burst:}
Resembles a hump with a nearly triangular shape growing from low frequency to a peak, and then dying back down in one or two seconds.
It is caused by scattered light driven by motion of the output mirrors.

\textbf{Low Frequency Lines:}
These appear as horizontal lines at low frequencies. Can be confused with Scattered Light (the latter shows some curvature) and Low Frequency Bursts (the former continues to look like a line in the 4s window).
    
\textbf{No Glitch:}
No glitch refers to images that do not have any glitch visible at all. The spectrograms would appear dark blue with only small fluctuations.

\textbf{Other:}    
This category is a catch-all for glitches that do not fit into the other categories. Therefore, it presents a great variability in its morphology.
    
\textbf{Power-line 60Hz:}
In US, the mains power is alternating current at 60Hz.
When equipment running on this power switches on or off, glitches can occur at 60Hz or harmonics (120, 180...).
These glitches usually look narrow in frequency, centered around 60Hz or harmonics.
    
\textbf{Repeating blips:}
Analogous to blips, but repeat at regular intervals, usually every 0.125, 0.25 or 0.5 seconds.

\textbf{Scattered Light:}    
After hitting optical components, some light from LIGO beam is scattered. It may then reflect off of other objects and re-enter the beam with a different phase.
It usually looks like upward humps, with frequency below 30 Hz.
It hinders searches of binary neutron stars, neutron star black hole binaries, and binary black holes.

\textbf{Scratchy:}
Wide band of mid-frequency signals that looks like a ripply path through time-frequency space.
This glitch hampers searches for binary black hole mergers.

\textbf{Violin Mode Harmonic:}
Test masses in LIGO are suspended from fibers with resonances. These are called violin modes, as they resemble violin strings resonances. Thermal excitations of the fibers produce movements at the violin mode frequencies, centered around 500Hz.
Thus, these glitches are short and located around 500 Hz and harmonics.

\textbf{Whistle:}
Usually appear with a characteristic W or V shape.
Caused by radio frequency signals beating with the LIGO Voltage Controlled Oscillators.
Whistles mainly contaminate searches for binary black hole mergers \cite{nuttall2015improving}.

For testing purposes, the astrophysicists at GravitySpy have labelled a set of $9997$ instances, including glitches from all the $15$ types explained above. 

%% file: experiments.tex
In this section, the proposed SVGPCR is empirically validated and compared against current methods, with a special focus on the LIGO data introduced in the previous section.
Three blocks of experiments are presented in sections \ref{sec:exp_MNIST}, \ref{sec:exp_LIGO_bin} and \ref{sec:exp_LIGO_full}. 
Firstly, the behavior of SVGPCR is thoroughly analyzed in a controlled crowdsourcing experiment based on the popular MNIST set.
Secondly, SVGPCR is compared with previous probabilistic (mainly GP-based) approaches on the LIGO data.
Since most of these methods were proposed for binary problems, we consider a binary task relevant to the GravitySpy project. 
Thirdly, SVGPCR is compared against state-of-the-art DL-based crowdsourcing methods in the full LIGO data set.

\subsection{Understanding the proposed method}\label{sec:exp_MNIST}
Before comparing against other crowdsourcing methodologies, let us analyze the behavior and main properties of the novel SVGPCR.
To do so, we simulate five different crowdsourcing annotators for the well-known MNIST data set.
The availability of simulated annotators and real training labels on this graphic data set constitutes a controlled setting that allows for a comprehensive analysis.

\begin{figure*}
    \footnotesize
    \centering
    \begin{tabular}{cccccc}
    & \;\;\;Annotator 1 & \;\;\;Annotator 2 & \;\;\;Annotator 3 & \;\;\;Annotator 4 & Annotator 5  \\
    \rotatebox{90}{\;\;\;\;\;\;\;\;\;\;\;\; True} &
    \includegraphics[height=2.7cm]{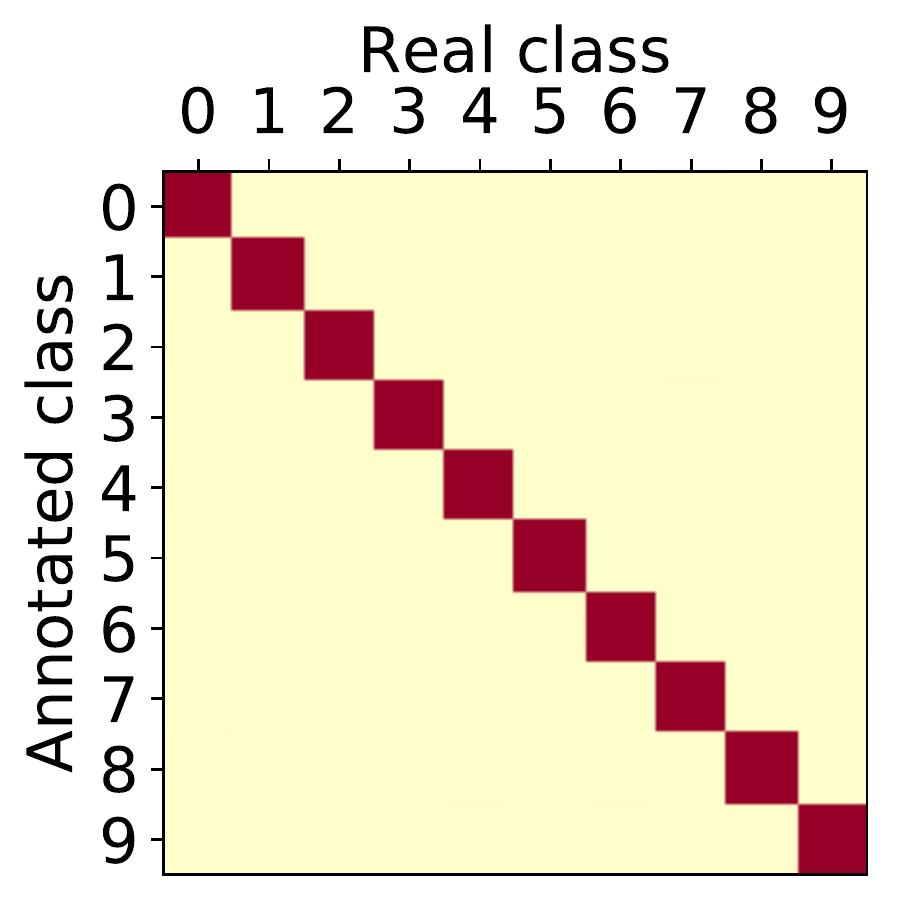}  & \includegraphics[height=2.7cm]{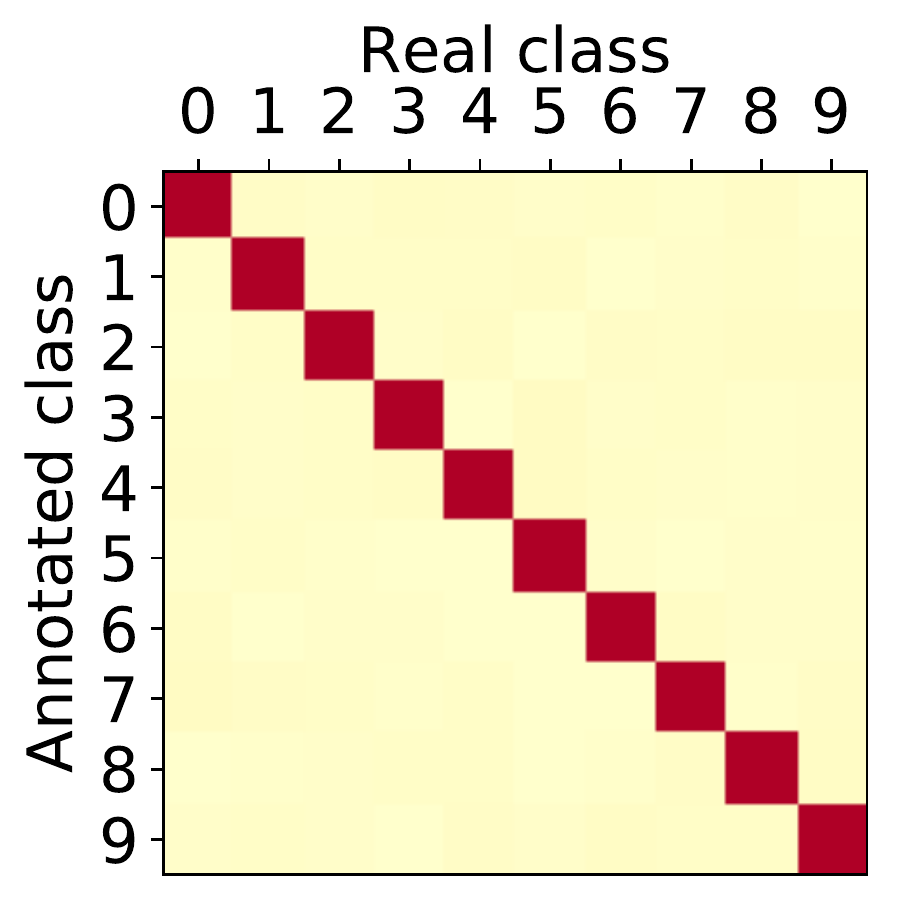} &
    \includegraphics[height=2.7cm]{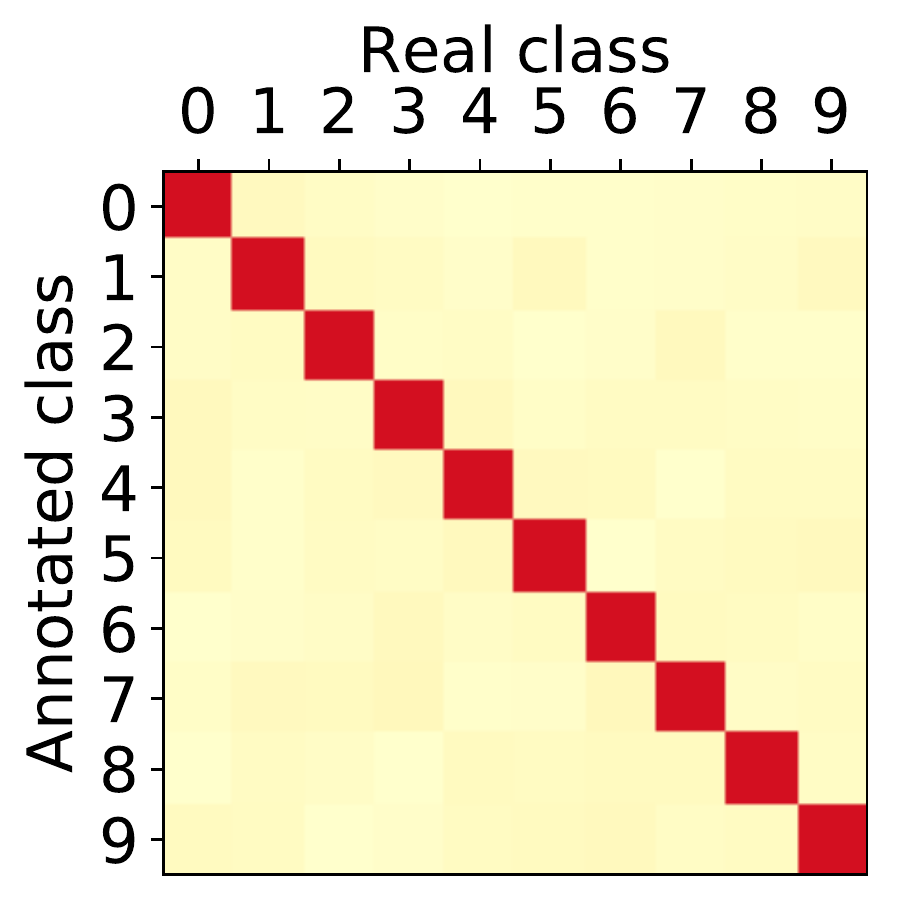}  & \includegraphics[height=2.7cm]{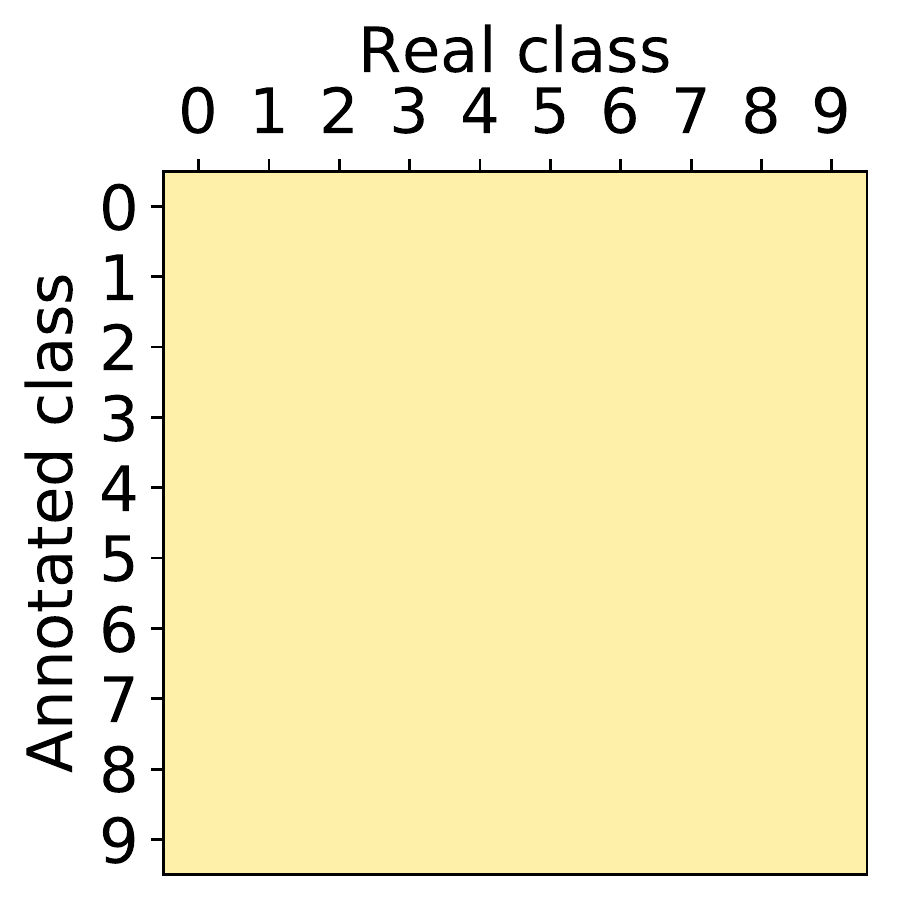} &
    \includegraphics[height=2.7cm]{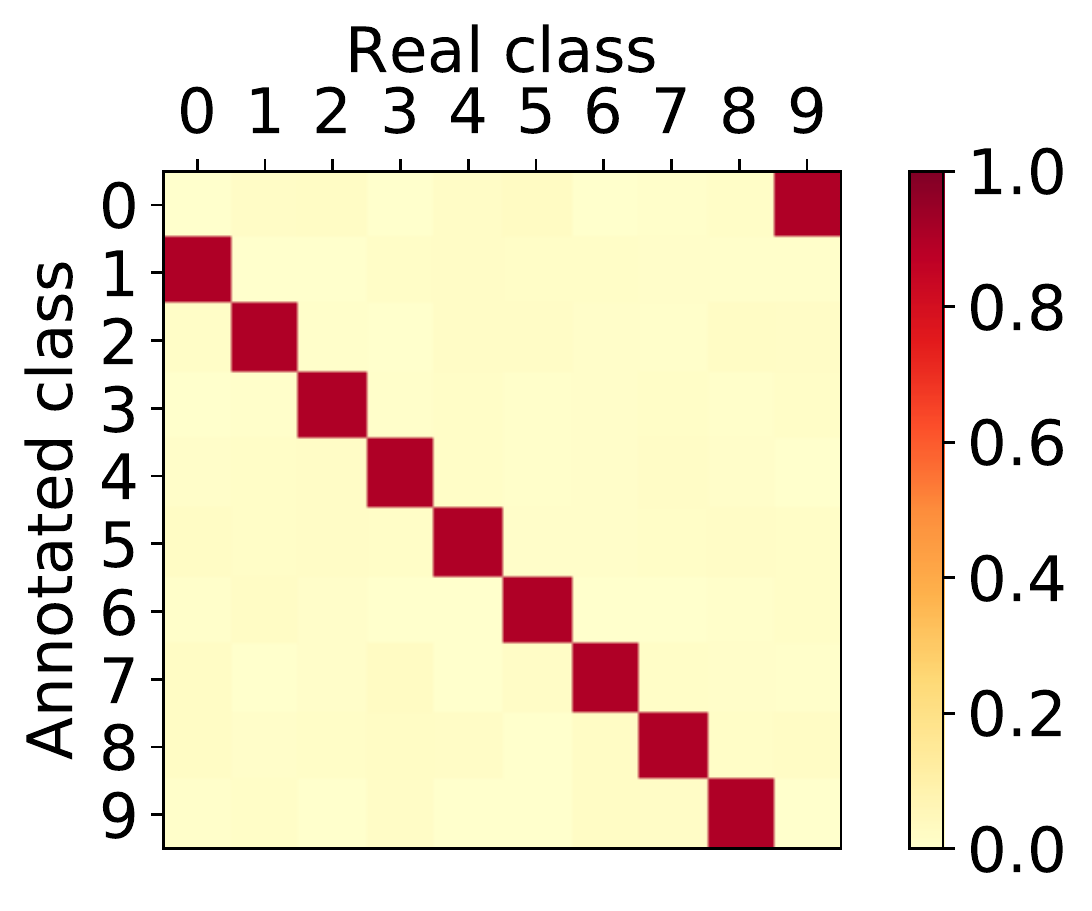} \\
    \rotatebox{90}{\;\;\;\;\;\;\;\;\;\;\;Estimated} &
    \includegraphics[height=2.7cm]{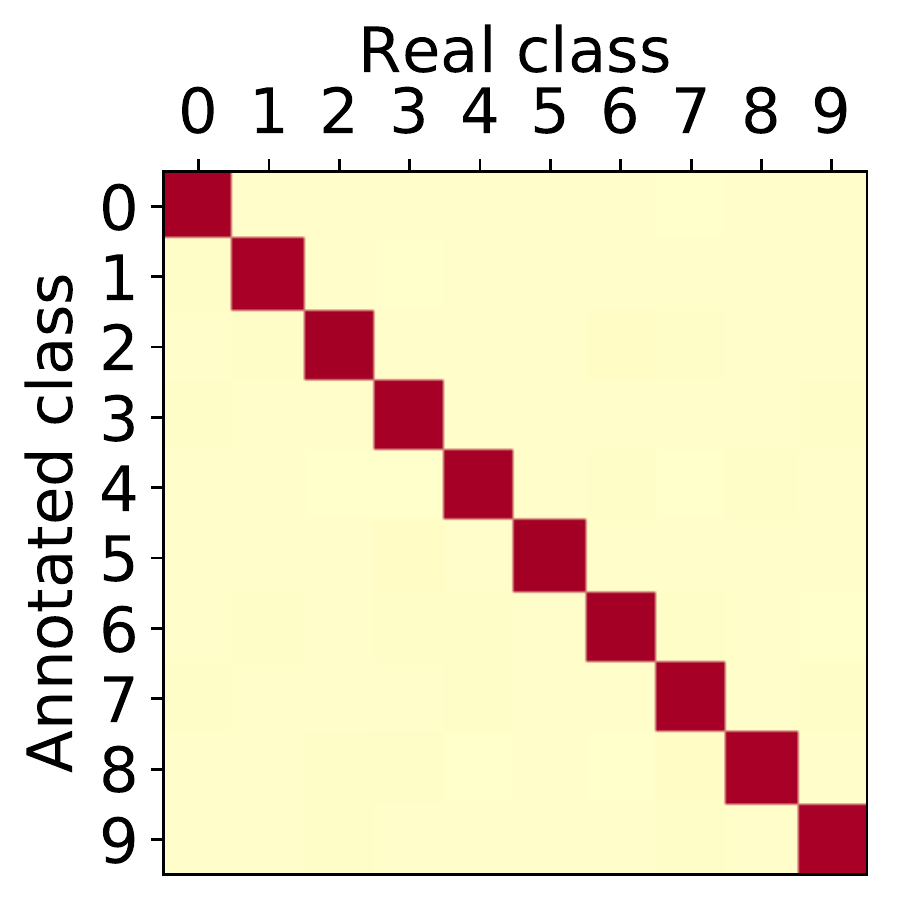} & \includegraphics[height=2.7cm]{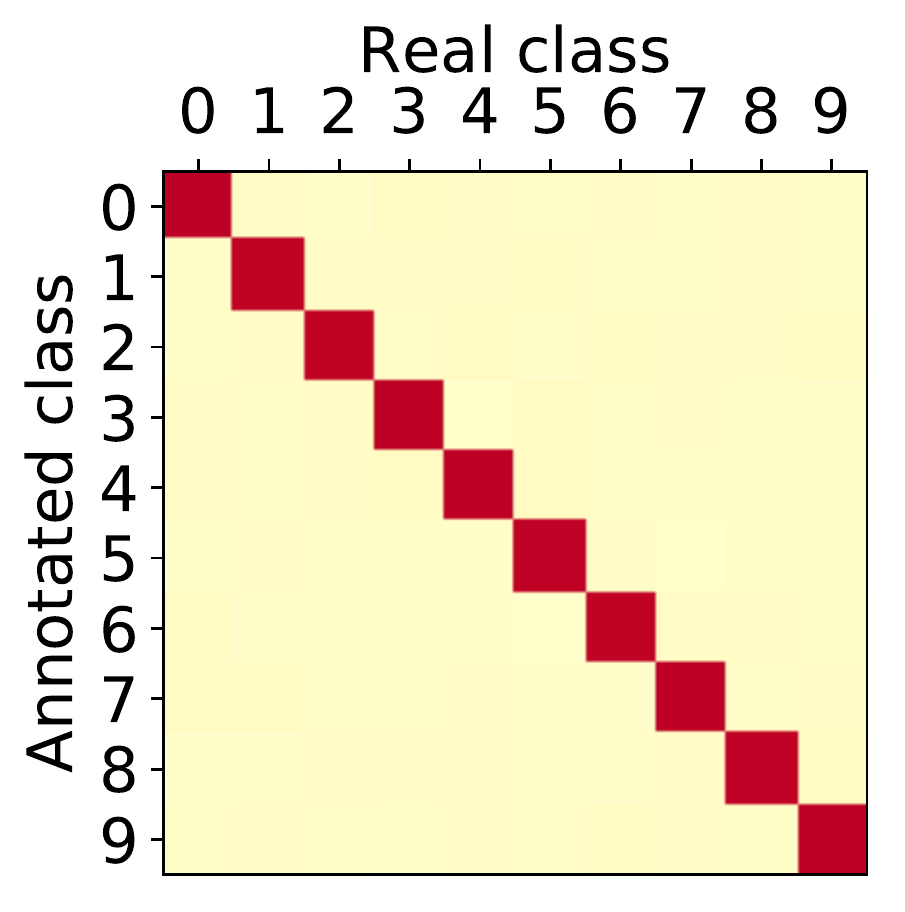} &
    \includegraphics[height=2.7cm]{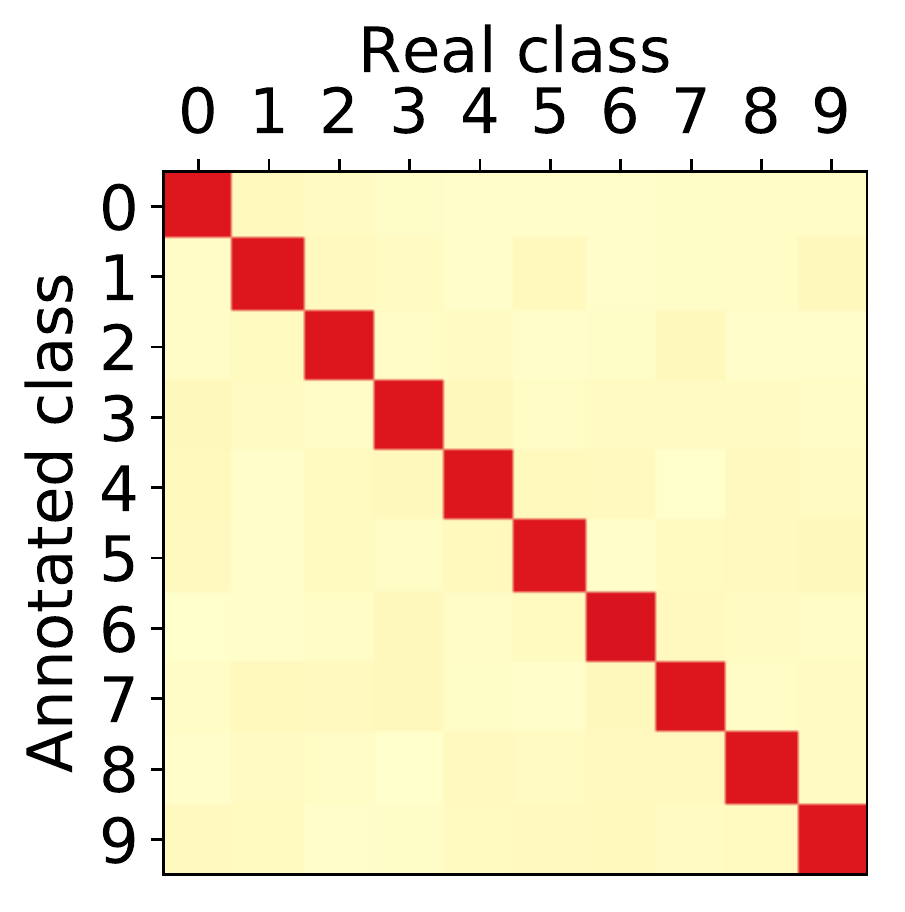}  & \includegraphics[height=2.7cm]{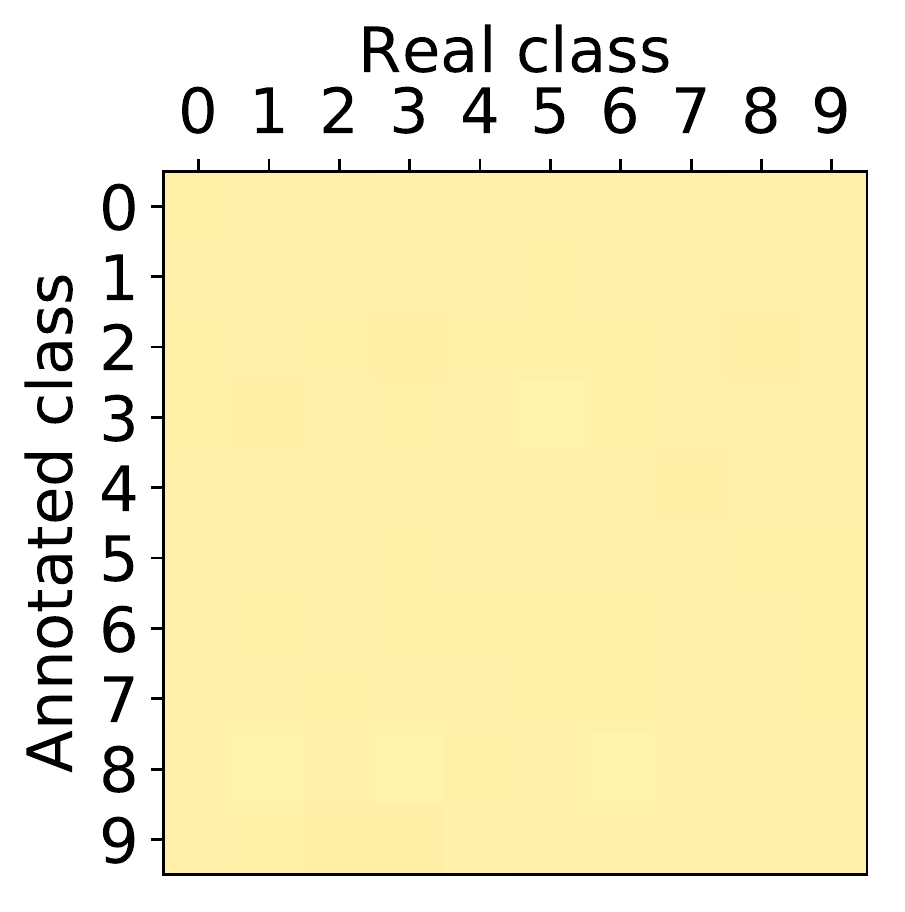} &
    \includegraphics[height=2.7cm]{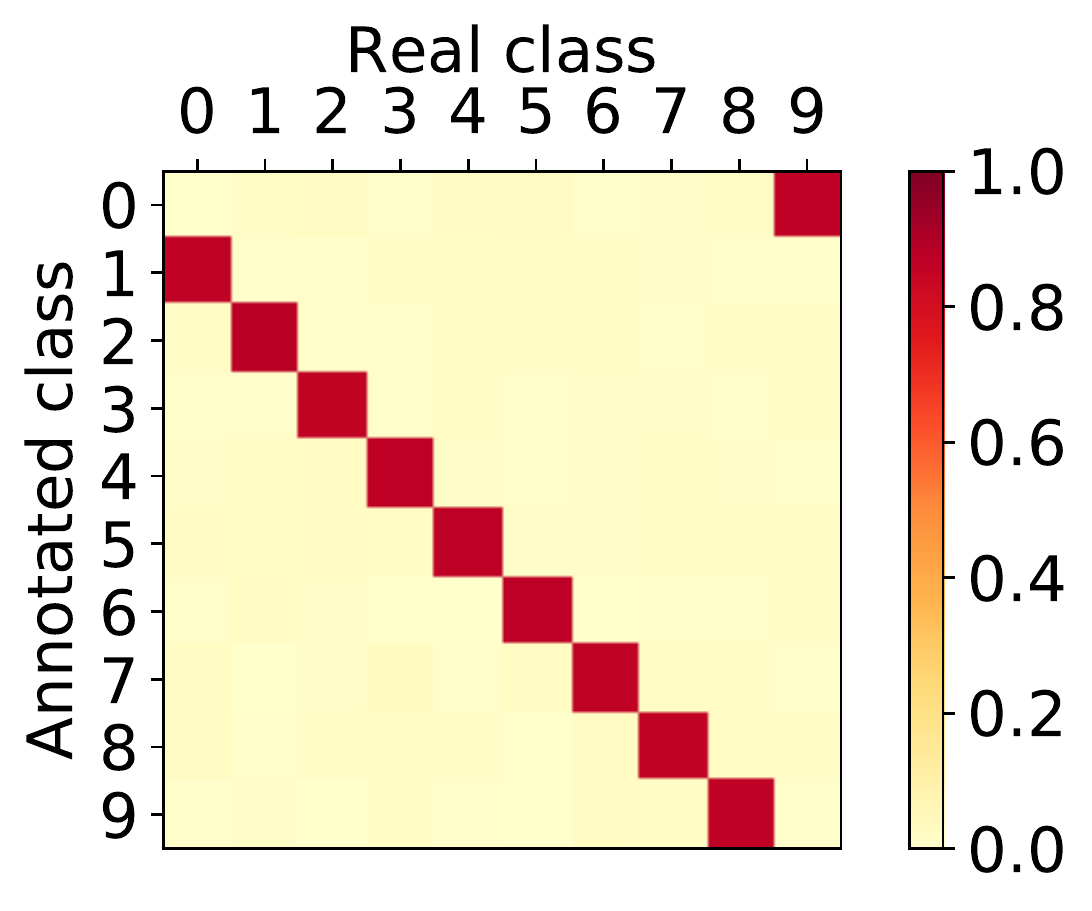} 
    \end{tabular}
    \caption{Estimation of the expertise degree of the simulated annotators in the MNIST problem by SVGPCR. Upper row: true confusion matrices. Lower row: mean of the estimated distribution for the confusion matrices. Notice that the proposed method perfectly identifies adversarial (fifth) and spammer (fourth) annotators.
    Moreover, not only the structure of the matrices is well identified, but also the actual values (the intensity of color is very similar). 
    }
    \label{fig:MNIST_annotators}
\end{figure*}

We use the standard train/test split of MNIST with 60K/10K hand-written digits from 0 to 9 (multi-class problem with 10 classes) \cite{lecun1998gradient}.
Notice that 60K training instances is already prohibitive for standard GPs.
Five decreasingly reliable annotators are considered. The first one has a $95\%$ accuracy for each class, that is, $r^1_{jj}=0.95$ for $j=0,\dots,9$ (the rest of values, $r^1_{ij}$, $i\neq j$, are randomly assigned to add the remaining $0.05$ probability by columns). The second and third ones are defined analogously, but with $90\%$ and $80\%$ accuracy, respectively. 
The fourth one is a \emph{spammer} annotator, that is, $r^4_{ij}=0.1$ for all $i,j=0,\dots,9$. This implies that, regardless of the real class, this annotator assigns a random label.
The fifth one is an \emph{adversarial} annotator. Specifically, in this case, with a $90\%$ of probability, the annotator labels as ($i+1$)-th class an instance whose real class is the $i$th (samples in class 9 are assigned to class 0).
The confusion matrices for these annotators are depicted in the first row of figure \ref{fig:MNIST_annotators} (the rest of the figure will be explained later).
The five annotators label all the instances, which yields $300$K annotations that are used to train SVGPCR.

Since we have available the true labels for the training instances, let us start by comparing SVGPCR with its theoretical upper bound, namely SVGP trained with the true labels, which we refer to as SVGP-gold.
Indeed, since SVGPCR can be considered an extension of SVGP to the noisy scenario of crowdsourcing (recall section \ref{sec:theory}), the performance of the former can be thought to be bounded by that of the latter.
Table \ref{tab:MNIST_gold} shows the global and per-class test accuracy and mean test likelihood for both approaches. 
Importantly, notice that the results are very similar for all classes and both metrics, and SVGPCR almost reaches the same global performance as SVGP-gold (in spite of the corrupted labels provided by annotators).
{
\renewcommand{\arraystretch}{1.2}
\begin{table}
\centering
\caption{Per-class and global test performance of SVGPCR and SVGP-gold in the MNIST problem. In spite of the corrupted labels, the proposed method almost recovers the golden results. 
}
\label{tab:MNIST_gold}
\begin{tabular}{r|cc|cc}
                            & \multicolumn{2}{c|}{Test accuracy} & \multicolumn{2}{c}{Test likelihood} \\ \cline{2-5} 
                            & SVGP-gold        & SVGPCR          & SVGP-gold        & SVGPCR           \\ \hline
0                           & 0.9867           & \textbf{0.9898} & 0.9777           & \textbf{0.9781}  \\
1                           & \textbf{0.9885}  & 0.9877          & \textbf{0.9853}  & 0.9845           \\
2                           & 0.9525           & \textbf{0.9535} & \textbf{0.9368}  & 0.9345           \\
3                           & 0.9703           & \textbf{0.9733} & 0.9475           & \textbf{0.9555}  \\
4                           & 0.9715           & \textbf{0.9735} & 0.9548           & \textbf{0.9570}  \\
5                           & \textbf{0.9630}  & 0.9540          & \textbf{0.9397}  & 0.9352           \\
6                           & \textbf{0.9749}  & \textbf{0.9749} & \textbf{0.9604}  & 0.9585           \\
7                           & \textbf{0.9591}  & 0.9543          & 0.9418           & \textbf{0.9432}  \\
8                           & \textbf{0.9620}  & \textbf{0.9620} & \textbf{0.9458}  & 0.9445           \\
9                           & \textbf{0.9346}  & 0.9316          & \textbf{0.9217}  & 0.9186           \\ \hline
\multicolumn{1}{l|}{Global} & \textbf{0.9665}  & 0.9657          & \textbf{0.9515}  & 0.9514          
\end{tabular}
\end{table}
}

This excellent performance of SVGPCR can be explained by its accurate prediction of the annotators behavior, which in turn allows SVGPCR to properly reconstruct the underlying true labels from the noisy annotations.
Indeed, firstly, figure \ref{fig:MNIST_annotators} shows the exceptional estimations obtained by SVGPCR for the annotators confusion matrices. 
Recall from eq.~\eqref{eq:parametric_form4} that the expertise degree of annotators is estimated through posterior Dirichlet distributions.
The bottom row of figure \ref{fig:MNIST_annotators} shows the mean of those distributions.
Interestingly, the maximum variance was $0.0016$, which implies a high degree of certainty about the predictions in figure \ref{fig:MNIST_annotators}.
Secondly, as previously mentioned, this allows SVGPCR to correctly puzzle out the underlying true labels from the noisy annotations. 
In fact, table \ref{MNIST_trueLabels} shows the excellent per-class and global performance of SVGPCR in this sense (recall that SVGPCR estimates the underlying true labels through the approximate posterior $\q(\bz)$ in eq.~\eqref{eq:parametric_form1}).

{
\renewcommand{\arraystretch}{1.2}
\begin{table*}
\centering
\caption{Per-class and global performance of SVGPCR to reconstruct the underlying true label for training instances in the MNIST problem. An excellent result is obtained across all the classes, with only 20 (out of the 60000 training examples) not correctly predicted.}
\label{MNIST_trueLabels}
\begin{tabular}{r|cccccccccc|c}
           & 0      & 1      & 2      & 3      & 4      & 5      & 6      & 7      & 8      & 9      & Global \\ \hline
Accuracy   & 0.9998 & 0.9997 & 1.0000 & 0.9995 & 0.9995 & 0.9993 & 1.0000 & 0.9997 & 0.9997 & 0.9995 & 0.9997 \\
Likelihood & 0.9997 & 0.9996 & 0.9997 & 0.9992 & 0.9994 & 0.9993 & 0.9999 & 0.9995 & 0.9996 & 0.9991 & 0.9995
\end{tabular}
\end{table*}
}



\begin{figure}
    \centering
    \includegraphics[width=\columnwidth]{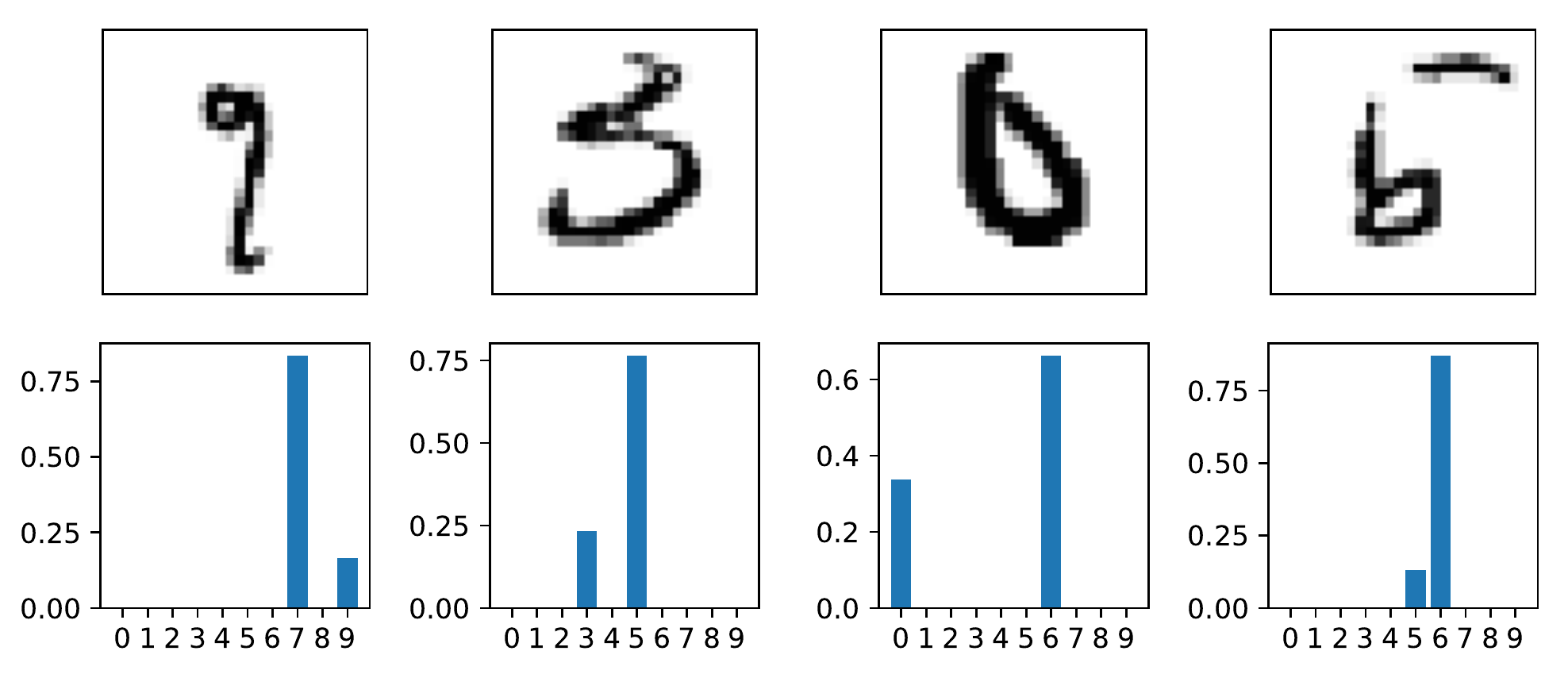}
    \caption{Upper row: four (out of the 20) examples for which SVGPCR is not able to reconstruct the real underlying class in the MNIST problem. Lower row: the corresponding probabilities assigned by SVGPCR. In all cases, the proposed method assigns the second highest probability to the real class. Notice that some of these examples are not easy, and have some features which might lead to confusion.}
    \label{fig:MNIST_SVGPCRfails}
\end{figure}

More in depth, we have analyzed the 20 examples where SVGPCR fails to reconstruct the true label, and some of them can be certainly considered as not-easy ones.
Figure \ref{fig:MNIST_SVGPCRfails} shows four of them, along with the probabilities assigned by SVGPCR for each one. 
In all cases, the true label is assigned the second highest probability by SVGPCR, and the digit presents some feature which certainly leads to confusion with the class that SVGPCR assigns more probability to. 

\begin{figure}
    \centering
    \includegraphics[width=\columnwidth]{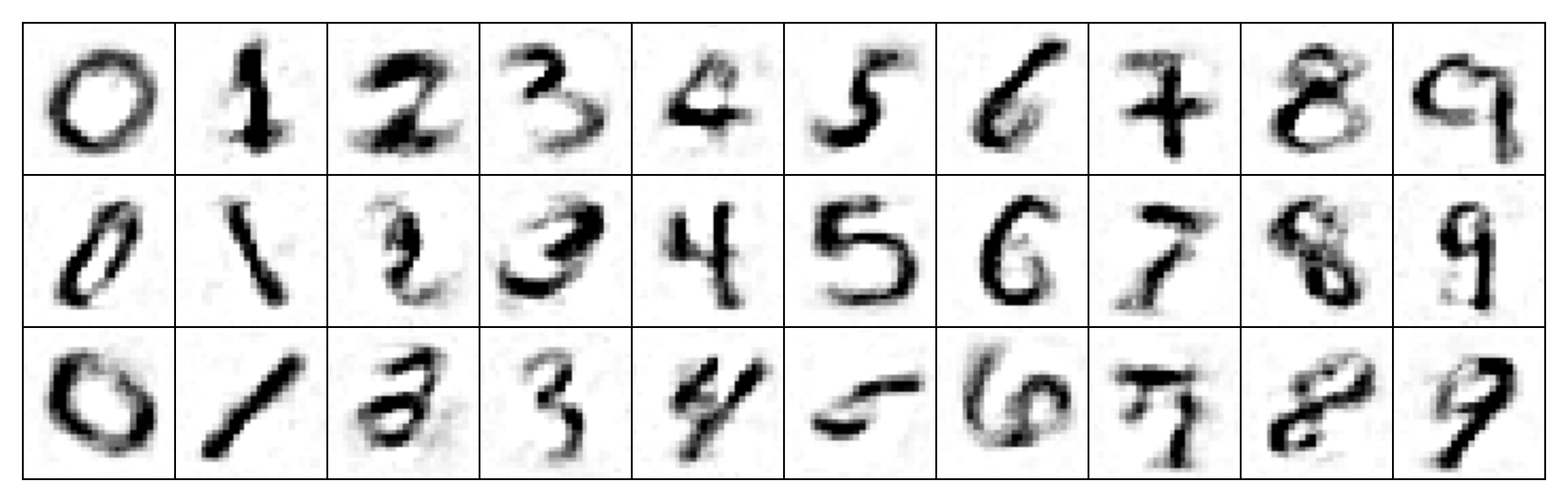}
    \caption{Some of the inducing point locations learned by SVGPCR in the MNIST problem. 
    They have been arranged by columns based on their classification.
    Notice that, for each digit (column), different representative patterns are learned in terms of shape, orientation and thickness.}
    \label{fig:MNIST_ipLocations}
\end{figure}

Another key aspect of the novel SVGPCR is the role of the inducing points.
In this example we are using $M=100$, and the next experiment will be devoted to analyze the influence of $M$ in the performance of SVGPCR.
But before, figure \ref{fig:MNIST_ipLocations} shows the locations to which 30 out of the 100 inducing points have converged after training (recall that the ELBO in eq.~\eqref{eq:ELBO} is also maximized w.r.t. the inducing locations $\tilde\bX$). 
For instance, the first column shows the locations of three inducing points which are classified as $0$ by SVGPCR (according to the estimated $\bm_k$, recall eq.~\eqref{eq:parametric_form3}), and analogously for the rest of the columns.
It is very interesting to notice that the inducing point locations comprise different calligraphic representations (in terms of shape, orientation and thickness) of the same digit.
This is related to their intuitive role of entities that summarize the training data.  

\begin{figure}
    \centering
    \begin{tabular}{cc}
        \includegraphics[width=0.47\columnwidth]{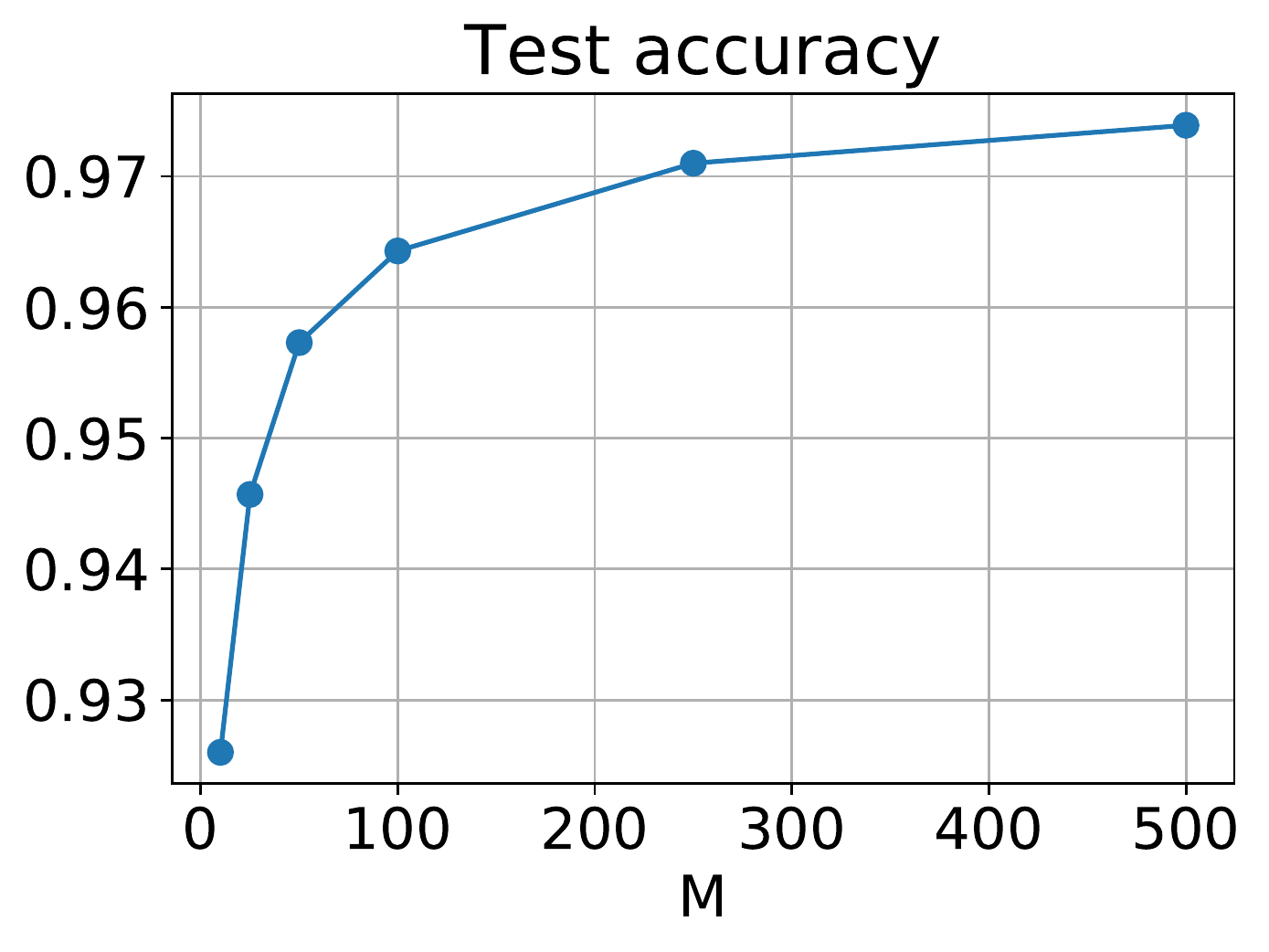} & 
        \includegraphics[width=0.47\columnwidth]{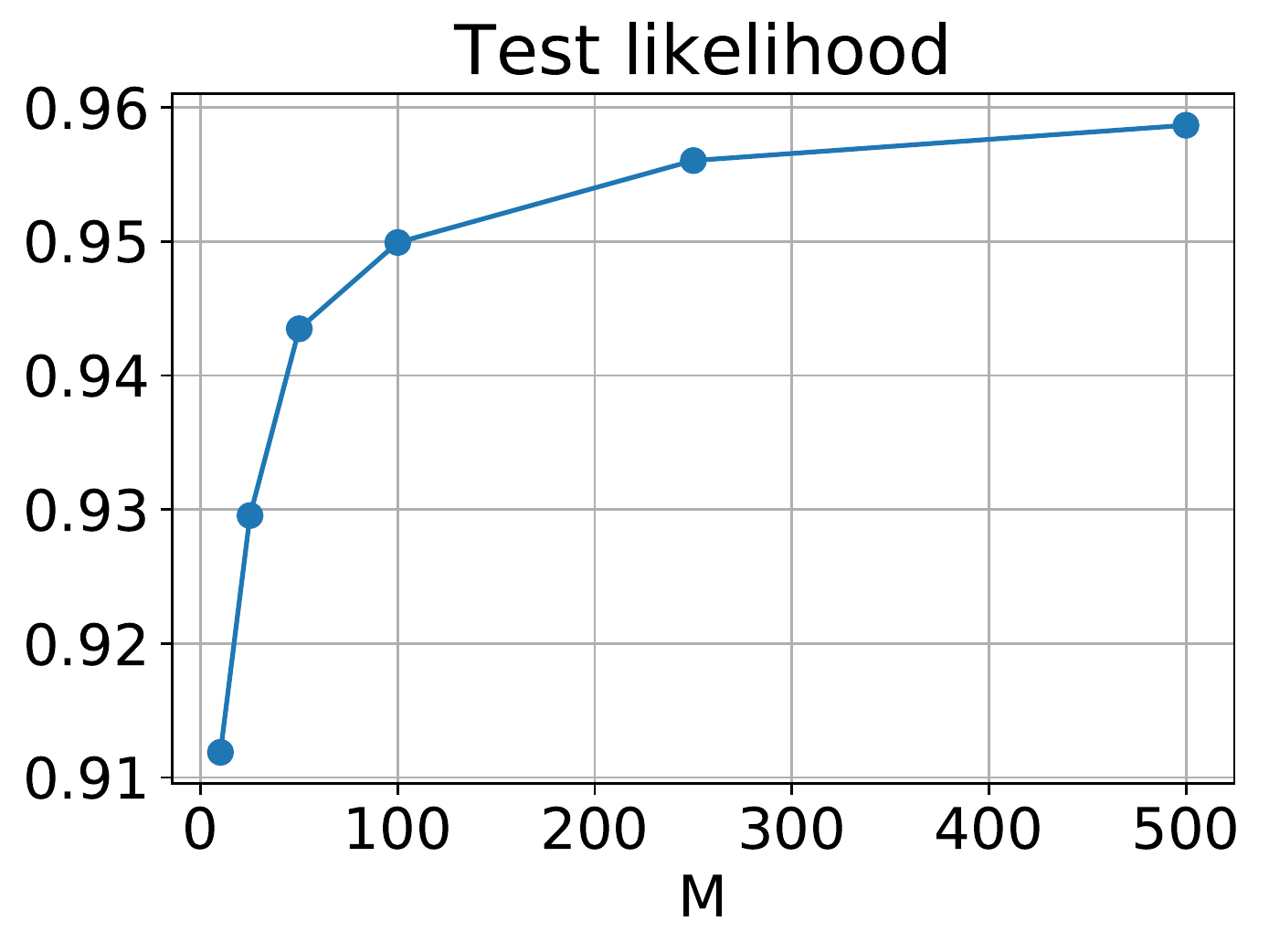}  
        \\
        \includegraphics[width=0.48\columnwidth]{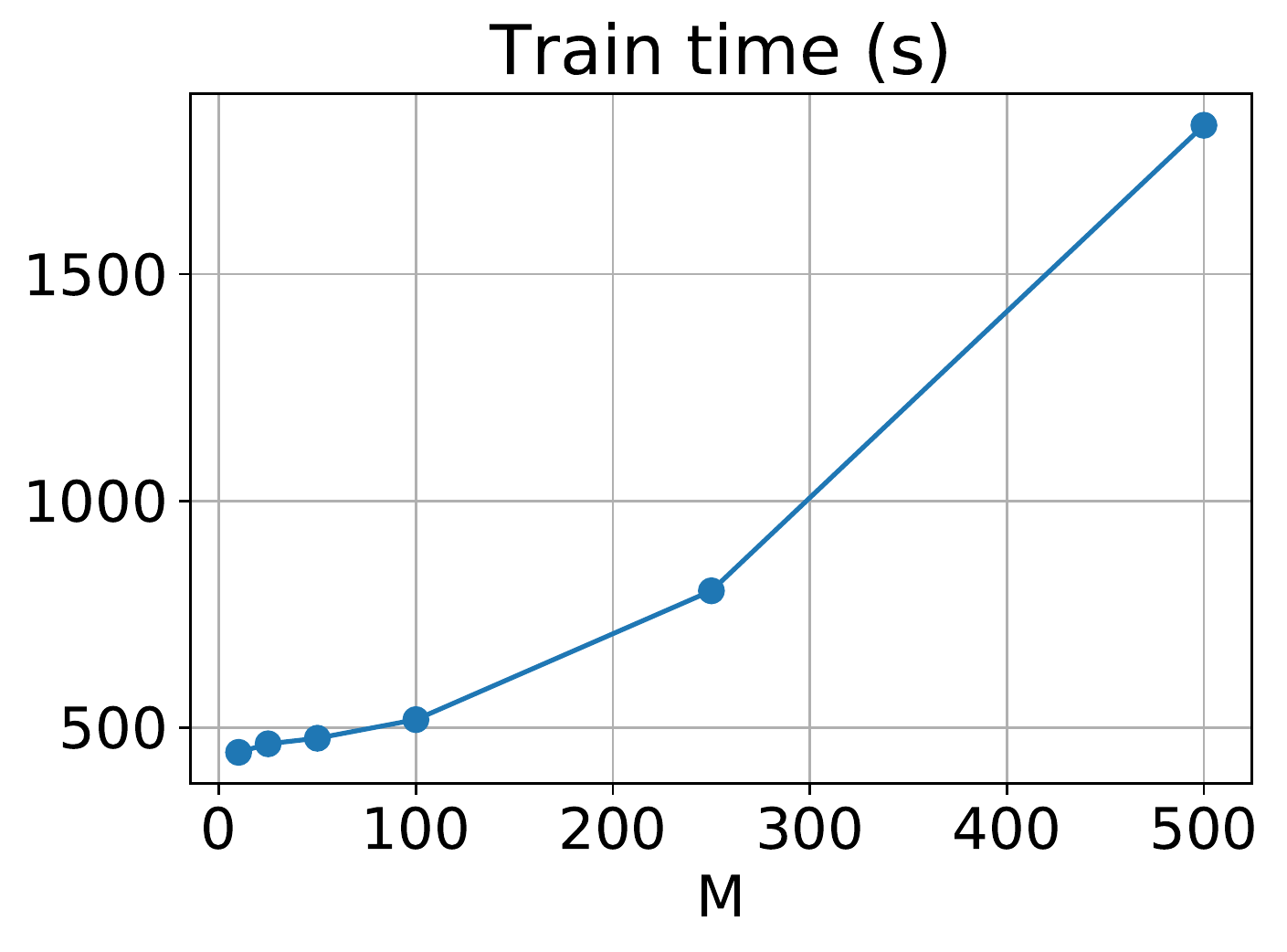} & 
        \includegraphics[width=0.47\columnwidth]{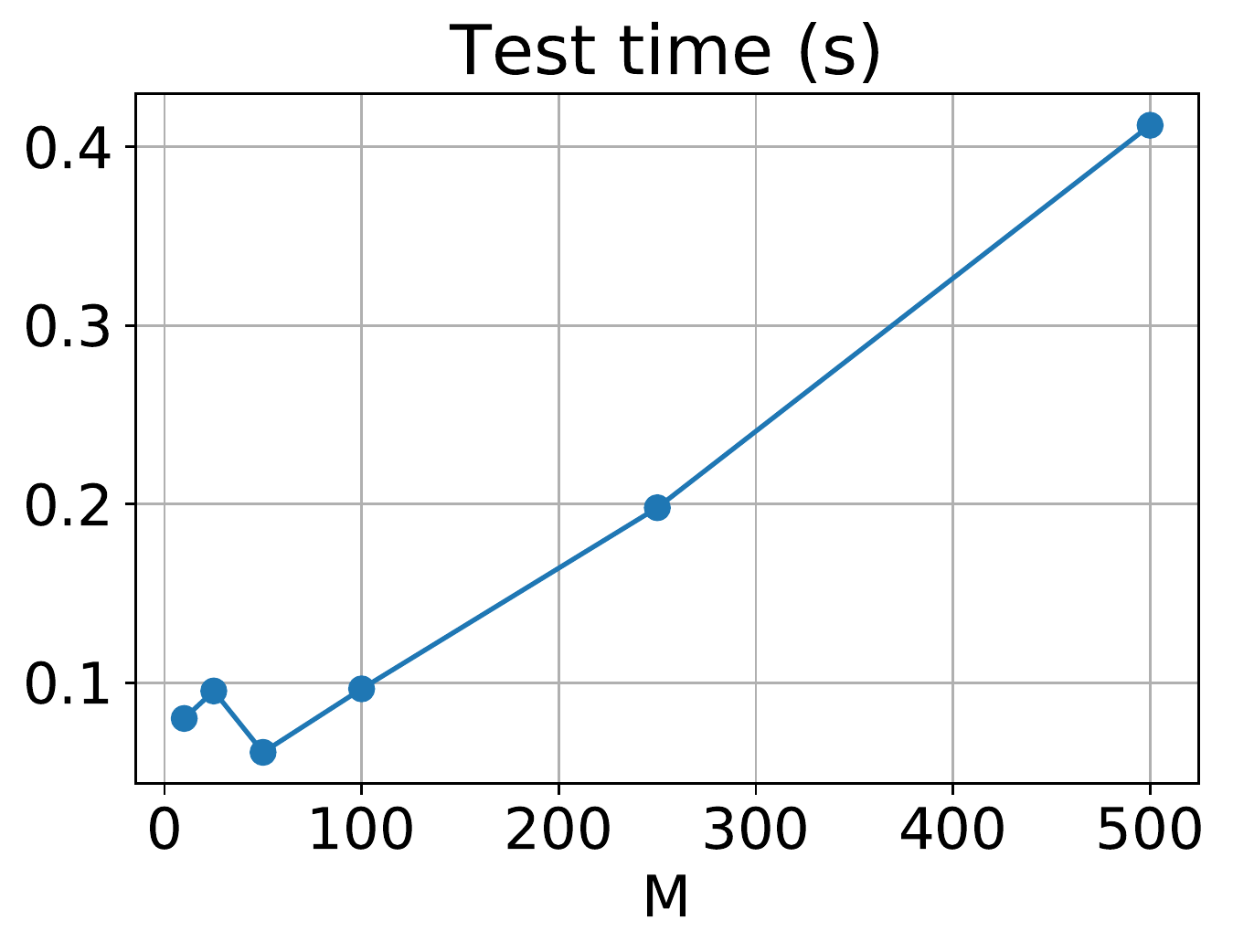}  
    \end{tabular}
    \caption{Influence of $M$ (number of inducing points) on the test performance (accuracy and likelihood) and computational cost (elapsed time at training and testing) for the novel SVGPCR in the MNIST problem. As theoretically expected, more inducing points lead to better test performance at the expense of a higher computational cost.
    }
    \label{fig:MNIST_depOnM}
\end{figure}

Next, let us study the influence of $M$ (the number of inducing points) on the behavior of SVGPCR. 
Figure \ref{fig:MNIST_depOnM} shows the dependence on $M$ of four different metrics: two measures of the test performance (accuracy and mean likelihood), and two related to the computational cost (at training and test steps).
As expected from the theoretical formulation in section \ref{sec:theory}, a greater number of inducing points implies a higher performance at test (in both accuracy and mean likelihood), since the expressiveness of the model is higher.
However, this also leads to heavier train and test costs, since there are more parameters to be estimated (inducing locations $\tilde\bX$, $\bm_k$, and $\bS_k$), and the size of several matrices increase. 

\begin{figure}
    \centering
    \includegraphics[width=0.7\columnwidth]{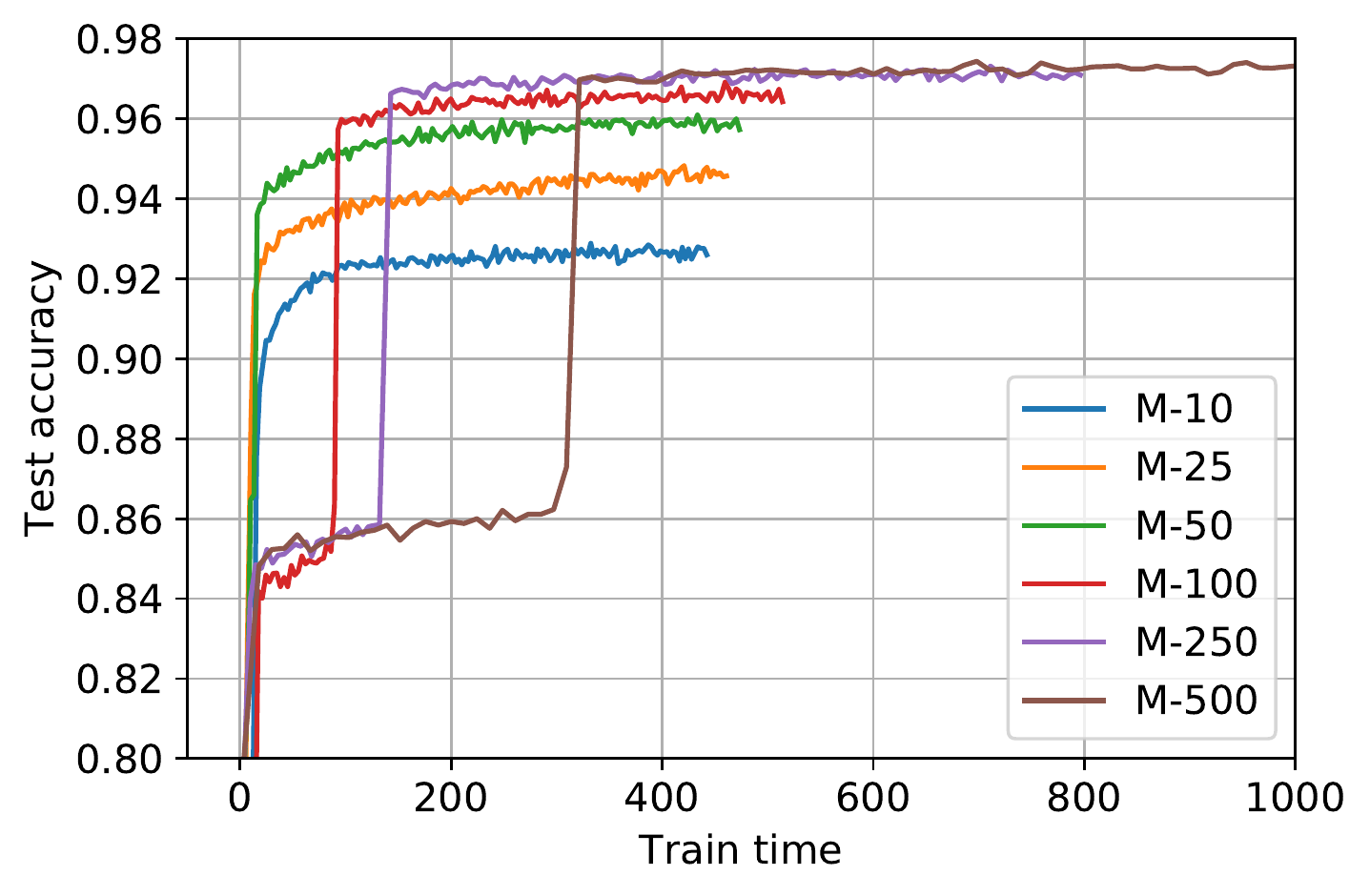}
    \caption{Evolution of test accuracy as a function of the training time for different values of $M$ in the MNIST problem. If there is no limit on the available training time, then high values of $M$ must be selected (as long as it allows for the kernel matrix inversion). However, lower values would be more appropriate for a fast training, since the amount of parameters to be trained significantly reduces. Moreover, when a certain (problem-dependent) $M$ has been reached, there is no a significant benefit by increasing it (observe the difference from $M=250$ to $M=500$).}
    \label{fig:MNIST_tradeoff}
\end{figure}

Moreover, for a given $M$, the model is expected to obtain better test performance as the training time evolves (i.e., when more epochs are run). 
In order to further investigate this, figure \ref{fig:MNIST_tradeoff} shows the test accuracy of SVGPCR as the training time evolves, for different values of $M$.
It is interesting to observe that, the more inducing points, the higher values of test accuracy can be potentially reached, but also a greater amount of training time is needed to reach that precision (notice that the steps which take $M=100,250,500$ to the level of their final precision happen increasingly later).
The conclusion is that, for a given computational budget, the $M$ to be selected is the highest one that can reach convergence in that time (logically, assuming that it allows for the inversion of the associated kernel matrix, i.e., usually $M<10^4$).

\begin{figure}
    \centering
    \begin{tabular}{cc}
        \includegraphics[width=0.47\columnwidth]{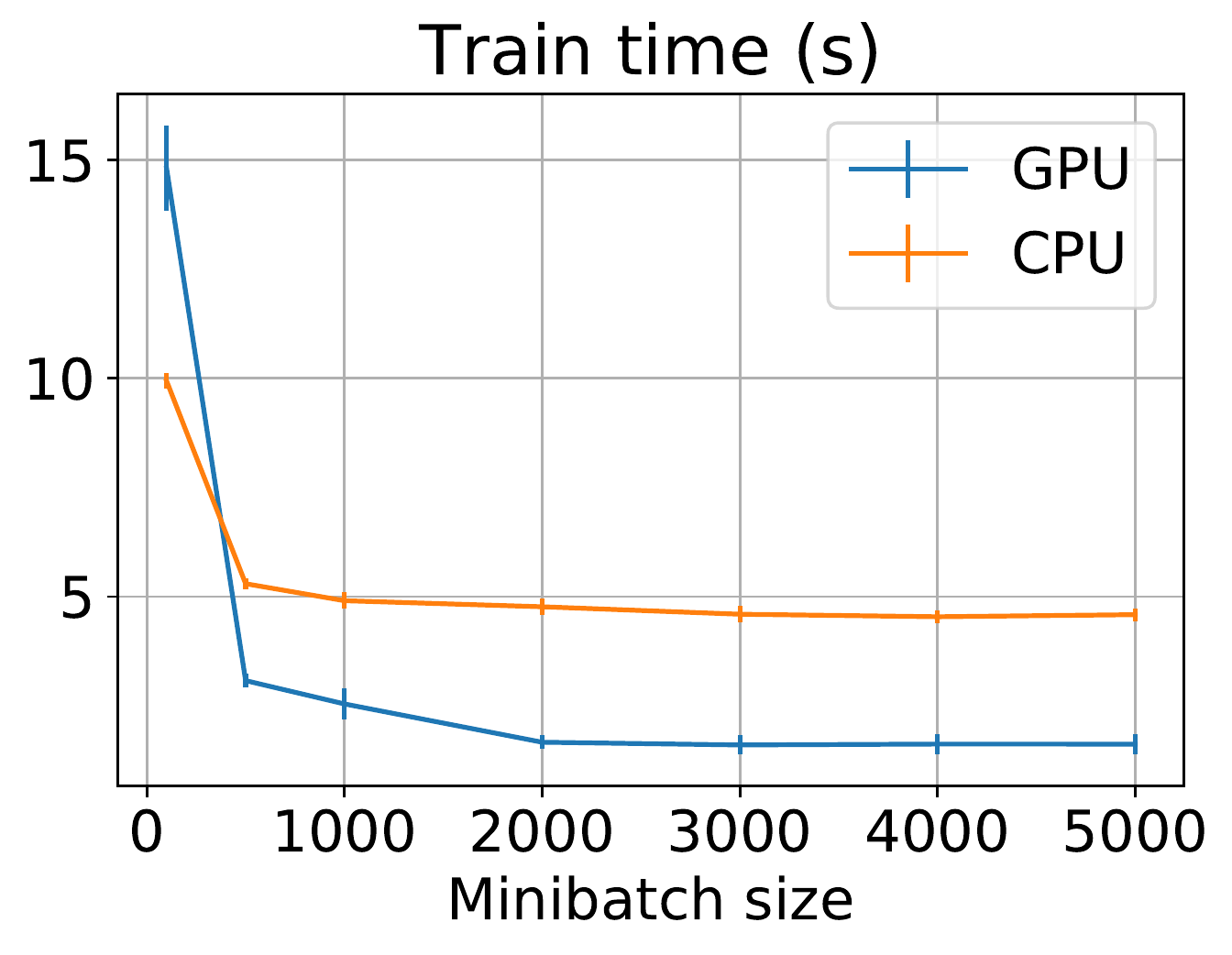} & 
        \includegraphics[width=0.49\columnwidth]{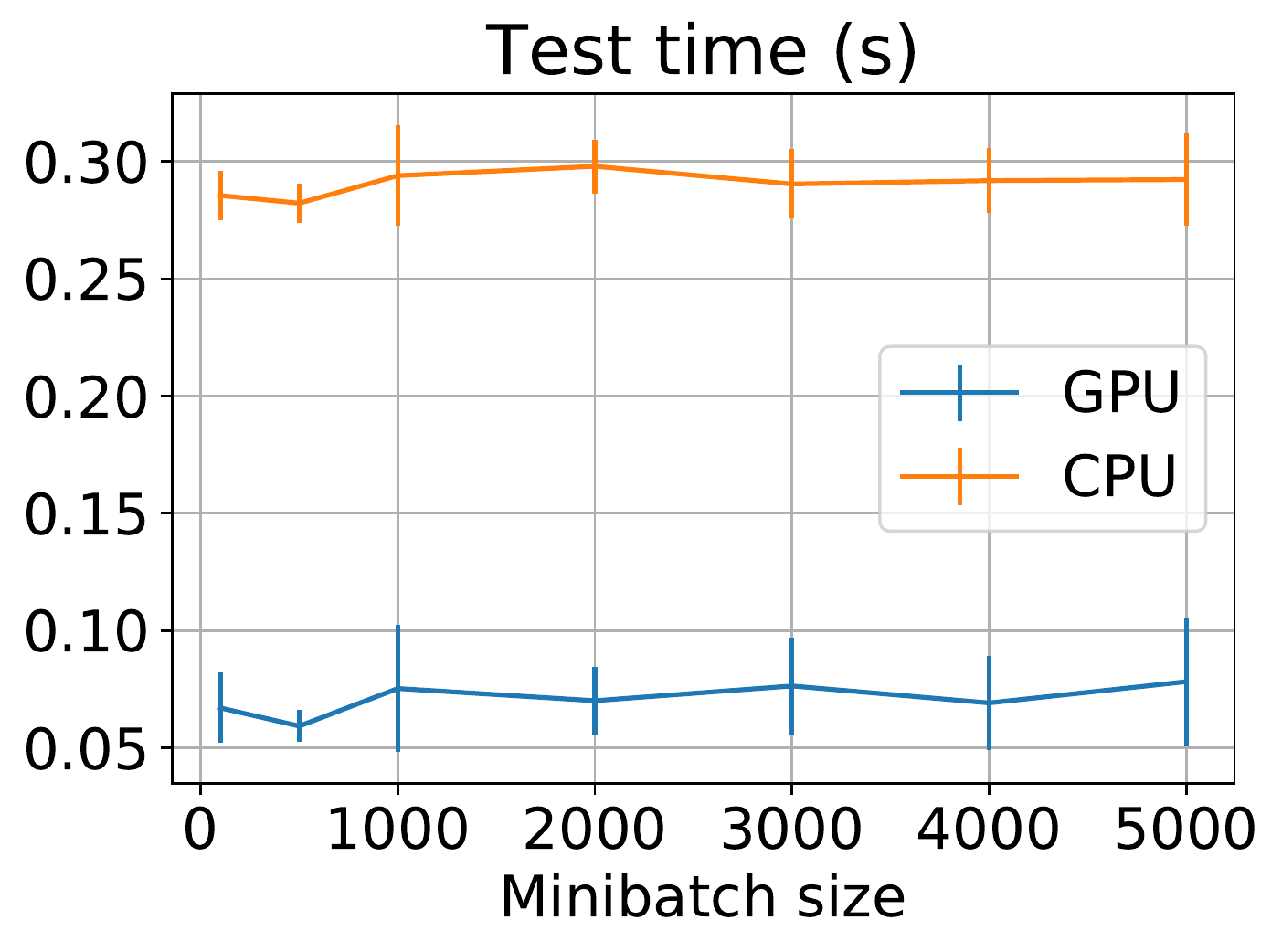}
    \end{tabular}
    \caption{Comparison between CPU and GPU
    implementations of SVGPCR in terms of computational cost for the MNIST problem. At training, the time depends on the minibatch size, since a greater minibatch implies computations with larger matrices and less memory copies, which benefits the GPU. With a minibatch of size 500 (the default used in this section), the GPU is around two times faster. These values are per epoch. At testing (production time), the GPU is over three times faster, and logically does not depend on the minibatch size. In this case, the shown values are for the whole test set.}
    \label{fig:MNIST_CPUGPU}
\end{figure}

Finally, since the associated code can leverage GPU acceleration through GPflow \cite{matthews2017gpflow}, let us compare CPU and GPU implementations.
Figure \ref{fig:MNIST_CPUGPU} shows that, for training, the GPU is usually the preferred choice, unless the minibatch size is very small, in which case the amount of memory copies from CPU to GPU does not compensate the advantage provided by the latter.
In test, the GPU is always faster, since it involves much less data transfers from CPU to GPU.

\subsection{Comparison to classical probabilistic approaches}\label{sec:exp_LIGO_bin}

As explained in section \ref{sec:intro}, the most popular approaches to crowdsourcing jointly model a classifier for the underlying true labels along with the annotators' behavior.
The first works used basic logistic regression as the classifier, e.g. \emph{Raykar} \cite{Raykar:10} and \emph{Yan} \cite{Yan:14} (the difference between them is the noise model considered for the annotators).
However, they struggled when dealing with complex non-linear data sets.
Then, Gaussian Processes became the preferred choice, since their non-parametric form and accurate uncertainty quantification yielded much better results, e.g. \emph{Rodr14} \cite{Rodrigues:14} and VGPCR \cite{CR_PR} (they differ in the inference procedure used, Expectation Propagation \cite{Minka:01} and Variational Inference \cite{blei2017variational}, respectively).
However, the poor scalability of GPs hampered the wide adoption of these approaches in practice.
This motivated the development of the so-called RFF and VFF algorithms, which leverage Random Fourier features approximations to GPs to propose two more scalable GP-based crowdsourcing methods \cite{CR_INFFUS}.
These approaches significantly improve the scalability, reducing it from cubic $\mathcal{O}(N^3)$ to linear $\mathcal{O}(ND^2)$ (with $D$  the number of Fourier frequencies used,  $D\ll N$, see \cite{CR_INFFUS}).
In practice, this implies moving from manageable data sets of $N\approx 10^4$ up to $N\approx 10^5/10^6$.
However,  RFF and VFF do not factorize in mini-batches, which prevents them from reaching data sets of virtually any size.

In the last few years, these classical (mainly GP-based) approaches have been replaced by crowdsourcing methods based on Deep Learning (DL) \cite{Albarqouni:16, Rodrigues:18}. These achieve excellent scalability through mini-batches, and can handle data sets of almost any size.
Because of this, they have become the state of the art approach for real-world crowdsourcing problems.
In the next section \ref{sec:exp_LIGO_full}, we will bring GP-based methods back to a state of the art level. We will show that the novel SVGPCR is competitive with DL-based methods, and additionally provides a very accurate control of uncertainty. 
But before this, it is worth to analyze here the advances that SVGPCR introduces over its predecessors classical (mainly GP-based) crowdsourcing approaches.

More specifically, let us compare SVGPCR with the aforementioned \emph{Raykar}, \emph{Yan} (based on logistic-regression), \emph{Rodr14}, VGPCR (based on GPs), and RFF, VFF (based on scalable approximations to GP). 
Since most of them were formulated for binary problems, we consider a binary task relevant to astrophysicists in GravitySpy.
Using the data set presented in section \ref{sec:data}, the goal is to distinguish between the glitch called ``Other'' and the rest of types. 
This is important in order to identify potential overlaps between that catch-all class and the rest of glitches.
Moreover, it introduces an imbalanced scenario, since ``Other'' represents only a $10.12\%$ of the total amount of annotations. We will use the area under the ROC curve (AUC) as test performance metric.

\begin{figure}
    \centering
    \includegraphics[width=0.9\columnwidth]{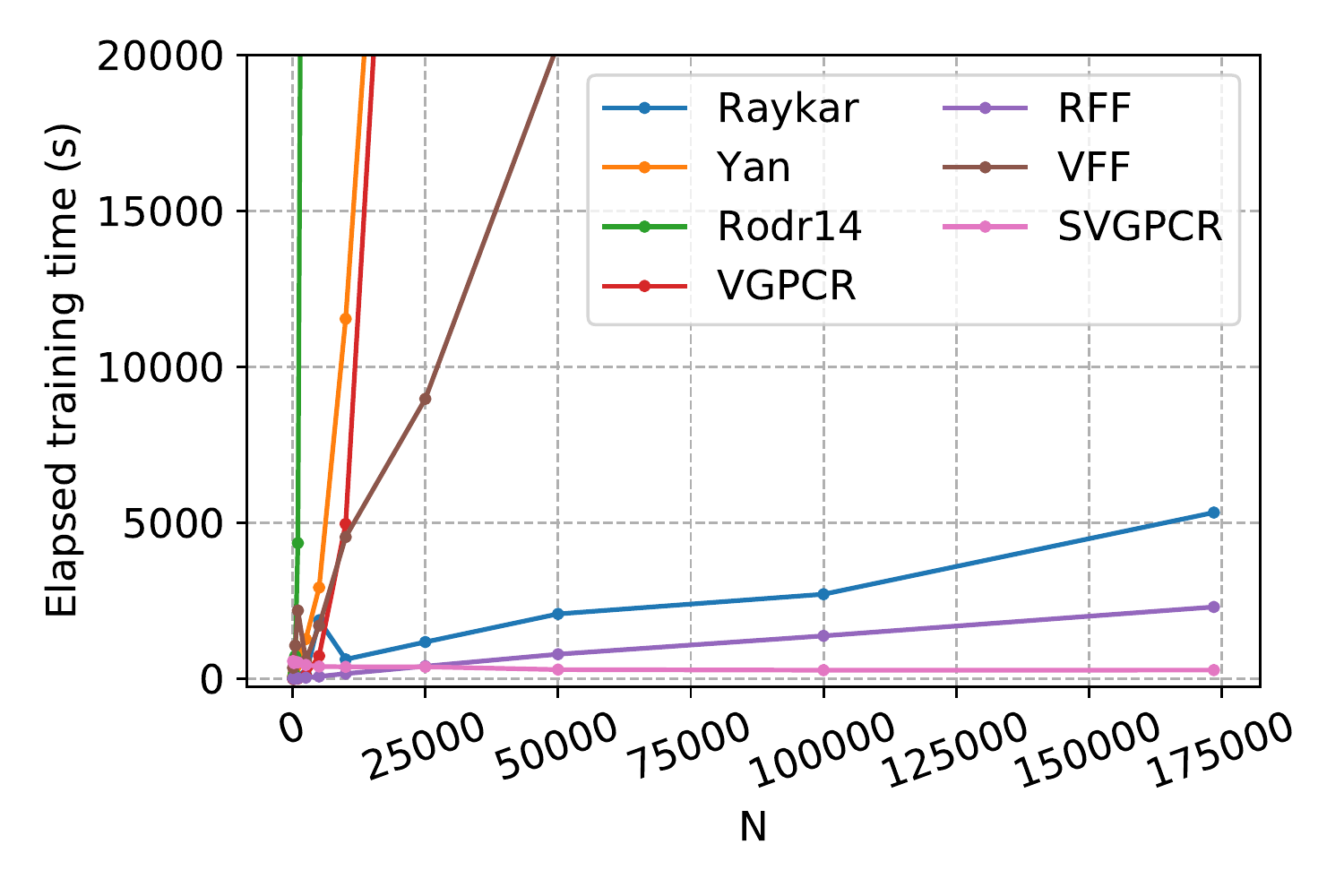}
    \includegraphics[width=0.9\columnwidth]{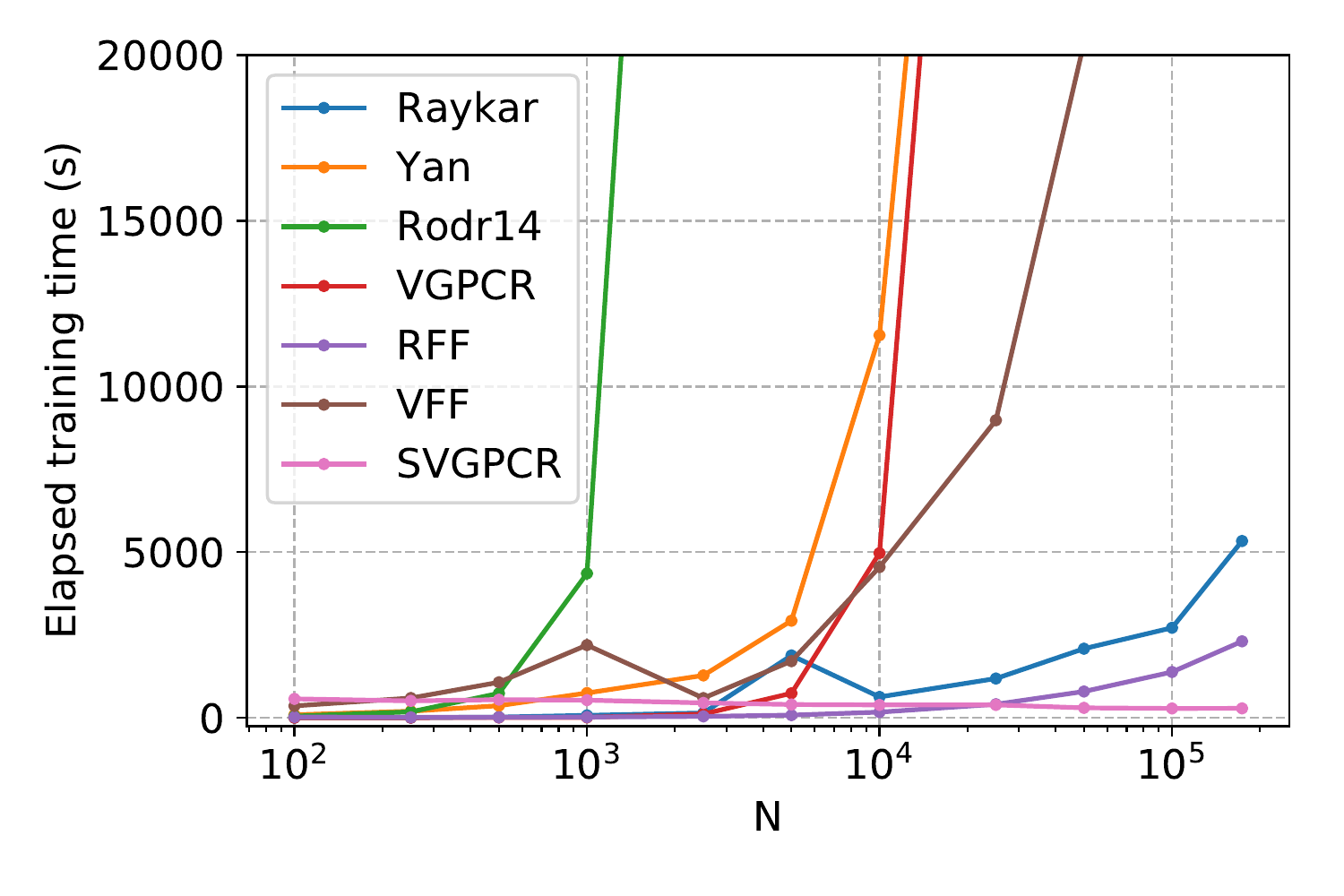}
    \caption{Elapsed training time as a function of the training set size $N$ in the binary LIGO data set. The mean over five independent runs is shown.
    Above, a standard linear scale is used for the x-axis. 
    Notice that SVGPCR exhibits a significantly better scalability than classical probabilistic methods, which is due to its factorization in mini-batches.
    Moreover, among previous approaches, we can distinguish between those that have already exploded for $N=25000$ (\emph{Yan}, \emph{Rodr14}, VGPCR, VFF), and those which have not yet for the full set size (\emph{Raykar}, RFF). 
    In order to better appreciate the differences, a logarithmic scale is used for the x-axis in the figure below.  
    This further shows that \emph{Rodr14} shoots up as early as $N=1000$, \emph{Yan}, VGPCR and VFF do it around $N=10000$, and \emph{Raykar} and RFF are starting beyond $N=10^5$.
    In order to clarify both figures, the y-axis upper limit has been set to $20000$ seconds (otherwise, most lines collapse onto the x-axis).  
    }
    \label{fig:binaryLIGO_time}
\end{figure}

Figure \ref{fig:binaryLIGO_time} compares the scalability of the compared methods as the training set grows.
The novel SVGPCR clearly stands out as the most scalable approach. This can be attributed to its training scheme through mini-batches, which considerably alleviates the dependence on the training set size.
The rest of methods explode at different moments: the heavy EP inference of \emph{Rodr14} only allows for training with up to $N=2500$, the GP-based formulation of VGPCR and the complex annotators noise model of \emph{Yan} make them reach $N=25000$ with difficulties. In spite of the GP approximation, VFF does not go beyond $N=10^5$ in this problem, because of the expensive optimization of Fourier features. Finally, \emph{Raykar} (which is based on cheap logistic regression) and RFF (which does not optimize over the Fourier features) can cope with the full data set, although they are significantly slower than SVGPCR.

\begin{figure}
    \centering
    \includegraphics[width=8cm]{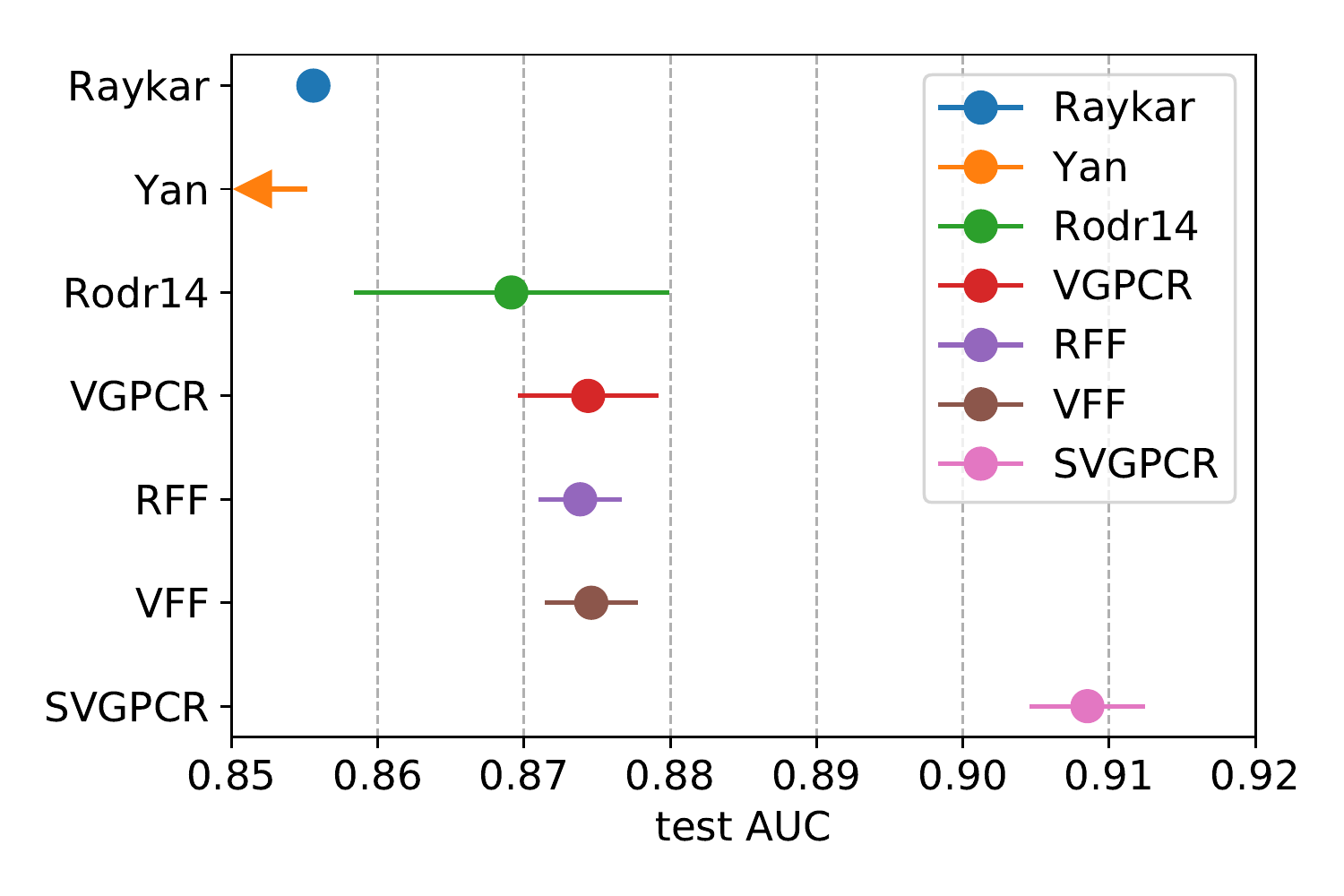}
    \caption{Test AUC achieved in the binary LIGO data set by the compared models (in each case, the largest one that could be trained). The mean and one standard deviations are shown. SVGPCR clearly outperforms methods which cannot cope with the full data set (\emph{Yan}, \emph{Rodr14}, VGPCR, VFF). Moreover, RFF and \emph{Raykar} are also beaten because of their more limited formulation (see the main text).}
    \label{fig:binaryLIGO_AUC}
\end{figure}

Moreover, figure \ref{fig:binaryLIGO_AUC} shows that their test performance is pretty far from that of the novel SVGPCR.
Indeed, the logistic regression model underlying \emph{Raykar} is not sufficient for the nonlinear problem at hand, and the GP approximation provided by RFF is known to be poor when the dimensionality of the problem is high \cite{CR_INFFUS} (like here, where we are working with $256$ features, recall section \ref{sec:data}).
The rest of methods are also clearly outperformed, since their limited scalability prevents them from processing the full data set.
Interestingly, figure \ref{fig:binaryLIGO_AUC} shows an intuitive and logical structure: the more simple logistic-regression based methods are located on the left (less test AUC), the classical GP-based ones in the central part, and the novel SVGPCR on the right.

\subsection{Comparison with state of the art DL-based methods}\label{sec:exp_LIGO_full}

In the last years, Deep Learning has emerged as a scalable alternative to model crowdsourcing problems. 
Two of the most popular approaches are AggNet \cite{Albarqouni:16} and the various crowd layers proposed in \cite{Rodrigues:18}.
The former considers a deep neural network (DNN) as underlying classifier, and a probabilistic noise model for annotators based on per-user confusion matrices.
Then, the training step follows an iterative expectation-maximization (EM) scheme between both parts of the model \cite[Section 9.4]{Bishop:06}.
Alternatively, the crowd layers in \cite{Rodrigues:18} allow for end-to-end training of the DNN, without the need for the EM scheme.
This is significantly cheaper in terms of computational cost, although the probabilistic formulation of AggNet allows for a better uncertainty quantification.
The three crowd layers studied in \cite{Rodrigues:18} will be considered here: CL-VW, CL-VWB and CL-MW. They differ in the parametric form of the annotator noise model, which is increasingly complex: a vector of per-class weights for CL-VW, an additional bias for CL-VWB, and a whole confusion matrix for CL-MW.

These four DL-based methods (AggNet, CL-VW, CL-VWB, CL-MW) are compared against three increasingly complex SVGPCR models: SVGPCR-10, SVGPCR-50, SVGPCR-100, where each number represents the amount of inducing points used.
As all these approaches are defined for multi-class tasks, the full LIGO problem in section \ref{sec:data} can be addressed now. 

Tables \ref{tab:LIGO_accuracy} and \ref{tab:LIGO_likelihood} show the global and per-class test performance of the compared methods.
Table \ref{tab:LIGO_accuracy} is devoted to the test accuracy, which relies only on the mode of the predictive distribution and is less influenced by the uncertainty quantification of the model. Table \ref{tab:LIGO_likelihood} shows the test likelihood, which additionally depends on the uncertainty of the predictive distribution, and therefore depends more heavily on its accurate control within the model.

In both tables, SVGPCR stands out as the best-performing method globally.
The difference is greater in the case of the test likelihood, which is logically explained by the excellent uncertainty quantification of GPs. 
Indeed, the better control of uncertainty also justifies that AggNet outperforms CL-based methods in test likelihood (whereas they are very similar in accuracy).
Moreover, observe that the global superiority of SVGPCR is not due to a great result in only one or two very populated classes.
Instead, SVGPCR performs consistently well across the 15 glitch types in both tables, winning in few of them (a bit more in test likelihood, as logically expected).
According to astrophysicists at GravitySpy, this regularity across classes is a very desirable property for a reliable glitch detection system.

It is also worth to notice that $M=50$ inducing points seem enough for the problem at hand. In both tables \ref{tab:LIGO_accuracy} and \ref{tab:LIGO_likelihood}, a significant improvement is observed from $M=10$ to $M=50$, but $M=100$ produces very similar results. This small value of $M=50$ hints at a not very complex internal structure of the data. 
It is also interesting to observe that, in general, the most difficult classes are ``Repeating Blips'' and ``Other'' (recall the 15 types in figure \ref{fig:glitches_full}).
This discovery is not surprising for astrophysicists in GravitySpy, since the former is usually confused with ``Blips'', and the latter is a catch-all class to which some conservative annotators resort too often. 
The case of ``Other'' is also related to the interest of astrophysicists to study it separately in the experiment of previous section \ref{sec:exp_LIGO_bin}.


\begin{table*}
\centering
\caption{Per-class and global test accuracy for the compared methods in the LIGO experiment. Mean and standard deviation over ten runs are shown. Globally, the novel SVGPCR with enough inducing points outperforms DL-based methods by 2\%. In per-class results, notice the regularity of SVGPCR, which performs well across all classes without standing out in many of them. 
}
\label{tab:LIGO_accuracy}
\begin{tabular}{r|ccccccc}
                   & AggNet                & CL-VW                & CL-VWB               & CL-MW                & SVGPCR (M=10) & SVGPCR (M=50)         & SVGPCR (M=100)        \\ \hline
1080LINE           & \textbf{.9791(.0045)} & .9676(.0063)          & .9732(.0059)          & .9746(.0063)          & .9781(.0075)  & .9720(.0076)          & .9720(.0069)          \\
1400RIPPLE         & \textbf{.9447(.0237)} & .1666(.3335)          & .3243(.4001)          & .0000(.0000)          & .0000(.0000)  & .8528(.0203)          & .8577(.0171)          \\
BLIP               & .9737(.0078)          & .9754(.0079)          & .9741(.0135)          & \textbf{.9826(.0062)} & .8712(.2904)  & .9643(.0044)          & .9622(.0052)          \\
EXTR.LOUD      & \textbf{.8704(.0566)} & .7193(.0652)          & .8034(.0660)          & .5272(.3502)          & .3681(.3705)  & .7261(.0398)          & .7295(.0427)          \\
KOIFISH            & .8760(.0634)          & .8749(.0151)          & \textbf{.9043(.0296)} & .9023(.0203)          & .8007(.2673)  & .8844(.0129)          & .8828(.0115)          \\
L.F.BURST  & .7649(.0802)          & .8639(.0299)          & .9023(.0211)          & \textbf{.9146(.0174)} & .7912(.2638)  & .8864(.0087)          & .8861(.0105)          \\
L.F.LINE   & .8603(.0742)          & .8603(.0274)          & .8514(.0293)          & .8566(.0279)          & .8354(.2785)  & .9150(.0105)          & \textbf{.9156(.0111)} \\
NOGLITCH           & \textbf{.9174(.0354)} & .8835(.0201)          & .8887(.0241)          & .9141(.0274)          & .7048(.2366)  & .7941(.0223)          & .7951(.0162)          \\
OTHER              & .2786(.0294)          & \textbf{.4167(.0348)} & .3660(.0376)          & .3503(.0297)          & .3622(.1213)  & .3977(.0083)          & .4011(.0091)          \\
P.L.60HZ      & .8885(.0676)          & .8591(.0290)          & \textbf{.9468(.0104)} & .9365(.0179)          & .7417(.2476)  & .8438(.0102)          & .8425(.0127)          \\
REP.BLIPS     & .5487(.0730)          & .0470(.1410)          & .0572(.1717)          & .0000(.0000)          & .0000(.0000)  & .6649(.0215)          & \textbf{.6700(.0210)} \\
SCATT.LIGHT     & .9088(.0492)          & .9497(.0126)          & .9601(.0109)          & \textbf{.9645(.0053)} & .8657(.2886)  & .9600(.0050)          & .9562(.0056)          \\
SCRATCHY           & .8980(.0400)          & .4433(.4437)          & .4426(.4452)          & .4440(.4444)          & .8093(.2702)  & .8953(.0206)          & \textbf{.9000(.0165)} \\
VIOLIN & .9755(.0057)          & \textbf{.9932(.0032)} & .9921(.0033)          & .9930(.0018)          & .8915(.2971)  & .9899(.0027)          & .9914(.0017)          \\
WHISTLE            & .9175(.0236)          & .9359(.0136)          & \textbf{.9438(.0137)} & .9377(.0230)          & .8122(.2712)  & .9166(.0070)          & .9201(.0047)          \\ \hline
GLOBAL             & .8957(.0227)          & .8886(.0126)          & .8985(.0105)          & .8956(.0104)          & .8355(.1919)  & \textbf{.9184(.0031)} & .9183(.0027)         
\end{tabular}
\end{table*}

\begin{table*}
\centering
\caption{Per-class and global test likelihood for the compared methods in the LIGO experiment. Mean and standard deviation over ten independent runs are shown. Globally, the novel SVGPCR with enough inducing points outperforms DL-based methods by almost 3\%.
It also exhibits a desirable regularity across the different classes.
Moreover, notice that, compared to the accuracy in Table \ref{tab:LIGO_accuracy}, there exists here a greater advantage against methods that do not quantify uncertainty (i.e. CL-based ones).}
\label{tab:LIGO_likelihood}
\begin{tabular}{r|ccccccc}
            & AggNet                & CL-VW                & CL-WVB               & CL-MW                & SVGPCR (M=10) & SVGPCR (M=50)         & SVGPCR (M=100)        \\ \hline
1080LINE    & \textbf{.9781(.0048)} & .9515(.0091)          & .9597(.0076)          & .9649(.0074)          & .8811(.2715)  & .9689(.0082)          & .9688(.0075)          \\
1400RIPPLE  & \textbf{.9416(.0242)} & .1644(.3290)          & .3196(.3937)          & .0118(.4874)          & .0067(.0199)  & .8475(.0182)          & .8509(.0156)          \\
BLIP        & .9709(.0077)          & .9753(.0079)          & .9735(.0123)          & \textbf{.9777(.0043)} & .8746(.2694)  & .9606(.0044)          & .9587(.0055)          \\
EXTR.LOUD   & \textbf{.8626(.0574)} & .7214(.0645)          & .8024(.0684)          & .5276(.3503)          & .3651(.3541)  & .7266(.0435)          & .7242(.0408)          \\
KOIFISH     & .8688(.0679)          & .8752(.0161)          & \textbf{.9013(.0300)} & .8959(.0190)          & .8022(.2457)  & .8788(.0109)          & .8784(.0117)          \\
L.F.BURST   & .7575(.0799)          & .7249(.0139)          & .8035(.0177)          & .8419(.0207)          & .7966(.2434)  & \textbf{.8851(.0083)} & .8838(.0098)          \\
L.F.LINE    & .8502(.0786)          & .5808(.0192)          & .6200(.0178)          & .6917(.0242)          & .8378(.2571)  & \textbf{.9125(.0094)} & .9118(.0103)          \\
NOGLITCH    & \textbf{.9091(.0402)} & .8109(.0247)          & .8196(.0342)          & .8196(.0244)          & .7062(.2146)  & .7919(.0238)          & .7932(.0146)          \\
OTHER       & .2692(.0285)          & \textbf{.4123(.0322)} & .3439(.0362)          & .3268(.0275)          & .3686(.1012)  & .3959(.0095)          & .3999(.0091)          \\
P.L.60HZ    & .8700(.0665)          & .8565(.0291)          & \textbf{.8993(.0108)} & .8929(.0117)          & .7435(.2260)  & .8384(.0107)          & .8380(.0107)          \\
REP.BLIPS   & .5323(.0664)          & .0466(.1400)          & .0584(.1752)          & .4878(.8447)          & .0067(.0199)  & .6581(.0191)          & \textbf{.6651(.0198)} \\
SCATT.LIGHT & .8716(.0480)          & .8782(.0165)          & .8947(.0196)          & .9416(.0067)          & .8683(.2673)  & \textbf{.9558(.0047)} & .9520(.0057)          \\
SCRATCHY    & .8823(.0399)          & .4425(.4429)          & .4411(.4440)          & .4414(.4419)          & .7971(.2441)  & .8911(.0215)          & \textbf{.8953(.0176)} \\
VIOLIN      & .9738(.0058)          & .9823(.0034)          & .9815(.0024)          & .9863(.0022)          & .8960(.2765)  & .9875(.0020)          & \textbf{.9886(.0014)} \\
WHISTLE     & .9167(.0215)          & .9341(.0149)          & \textbf{.9427(.0141)} & .9388(.0213)          & .8124(.2490)  & .9155(.0064)          & .9179(.0046)          \\ \hline
GLOBAL      & .8871(.0240)          & .8175(.0106)          & .8387(.0113)          & .8528(.0101)          & .8126(.2487)  & \textbf{.9154(.0033)} & .9149(.0027)         
\end{tabular}
\end{table*}

\begin{figure}
    \centering
    \begin{tabular}{cc}
        \includegraphics[width=0.48\columnwidth]{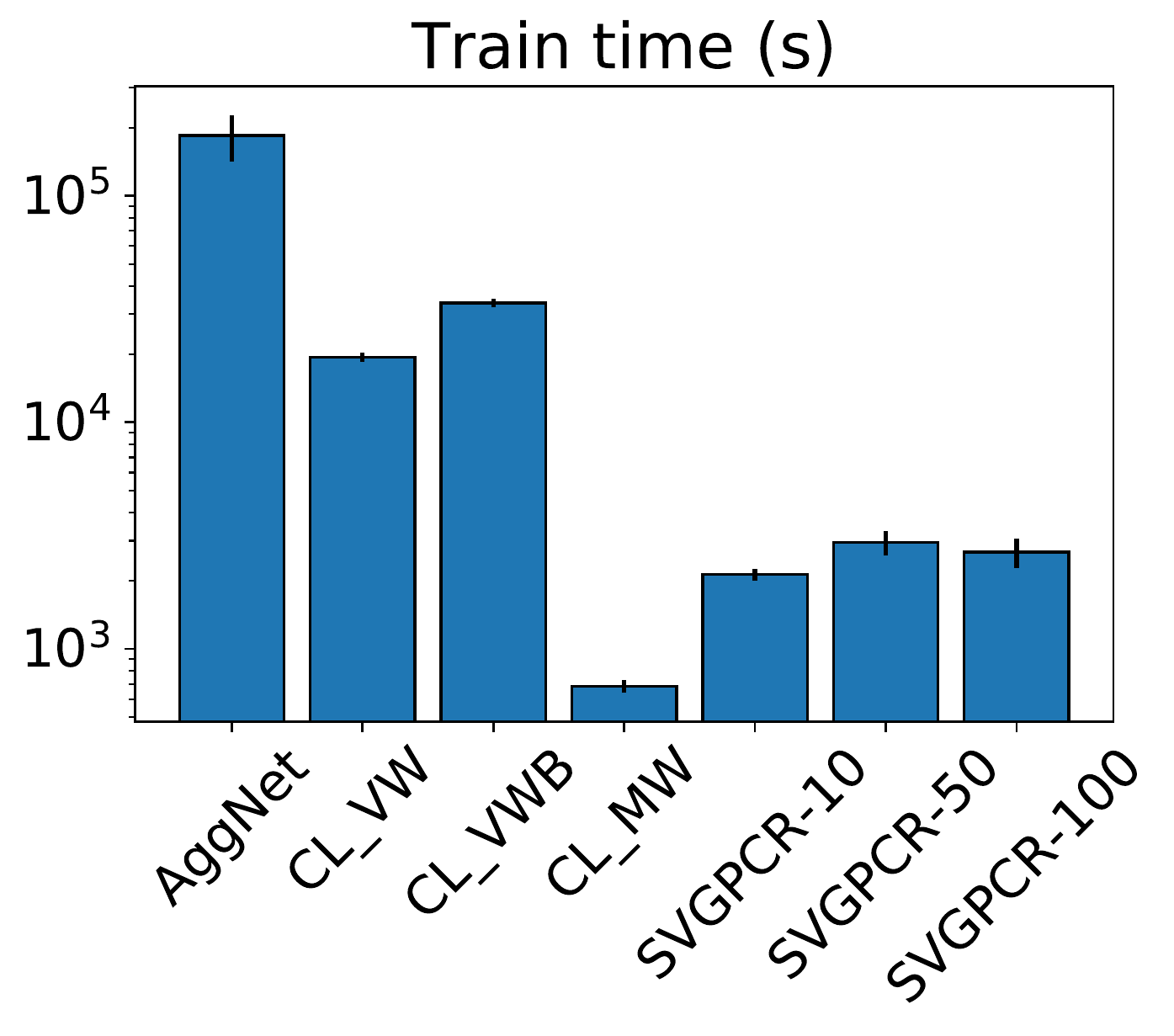} & 
        \includegraphics[width=0.48\columnwidth]{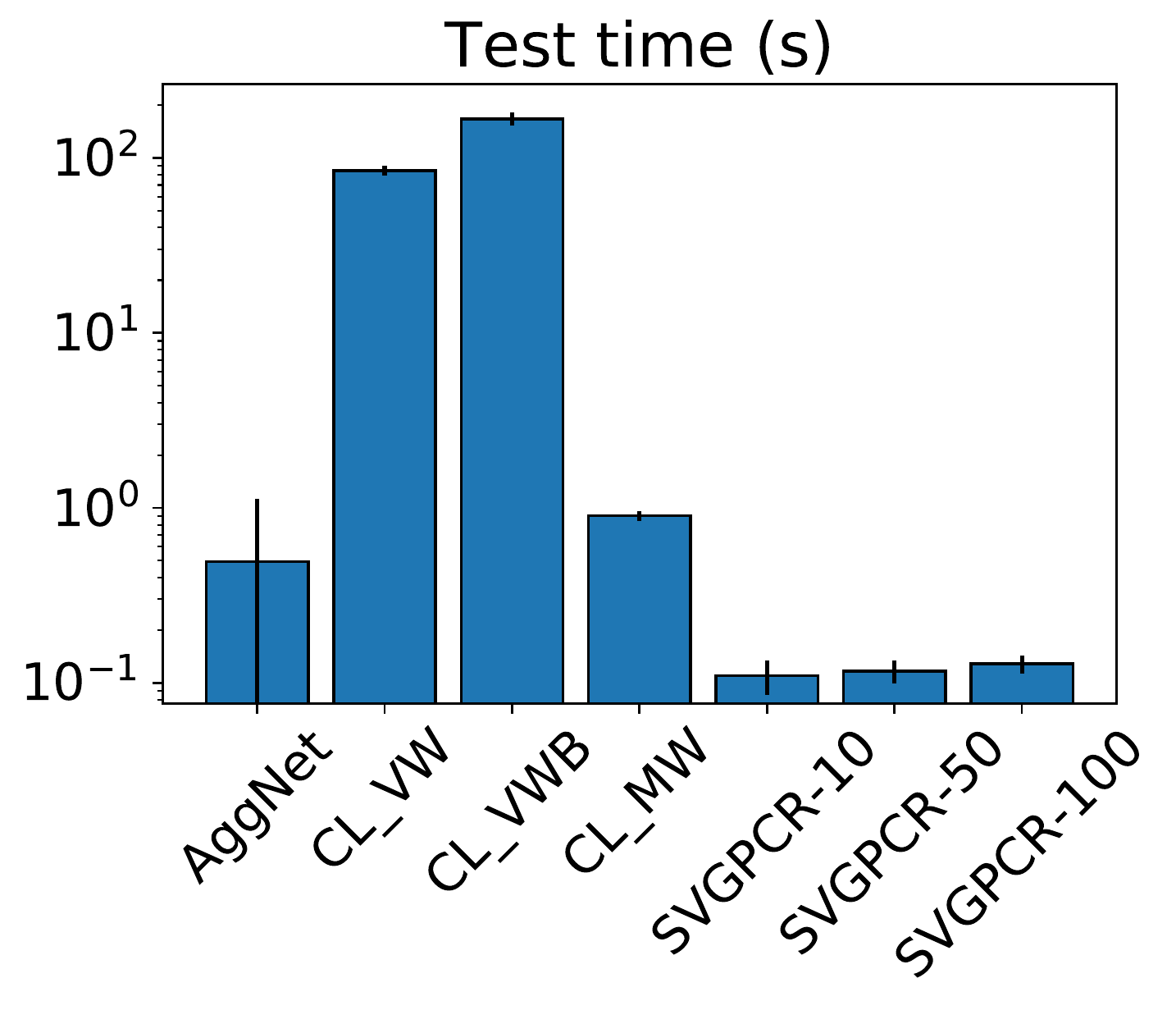}
    \end{tabular}
    \caption{Train and test times (mean and standard deviation over ten independent runs) for the compared methods in the LIGO experiment. Notice the logarithmic scale in y-axis. 
    The proposed SVGPCR is competitive with all the DL-based methods in terms of computational cost at both training and testing.}
    \label{fig:LIGO_times}
\end{figure}

It is also important to highlight that all these methods are scalable enough so as to cope with the full LIGO data set.
More specifically, figure \ref{fig:LIGO_times} shows the elapsed time at training and testing for the compared methods.
In general, the proposed SVGPCR is competitive with DL-based methods in these aspects.
At training, SVGPCR is significantly faster than AggNet due to the heavy iterative EM scheme of the latter, and is slower than CL-MW\footnote{Results of CL-VW/CL-VWB being worse than CL-MW in figure \ref{fig:LIGO_times} might be attributed to implementation inefficiency, since the former include for loops whereas matrix multiplication is used in the latter.}.
Nonetheless, less than one hour of training is a competitive result for a data set with $173565$ instances (recall section \ref{sec:data}).
At testing, SVGPCR is the fastest approach, which is convenient for real-time applications the system might be used for.

\begin{figure}
    \footnotesize
    \centering
    \begin{tabular}{ccc}
    SVGPCR-50 & AggNet & CL-MW \\
    \includegraphics[width=0.29\columnwidth]{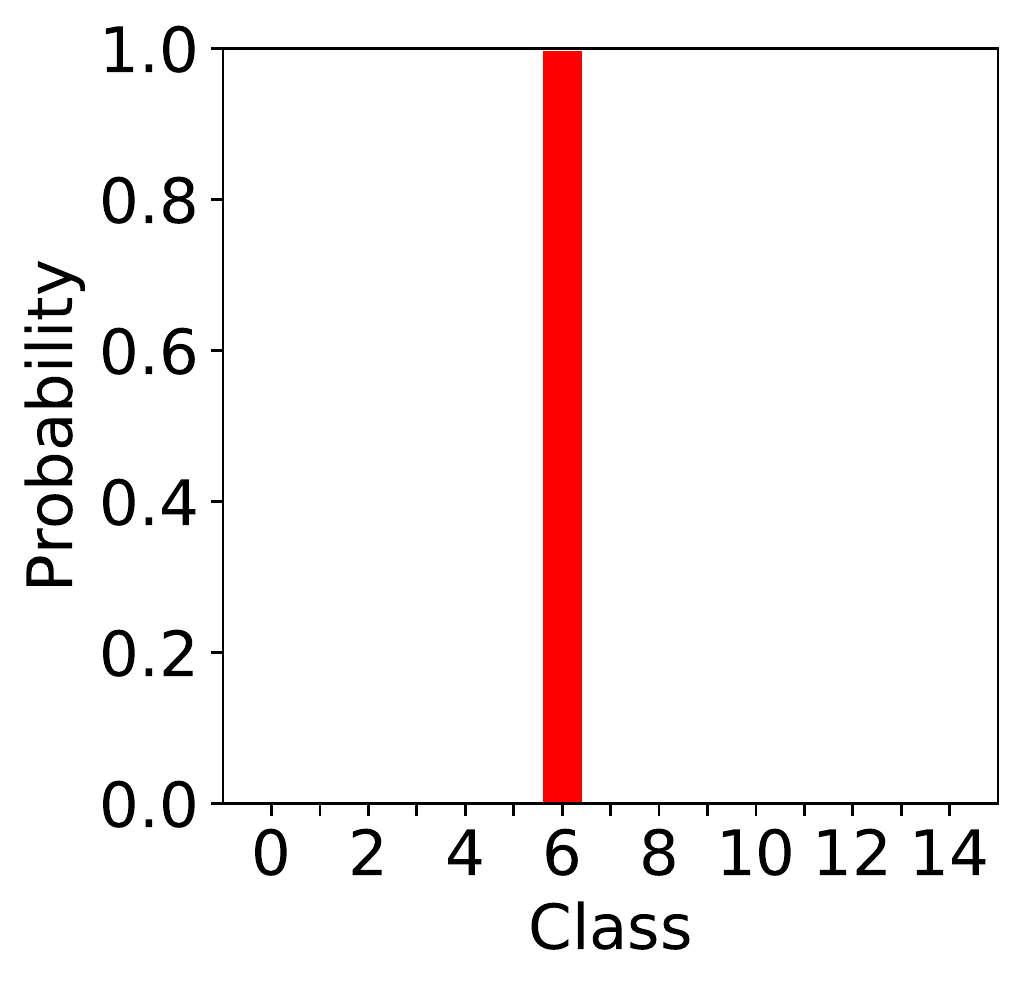} &
    \includegraphics[width=0.29\columnwidth]{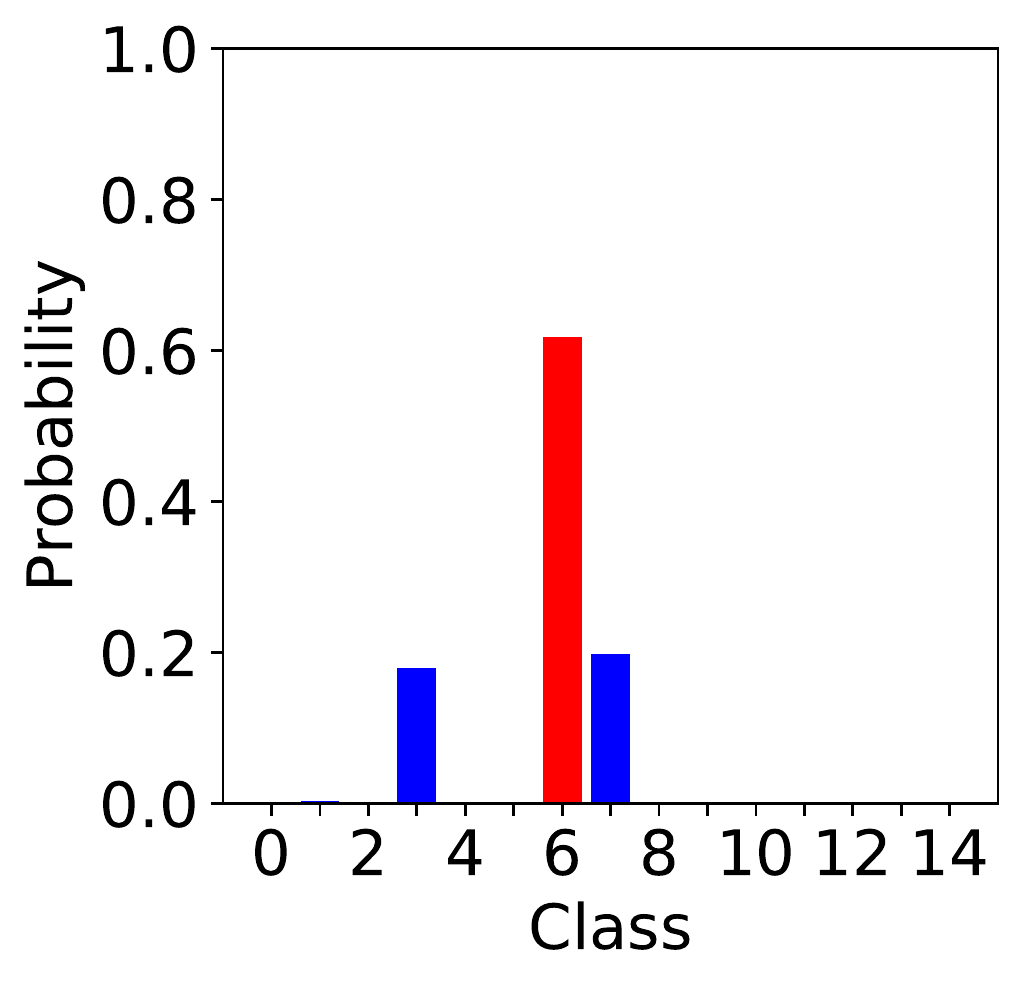} &
    \includegraphics[width=0.29\columnwidth]{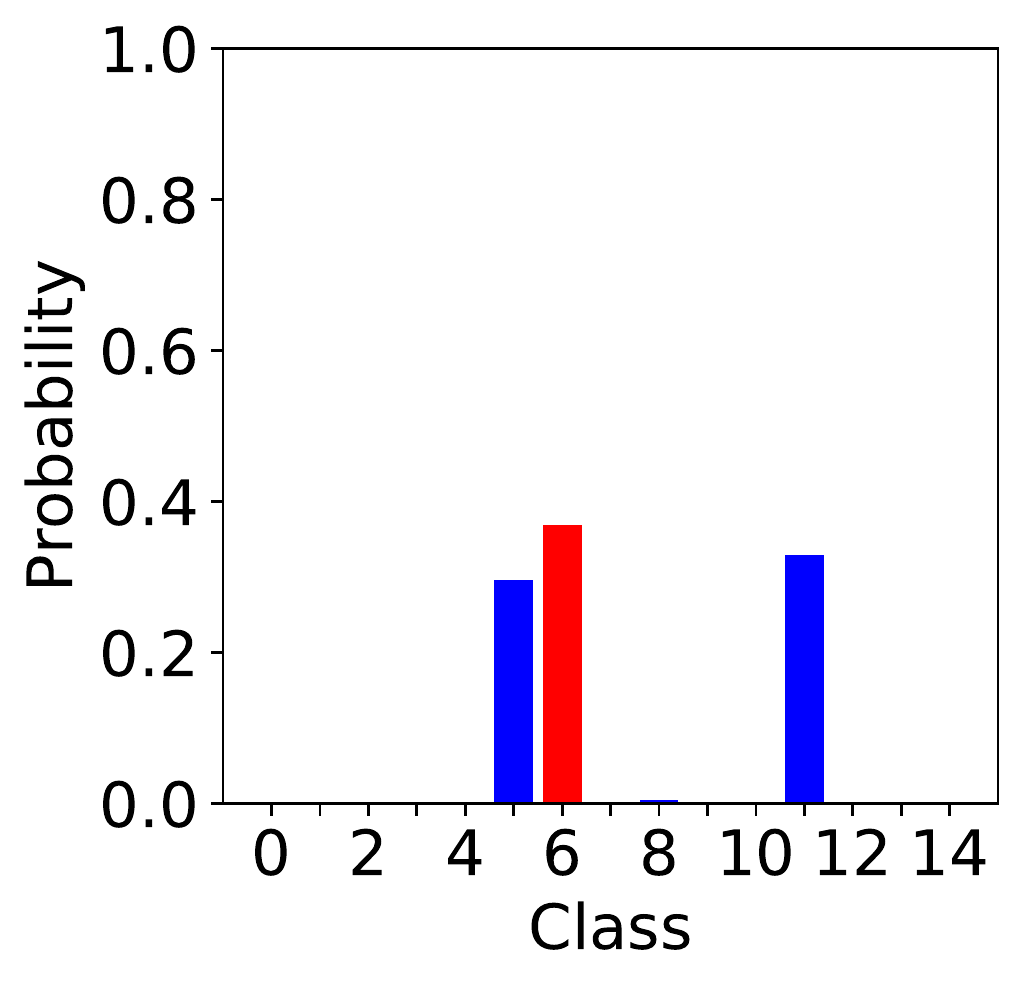} \\
    \includegraphics[width=0.29\columnwidth]{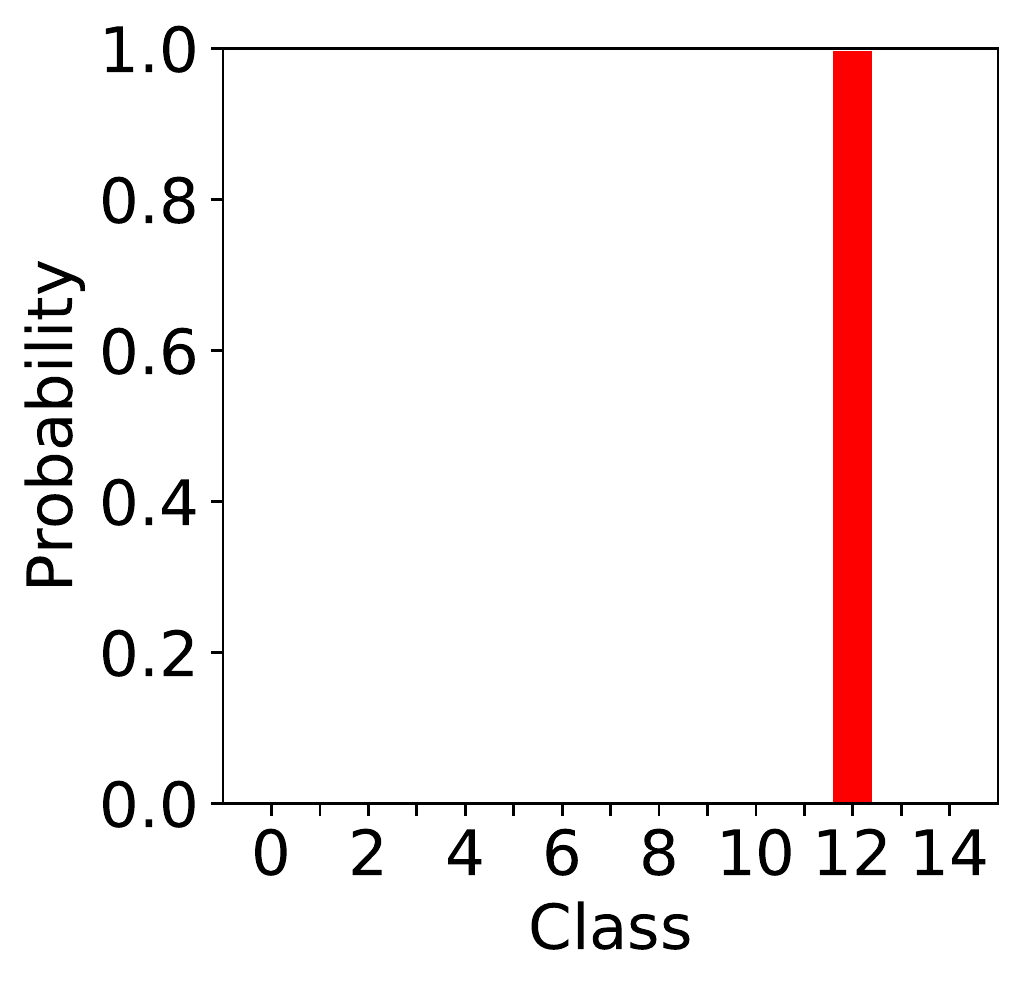} &
    \includegraphics[width=0.29\columnwidth]{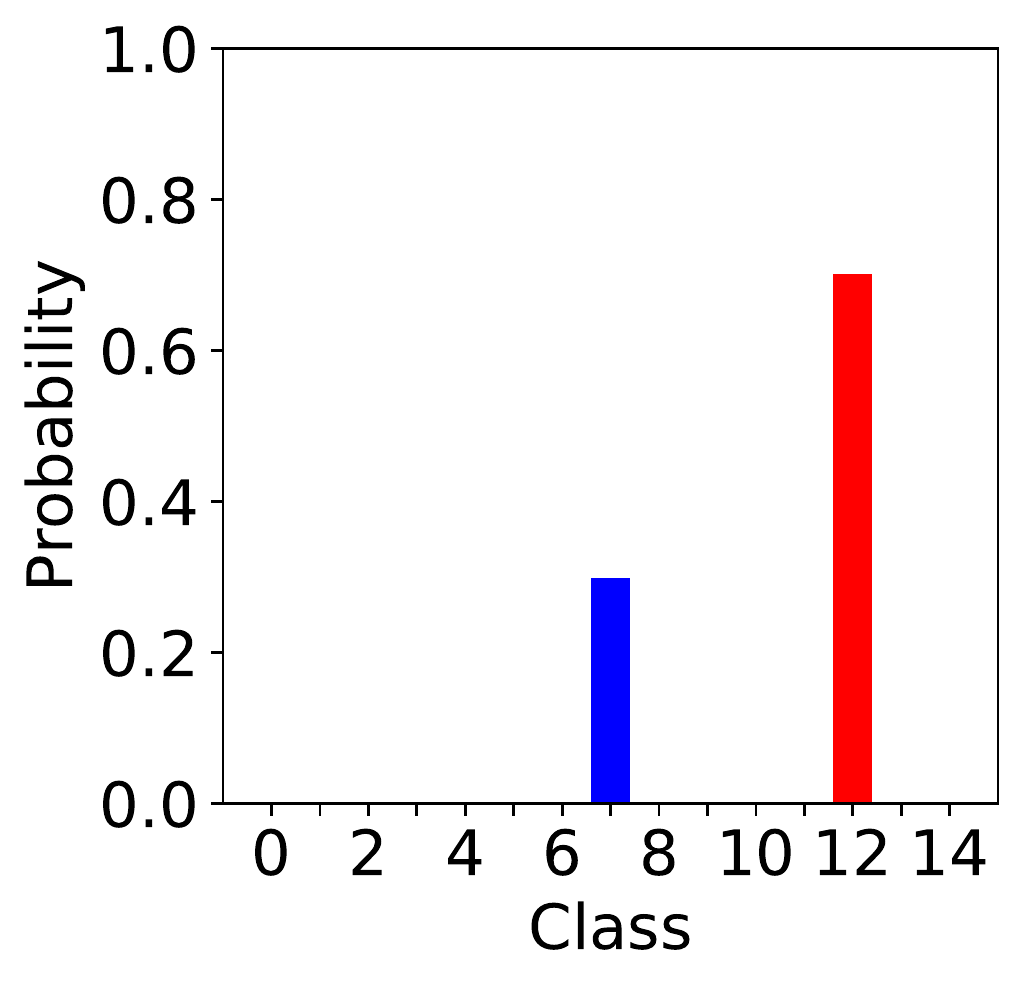} &
    \includegraphics[width=0.29\columnwidth]{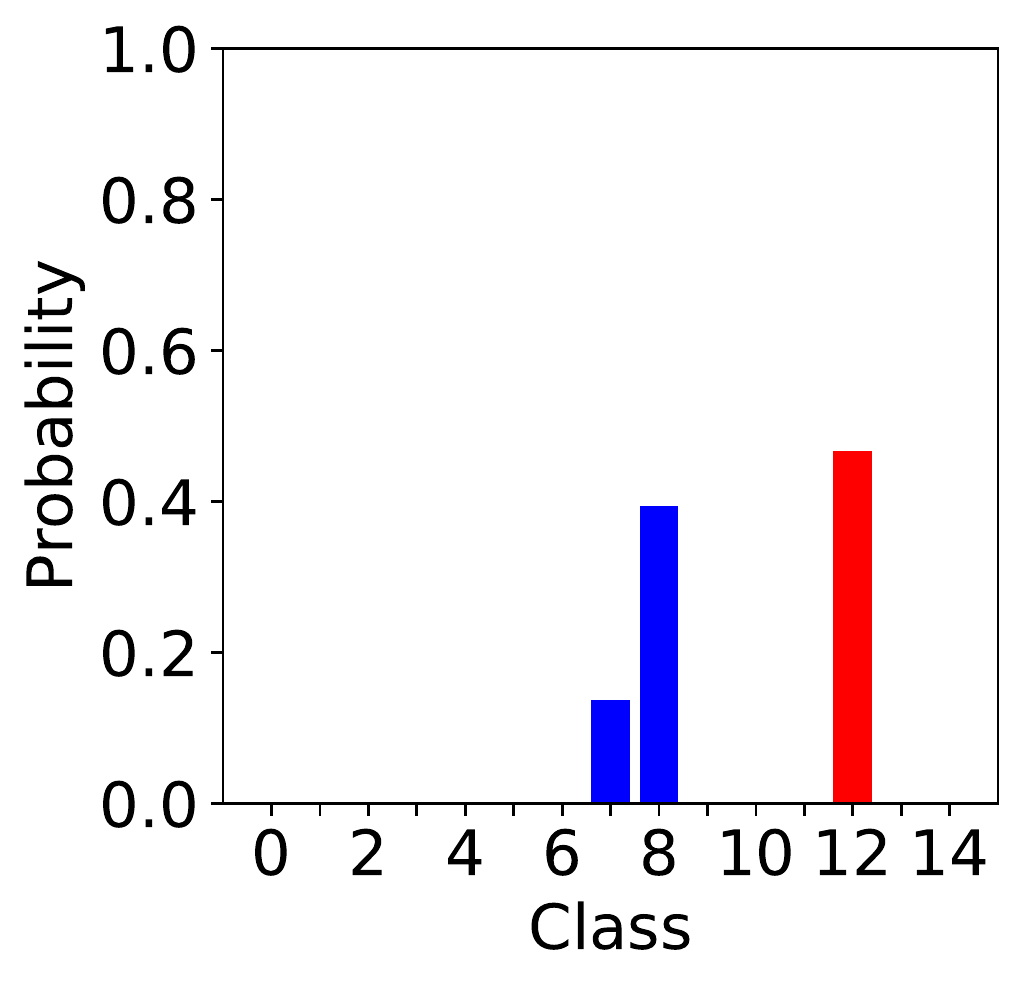} \\
    \includegraphics[width=0.29\columnwidth]{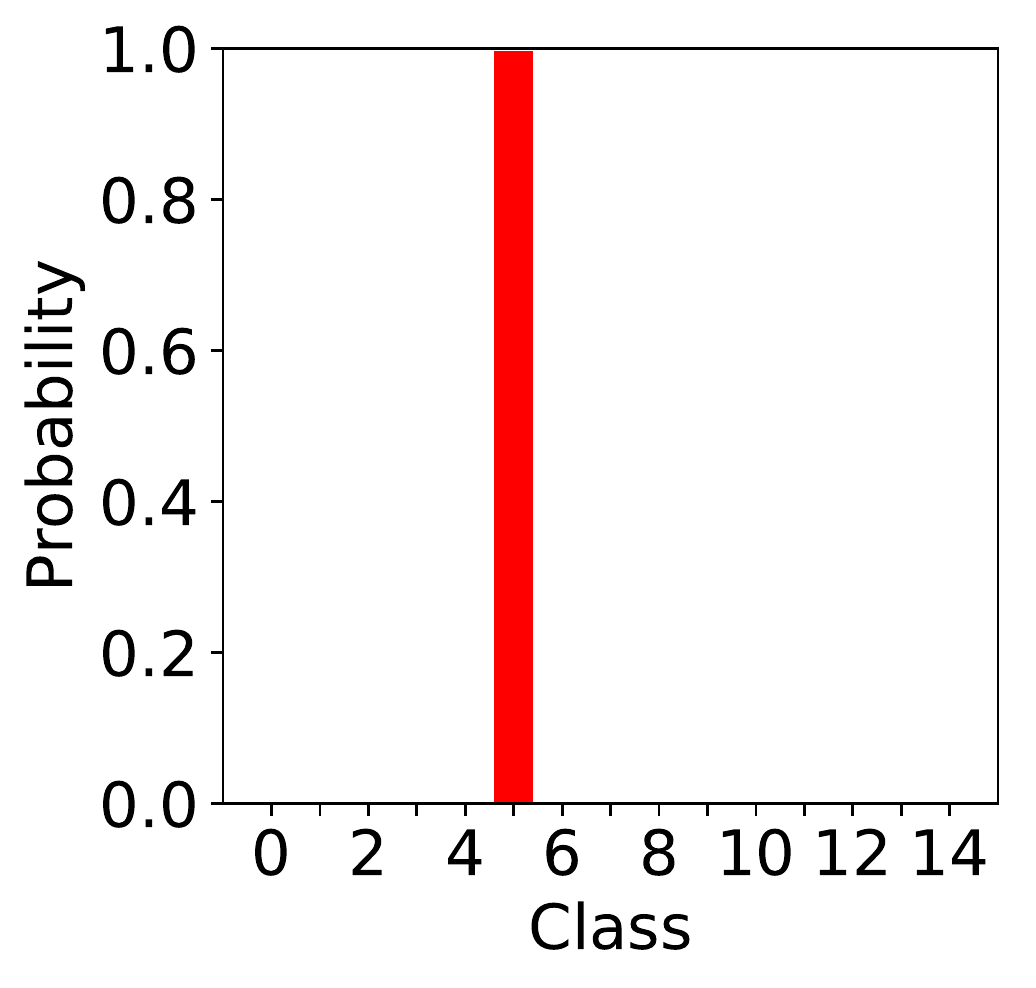} &
    \includegraphics[width=0.29\columnwidth]{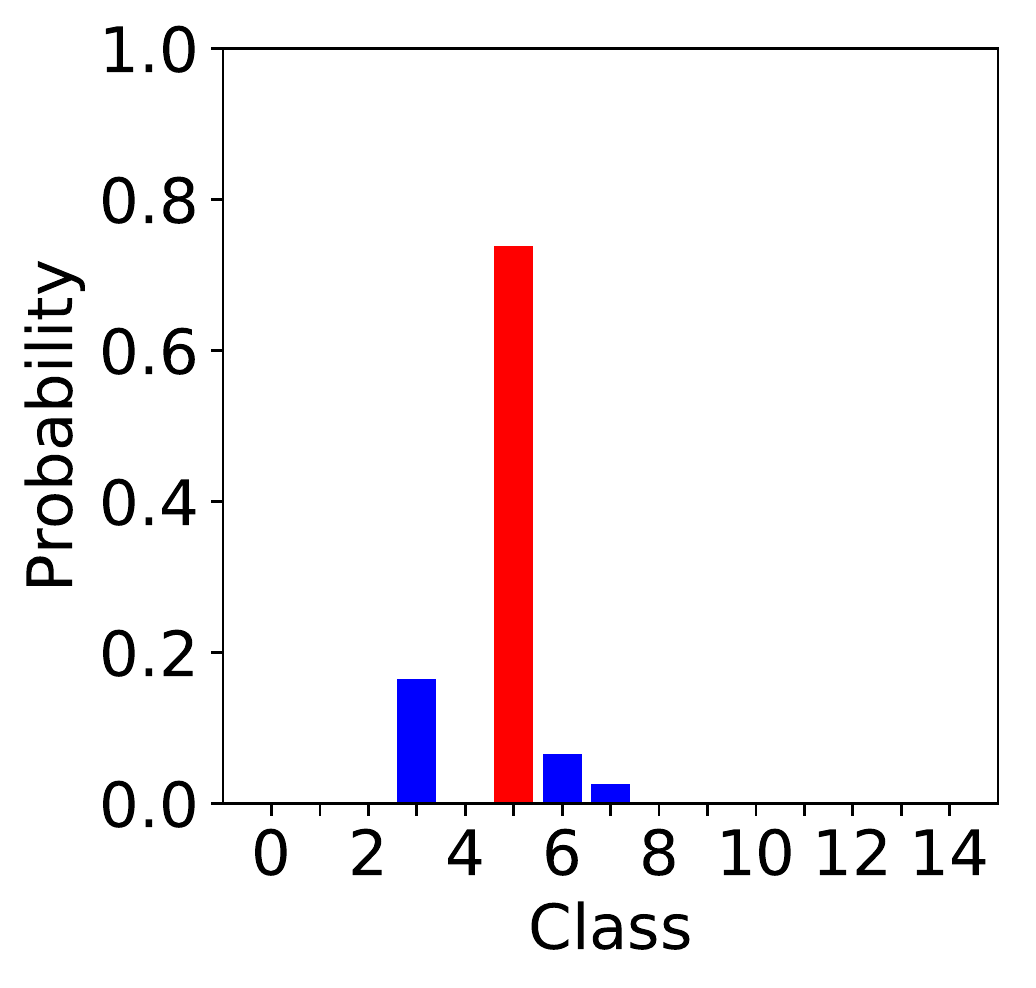} &
    \includegraphics[width=0.29\columnwidth]{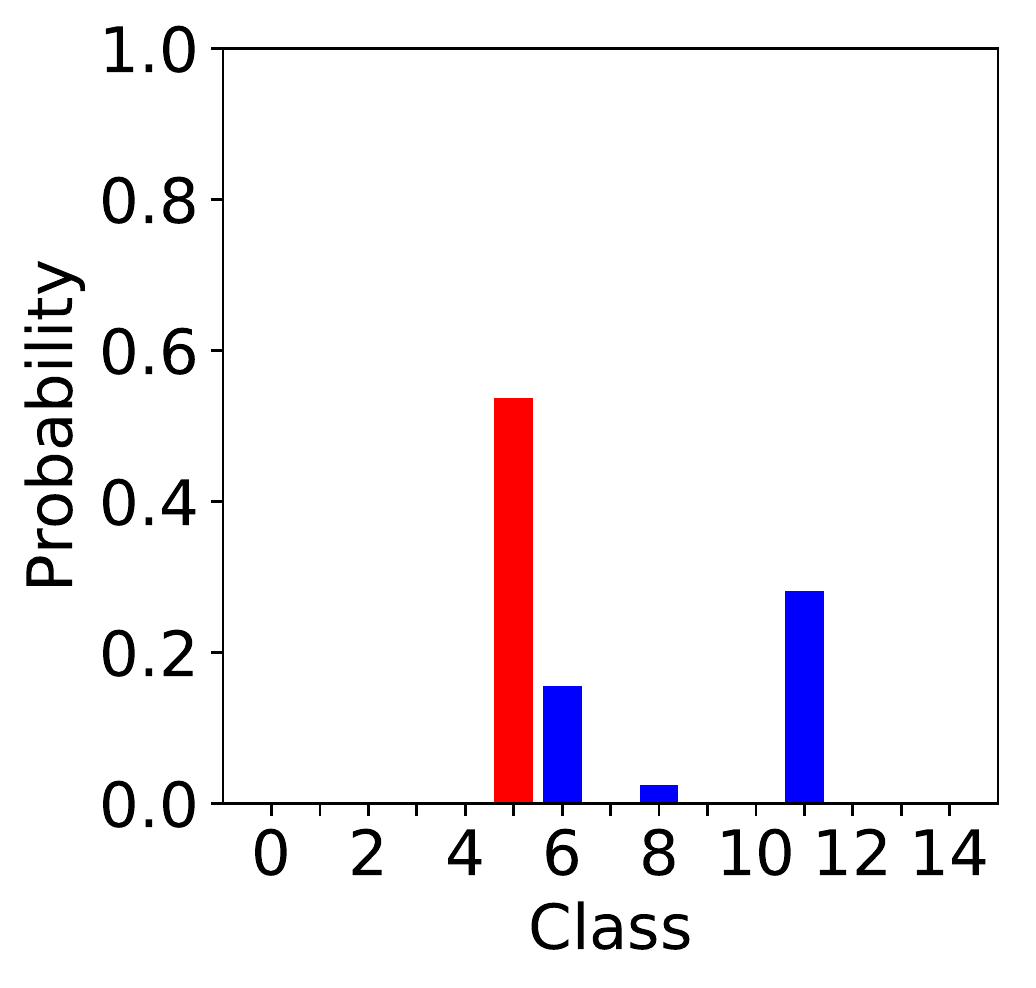} \\
    \includegraphics[width=0.29\columnwidth]{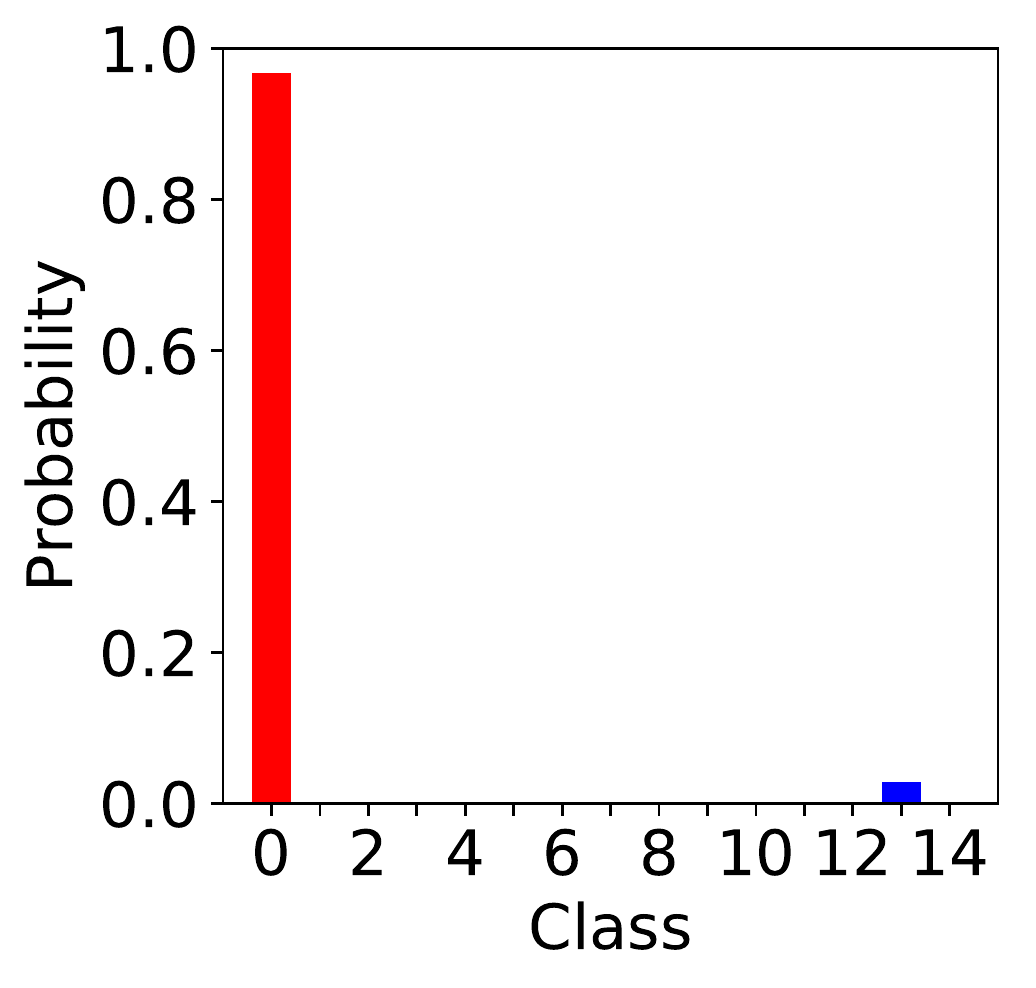} &
    \includegraphics[width=0.29\columnwidth]{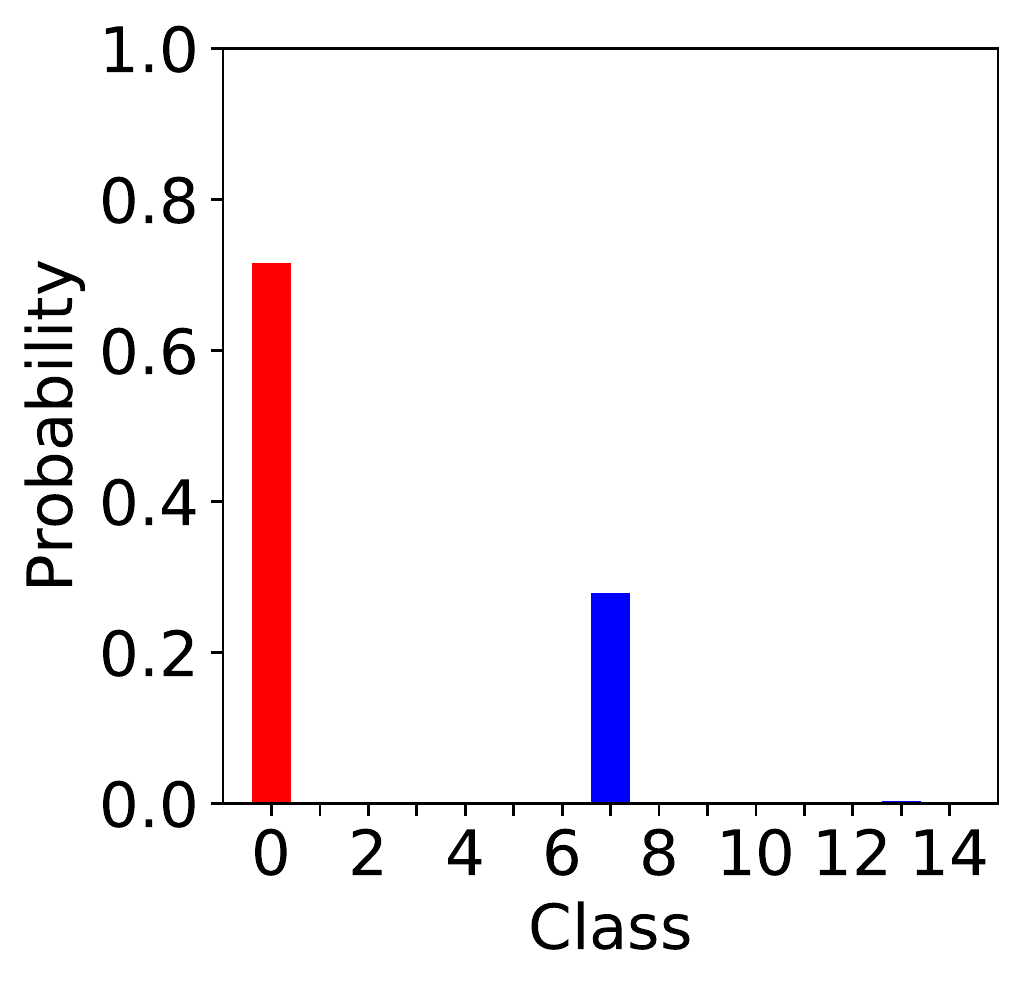} &
    \includegraphics[width=0.29\columnwidth]{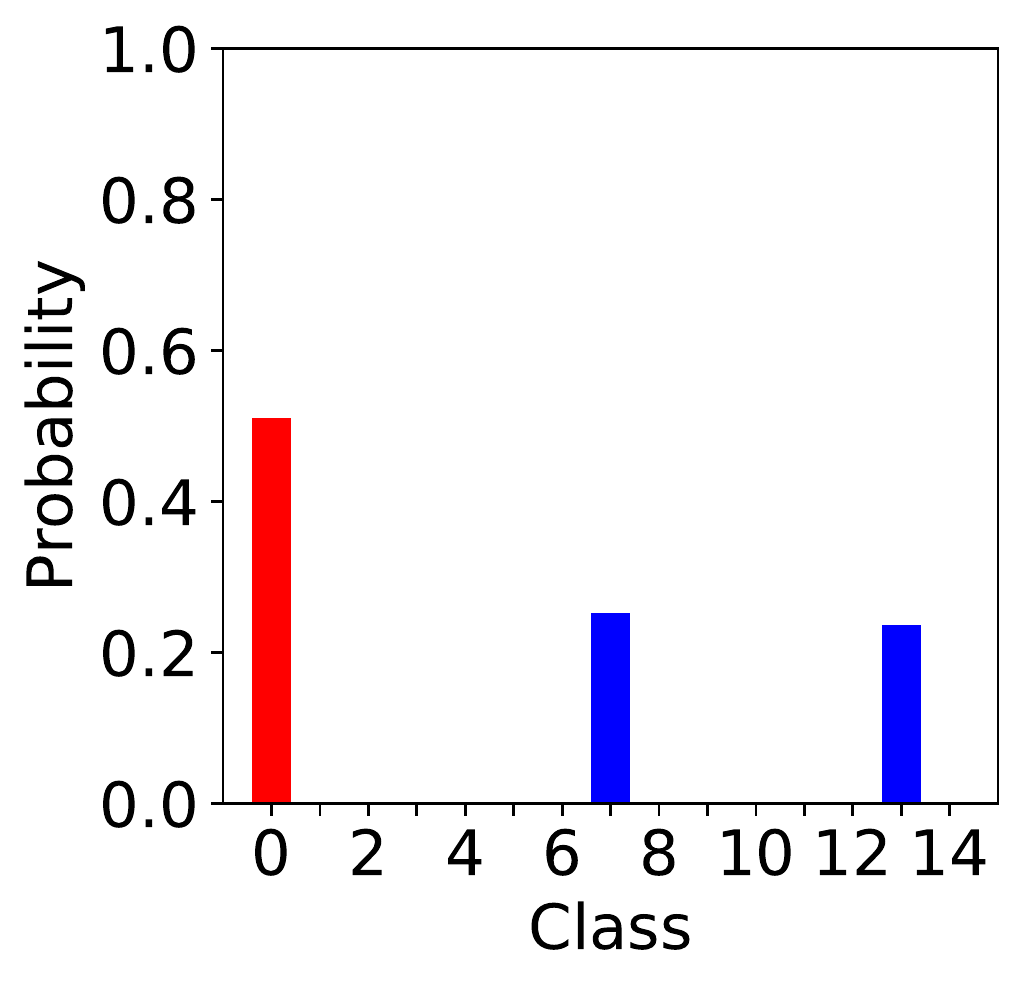}
    \end{tabular}
    \caption{Test predictive distribution obtained by the compared methods for four different test instances in the LIGO experiment.
    All the methods assign the highest probability to the correct class (which is highlighted in red).
    However, from left to right, the quality of the predictive distribution decreases, as greater probability is assigned to wrong classes. 
    This is related to the uncertainty quantification capabilities of each method, an aspect at which the GP modelling of SVGPCR stands out. 
    Moreover, these differences in the predictive distributions are behind the superiority of SVGPCR in test likelihood (table \ref{tab:LIGO_likelihood}).}
    \label{fig:LIGO_predDist}
\end{figure}

As already pointed out in Table \ref{tab:LIGO_likelihood}, the underlying GP model of SVGPCR implies an advantage over DL-based methods in terms of uncertainty quantification.
Importantly, the test likelihood metric is a \emph{global} measure of the quality of the predictive distribution obtained for each \emph{individual} test instance.
To clearly understand the benefits of the GP modelling, figure \ref{fig:LIGO_predDist} shows the predictive distributions for some test instances which are behind the better global performance of SVGPCR.
Only the best method (in terms of test likelihood) of each type (i.e. CL-based ones, SVGPCR ones, and AggNet) is considered, which yields the three columns in figure \ref{fig:LIGO_predDist}. 
Each row represents a different test instance.

Interestingly, we observe that the three approaches correctly classify the four instances, that is, they assign the highest probability to the correct class (which is highlighted in red). 
In particular, this means that these four instances contribute equally to the test accuracy of the three methods.
However, notice that the quality of the predictive distribution worsens from left to right (i.e., from better to worse uncertainty quantification theoretical properties), since the methods become less certain about the correct answer and assign more probability to wrong ones. 
This is precisely what is accounted for in the test likelihood metric.
From a practical perspective, this better quality of the predictive distributions has been particularly appreciated by astrophysicists at GravitySpy, in addition to the improvement in test accuracy (recall table \ref{tab:LIGO_accuracy}).


\begin{figure*}
    \centering
    \includegraphics[width=0.95\textwidth]{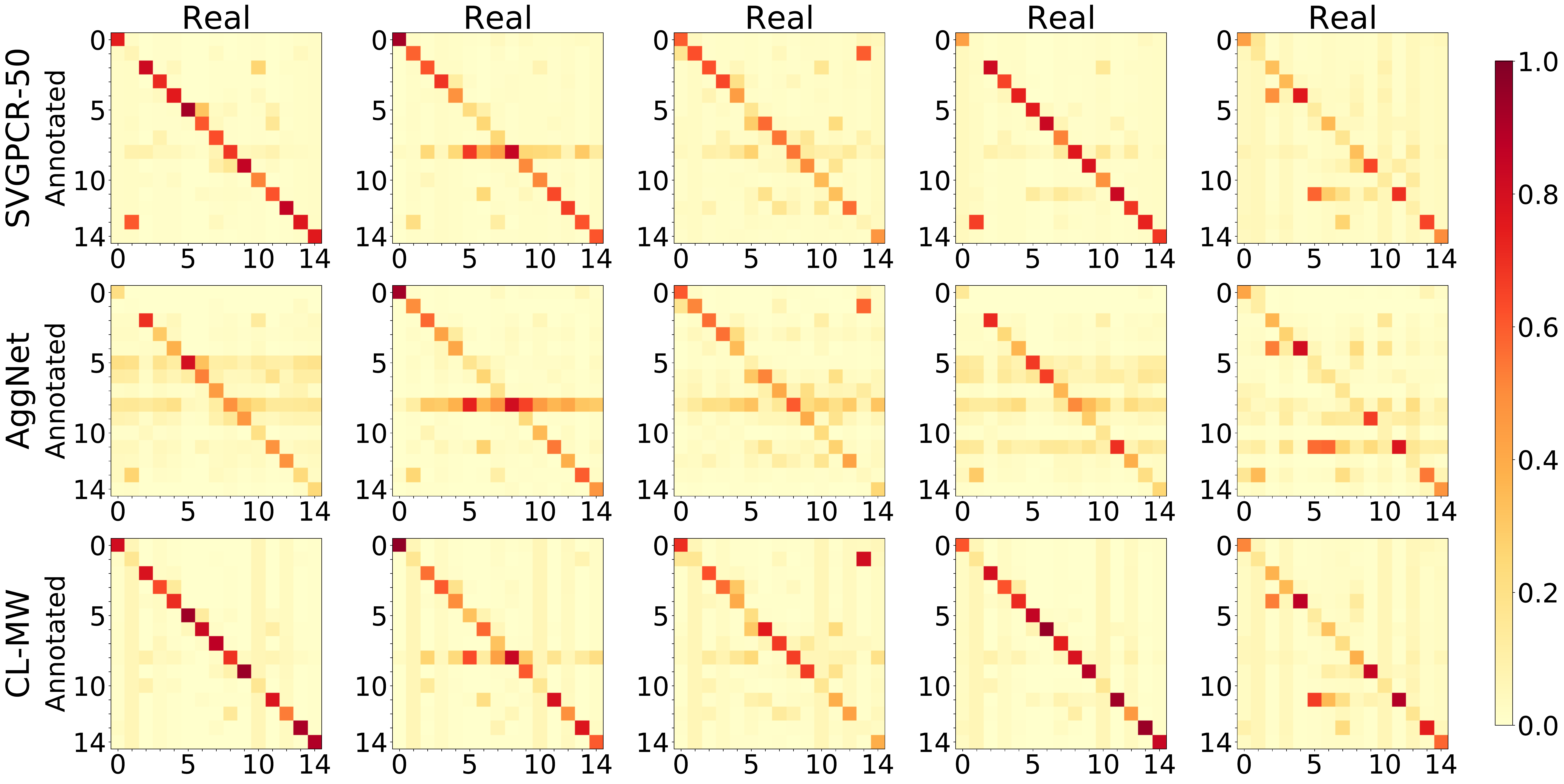}
    \caption{Confusion matrices estimated by the three families of algorithms (rows) and for five different annotators (columns) in the LIGO experiment. In each matrix, the real class is represented in the x-axis, and the annotated one in the y-axis (that is, the matrices add up to one by columns).
    For every annotator, the general structure of the three estimated matrices is similar. In particular, the highest values are located in the diagonal, which means that annotators have been correctly instructed in general. 
    Moreover, recall that SVGPCR provides a full probability distribution (in particular uncertainties) for these predictions.
    }
    \label{fig:LIGO_confMatrices}
\end{figure*}

Finally, a key aspect of crowdsourcing methods is the identification of the different annotators behavior. 
Unlike in section \ref{sec:exp_MNIST}, where we had simulated annotators to check the good estimations of SVGPCR, in this real experiment we do not have available a ground-truth. 
Nonetheless, let us compare the predictions obtained by the different methods. 
We will see that they capture similar patterns, some of which can be explained from the experience of astrophysicists. 
Figure \ref{fig:LIGO_confMatrices} shows the confusion matrices predicted by the compared methods for five different annotators. In the CL-based family we only consider CL-MW, as it is the best in test likelihood and the only one which provides a confusion matrix.

One of the most distinctive features for all instances and methods is the predominance of high values in the diagonal.
This was considered as a positive feedback by astrophysicists, as it means that annotators have been generally well instructed to distinguish among glitches.
Additionally, other patterns out of the diagonal are worth an analysis.
For the first column (first volunteer), SVGPCR and AggNet detect that glitches of type 1 (i.e. ``1400Ripple'', recall figure \ref{fig:glitches_full}) are classified as class 13 (``Violin Mode Harmonic''). 
This is a very frequent mistake according to experts, since the general appearance of both glitches is similar.
We also observe that CL-MW does not agree on this prediction. 
This discrepancy of CL-MW for some particular patterns is recurrent across different annotators, and can be attributed to the different modelling of the annotators noise (non-probabilistic one, but through weights in the DNN). 
The second column shows a typical conservative annotator, who resorts too frequently to the catch-all ``Other'' class. This is reflected in the persistent high values of the row number 8 in the matrices, regardless of the column (the real class).
For the third column, the three methods identify the confusion from ``Violin Mode Harmonic'' to ``1400Ripple''. Notice that this is the opposite to the first annotator, where the confusion was the other way round.
In the fourth annotator, AggNet exhibits a noisy behavior compared to SVGPCR and CL-MW. Although perhaps less explicitly, this can be also observed across different annotators, and might be due to the iterative nature of AggNet, which does not allow for an end-to-end learning and leaves some extra noise after training.
In the fifth annotator, the three methods identify a very common confusion, which is labelling instances whose real class is ``Blip'' as ``Koifish'' (classes 2 and 4, respectively). 
Although these glitches seem pretty different in the paradigmatic examples shown in figure \ref{fig:glitches_full}, wider ``Blip'' and narrower ``Koifish'' are frequent in the data set, and might mislead a non-expert volunteer.

Most importantly, the identification of all these wrong behaviors allows crowdsourcing methods to take full advantage of the noisy annotations.
It is also worth noticing that the Bayesian nature of SVGPCR provides uncertainties for the confusion matrices obtained here (recall the full posterior Dirichlet distributions in eq.~\eqref{eq:parametric_form4}), which is not available for the DL-based methods.

%% file: conclusion.tex
In this work we have introduced SVGPCR, a novel GP-based crowdsourcing classification algorithm that can scale up to very large data sets through its mini-batch training scheme.
The motivation for this methodology is the problem of glitch classification in the laureate LIGO project, which is addressed with crowdsourcing techniques in the GravitySpy sub-project.
To that end, and in order to obtain accurate predictive distributions, astrophysicists were interested in combining the excellent uncertainty quantification of GP-based crowdsourcing methods with the scalability of those based on deep learning (DL). 
The proposed SVGPCR resorts to the most popular sparse GP approximations in Machine Learning in order to make such combination a reality, and brings back GP-based methods to the state of the art in crowdsourcing.

SVGPCR is competitive with DL-based approaches in terms of test accuracy and computational cost, and stands out in terms of predictive distribution quality.
Moreover, its behavior naturally follows its theoretical formulation: it provides very accurate estimations for the annotators expertise degree, and the inducing points influence the test performance and the computational cost as expected.
Moreover, the code is based on the popular GPflow library, which leverages GPU acceleration through TensorFlow.

In the LIGO problem, the glitches were given by relevant features extracted by astrophysicists. However, in the case of more complex data such as images, audio or natural language, DL-based methods can benefit from convolutional layers in the deep neural network. 
From a probabilistic perspective, this could be addresed through Deep Gaussian Processes \cite{salimbeni2017doubly} and the very recent attempts to introduce convolutional structure in GPs \cite{van2017convolutional, blomqvist2018deep}.